\RequirePackage{subcaption}
\documentclass[lettersize,journal]{IEEEtran}
\usepackage{listings}
\usepackage{makecell}
\usepackage{amsmath,amsfonts}
\usepackage{algorithmic}
\usepackage{algorithm}
\usepackage{xcolor}
\usepackage{array}
\usepackage{booktabs}
\usepackage{diagbox}
\usepackage[caption=false,font=normalsize,labelfont=sf,textfont=sf]{subfig}
\usepackage{textcomp}
\usepackage{stfloats}
\usepackage{hyperref}
\usepackage{url}
\usepackage{verbatim}
\usepackage{graphicx}
\usepackage{cite}
\usepackage{multirow} 
\usepackage{overpic}
\usepackage{wrapfig}
\usepackage{rotating}
\usepackage{soul}
\usepackage{placeins}

\newcommand{\nonconverted}[1]{\mbox{}}
\usepackage{float}
\usepackage{amsthm}

\newcounter{broj}

\newcounter{broj1}

\begin{document}

% \title{OpenBreastUS: Benchmarking Forward and Inverse Neural Operators with Breast Ultrasound Computed Tomography} 

\title{OpenBreastUS: Benchmarking Neural Operators for Wave Imaging Using Breast Ultrasound Computed Tomography}

\author{Zhijun Zeng, Youjia Zheng, Hao Hu, Zeyuan Dong, Yihang Zheng, Xinliang Liu, Jinzhuo Wang, Zuoqiang Shi, Linfeng Zhang, Yubing Li, He Sun
\thanks{Zhijun Zeng is with the Department of Mathematical Sciences, Tsinghua University, Beijing 100084, China (e-mail: zengzj22@mails.tsinghua.edu.cn).}%
\thanks{Youjia Zheng, Hao Hu Yihang Zheng and Jinzhuo Wang are with the College of Future Technology, Peking University, Beijing, China. (e-mail: 2401112587@stu.pku.edu.cn, 211200005@smail.nju.edu.cn, 2000012969@stu.pku.edu.cn, wangjinzhuo@pku.edu.cn). }
\thanks{Xinliang Liu is with the  Computer, Electrical and Mathematical Science and Engineering Division, King Abdullah University of Science and Technology, Thuwal, Saudi Arabia. (e-mail: xinliang.liu@kaust.edu.sa). }%
\thanks{Zeyuan Dong and Yubing Li are with the State Key Laboratory of Acoustics and Marine Information, Institute of Acoustics, Chinese Academy of Sciences, Beijing, China. (e-mail: dongzeyuan@mail.ioa.ac.cn, liyubing@mail.ioa.ac.cn). Yubing Li is the Corresponding author.}
\thanks{Zuoqiang Shi is with the Yau Mathematical Sciences Center, Tsinghua University, Beijing, China and Yanqi Lake Beijing Institute of Mathematical Sciences and Applications, Beijing, China. (e-mail: zqshi@tsinghua.edu.cn). }%
\thanks{Linfeng Zhang is with the DP Technology, Beijing, 100080, China. (e-mail: linfeng.zhang.zlf@gmail.com). }%
\thanks{He Sun is with College of Future Technology, Peking University, Beijing, China and National Biomedical Imaging Center, Peking University, Beijing, China(e-mail: hesun@pku.edu.cn). Corresponding author.}\thanks{Zhijun Zeng, Youjia Zheng and Hao Hu contributed equally to this work.}}
\markboth{Journal of \LaTeX\ Class Files,~Vol.~14, No.~8, August~2021}%
{Shell \MakeLowercase{\textit{et al.}}: A Sample Article Using IEEEtran.cls for IEEE Journals}

\IEEEpubid{}
% Remember, if you use this, you must call \IEEEpubidadjcol in the second

\maketitle

\begin{abstract}
Accurate and efficient simulation of wave equations is crucial in computational wave imaging applications, such as ultrasound computed tomography (USCT), which reconstructs tissue material properties from observed scattered waves. Traditional numerical solvers for wave equations are computationally intensive and often unstable, limiting their practical applications for quasi-real-time image reconstruction. Neural operators offer an innovative approach by accelerating PDE solving using neural networks; however, their effectiveness in realistic imaging is limited because existing datasets oversimplify real-world complexity. 
In this paper, we present OpenBreastUS, a large-scale wave equation dataset designed to bridge the gap between theoretical equations and practical imaging applications. OpenBreastUS includes 8,000 anatomically realistic human breast phantoms and over 16 million frequency-domain wave simulations using real USCT configurations. It enables a comprehensive benchmarking of popular neural operators for both forward simulation and inverse imaging tasks, allowing analysis of their performance, scalability, and generalization capabilities. 
By offering a realistic and extensive dataset, OpenBreastUS not only serves as a platform for developing innovative neural PDE solvers but also facilitates their deployment in real-world medical imaging problems. For the first time, we demonstrate efficient \textit{in vivo} imaging of the human breast using neural operator solvers.
\end{abstract}


\section{Introduction}

\IEEEPARstart{C}{omputational} imaging aims to recover hidden images by decoding wave–matter interactions from observed signals, with partial differential equations (PDEs) playing a central role in modeling wave propagation in acoustics, electromagnetics, and seismology. However, high‐wavenumber wave equations are notoriously difficult to solve: accurately resolving rapidly oscillating wavefields demands very fine spatial grids, which produce large, complex‐valued linear systems that are highly ill‐conditioned and indefinite. To address this challenge, data‐driven neural PDE solvers have emerged as fast, numerically stable surrogates that learn mappings from parameter spaces to physical fields, demonstrating success in turbulent flow modeling, weather forecasting, and materials design~\cite{lu_DeepONet_2019,li_neural_2020,li_fno_2021,lu_learning_2021}. Beyond network architecture, however, the surrogate’s fidelity depends critically on training data quality. Existing open‐source PDE datasets—such as PDEBench~\cite{takamoto2022pdebench}, OpenFWI~\cite{deng2022openfwi}, and WaveBench~\cite{liu2024wavebench}—often simulate overly simplified scenarios (e.g., small regions of interest, simple geometric boundaries, or unrealistic random media), as shown in Fig.~\ref{fig:data_compare}. These limitations can lead to overly optimistic performance estimates and restrict applicability to real‐world problems. To advance practical deployment of neural operators, we therefore require application‐driven, realistic, and large‐scale PDE datasets. 

\begin{figure*}[htbp!]
  \centering
  \includegraphics[width=\linewidth]{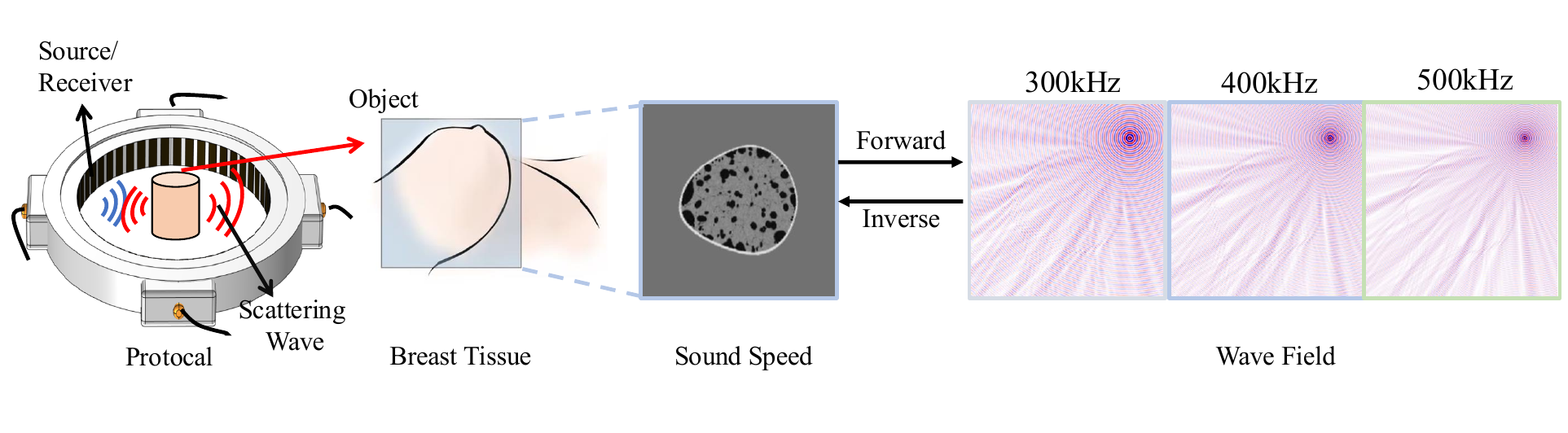}%
  \caption{\textbf{Schematic diagram of a USCT system and the OpenBreastUS dataset.}The imaging target is placed inside an annular transducer array, with each transducer emitting waves sequentially while the others act as receivers.The OpenBreastUS dataset includes anatomically realistic human breast phantoms and their corresponding wavefields at different frequencies.}
  \label{fig:usct_intro}
  \end{figure*}

Ultrasound computed tomography (USCT) is a promising wave‐imaging modality for medical diagnostics, offering high‐resolution 2D and 3D visualization of human tissues~\cite{guasch_full-waveform_2020,li_3-d_2022}. As illustrated in Fig.\ref{fig:usct_intro}, USCT uses a specialized transducer array—annular, cylindrical, or hemispherical—for fully automatic data acquisition. Unlike conventional B‐mode ultrasound, which relies on manual operation and only reflected signals, USCT sequentially emits waves from each transducer and measures both transmitted and reflected signals across the array\cite{cueto_spatial_2022,wu_ultrasound_2023,zhou_frequency-domain_2023}. In this setting, wave scattering within tissues is significant because ultrasonic wavelengths are comparable to tissue structures. USCT therefore employs wave PDEs to model the image‐formation process and solves a nonlinear PDE‐constrained inverse problem—known as full waveform inversion (FWI)—to reconstruct high‐dimensional tissue properties such as attenuation and sound speed~\cite{bernard_ultrasonic_2017,perez-liva_time_2017}. The computational intensity and numerical instability of traditional wave‐equation solvers make wave simulation a bottleneck for quasi‐real‐time USCT imaging, limiting its widespread clinical adoption~\cite{ali20242}. Consequently, USCT provides a rigorous and challenging benchmark for neural PDE solvers. 

In this paper, we introduce OpenBreastUS, a large‐scale USCT dataset designed to benchmark neural operators in wave imaging. OpenBreastUS bridges theoretical wave equations with a practical medical imaging application by providing over 16 million frequency‐domain wave simulations (breast phantoms $\times$ frequencies $\times$  source locations $\to$  wavefields: 8,000 $\times$  8 $\times$  256 $\to$ 16,384,000). The dataset features anatomically realistic human breast phantoms across four categories and simulates wavefields under settings that mirror a real USCT system (e.g., exact transducer positions and operating frequencies). OpenBreastUS thus offers a unified platform for designing optimal neural operator algorithms and for training models deployable in clinical imaging systems. We evaluate several popular machine‐learning surrogates for forward simulation and for inverse reconstruction, using relative root mean square error (RRMSE) and max error to assess simulation accuracy and structural similarity index (SSIM) and peak signal‐to‐noise ratio (PSNR) to evaluate image reconstructions. In experiments on two \textit{in vivo} human breast datasets, we demonstrate the strengths and limitations of existing neural operators and show that, with OpenBreastUS, neural PDE solvers can generalize effectively to realistic breast ultrasound imaging tasks.

\section{Related Work}\label{sec:related-work}
\subsection{Neural Operators} \label{subsec:neural-operator}
Neural operators are deep learning models designed to learn mappings between infinite-dimensional function spaces, offering data-driven solutions to PDEs. They are versatile tools for both forward simulations, predicting PDE solutions given parameters, and inverse problems, inferring underlying parameters from observations. Baseline frameworks include U-Net \cite{ronneberger2015u}, which employs a convolutional encoder–decoder architecture; the Fourier Neural Operator (FNO) \cite{li_fno_2021, li_neural_2020} and its variants—U-FNO (UFNO) \cite{wen_u-fnoenhanced_2022}, Born FNO (BFNO) \cite{zhao_deep_2023}, and Adaptive FNO (AFNO) \cite{guibas_adaptive_2022}, which leverage Fourier transforms to capture global information; and the Multigrid Neural Operator (MgNO) \cite{he2023mgno}, which integrates multigrid numerical schemes with neural networks to address multi-scale problems. In the inverse problem setting, approaches include InversionNet\cite{zeng_inversionnet3d_2022}—a convolutional network that directly models the inversion operator; Deep Operator Network (DeepONet)\cite{lu_DeepONet_2019,lu_learning_2021,cai_deepmmnet_2021,di_leoni_deeponet_2021,lin_operator_2021}, which employs a “branch and trunk” design to separate input functions from evaluation locations for efficient operator learning; and extensions such as Fourier-DeepONet\cite{zhu_fourier-deeponet_2023} and the Neural Inverse Operator (NIO)\cite{molinaro2023neural}, which combine DeepONet with FNO principles to integrate local and global representations, thereby improving accuracy and efficiency when mapping observations to PDE parameters.

\subsection{Physics Datasets} \label{subsec:dataset}
High-quality datasets are crucial for advancing deep learning approaches to PDEs, as they provide benchmarks for training and evaluating neural operator models. PDEBench \cite{takamoto2022pdebench} is a widely used benchmark dataset that covers various forms of PDEs primarily in fluid mechanics, such as Darcy flow, advection, diffusion, and Navier-Stokes equations, but it lacks wave propagation PDEs. OpenFWI \cite{deng2022openfwi} specifically targets wave equations for geophysical problems, benchmarking neural networks for direct inversion from partial seismic wavefield observations. Recently, WaveBench \cite{liu2024wavebench} has been introduced to benchmark neural operators for forward simulations using extensive datasets of time-harmonic and time-varying wave simulations.

\begin{figure*}[h!]
    \centering
    % 第一张图片
    \includegraphics[width=0.9\linewidth]{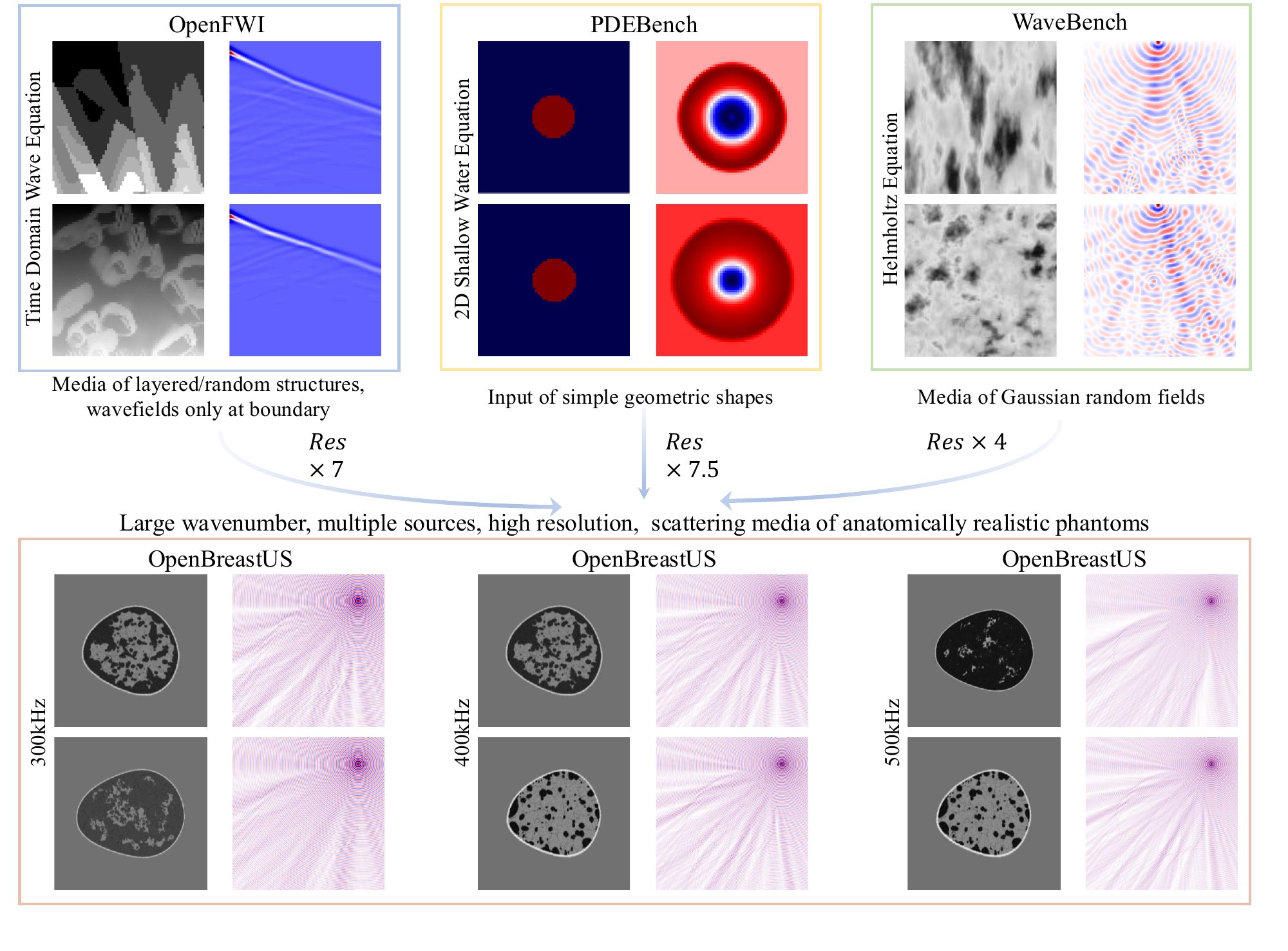}
    \caption{\textbf{Comparison of existing wave PDE dataset.} Representative data samples (scattering media and wavefields) from OpenFWI, PDEBench, WaveBench and OpenBreastUS datasets are illustrated.}
    \label{fig:data_compare}
\end{figure*}

Despite their contributions, both OpenFWI and WaveBench assume oversimplified scattering media or sources—OpenFWI uses layered structures, and WaveBench employs Gaussian random fields and MNIST 
 \cite{lecun1998gradient} with fixed source locations—and limit simulations to small ROIs (see Fig.~\ref{fig:data_compare}). 
These simplifications may lead to overly optimistic evaluations that fail to accurately assess neural operator performance in realistic applications, such as biomedical imaging scenarios where physical properties vary more complexly and ROIs exceed 100 wavenumbers. This underscores the need for a dataset that captures the complexities of real-world wave phenomena, motivating us to create OpenBreastUS, a more accurate benchmark for neural operator models in practical biomedical imaging settings. 
\label{app:figures}

\section{OpenBreastUS: A Realistic Application-Driven Benchmark for Wave Imaging} \label{sec:dataset}
In this section, we describe the USCT imaging problem addressed by the OpenBreastUS dataset, and provide detailed dataset statistics and its creation process.

\subsection{Problem Definition} \label{subsec:problem}
The primary goal of the OpenBreastUS dataset is to facilitate the development of neural operators and other deep learning techniques for real-world wave imaging applications, with USCT serving as a representative example. In the dataset, wave phenomena are modeled in the frequency domain, so the propagation of ultrasonic waves can be approximated by a heterogeneous Helmholtz equation, assuming negligible shear motion and nonlinear effects:
\begin{equation}
\left[ \nabla^{2}+\left( \frac{\omega}{c(x)} \right)^2 \right] u(x)=-s(x),
\label{eq:helmholtz}
\end{equation}
where $\omega$ is the angular frequency of ultrasound waves, $c(x)$ is the spatial distribution of sound speed in the scattering medium, $s(x)$ is the source term, and $u(x)$ is the resulting complex acoustic field. We further assume that the variation in sound speed, $c(x)$, is confined to a pre-defined region of interest (ROI), while outside this region, the sound speed remains constant at $c_0$. This results in the following Sommerfeld radiation condition at infinity:
\begin{equation}
\lim_{r\to \infty} r^{\frac{n-1}{2}} \left( \frac{\partial}{\partial r}-i\frac{\omega}{c_0} \right) u(x)=0.
\label{eq:sommerfeld}
\end{equation}
Equations \ref{eq:helmholtz} and \ref{eq:sommerfeld} define the forward model for the USCT imaging problem, namely the relationships between $\omega$, $c(x)$, $s(x)$, and $u(x)$—where $c(x)$ describes tissue's mechanical properties (sound speed), $\omega$ is the transducer’s operating frequency, $s(x)$ denotes the point source on the annular ring, and $u(x)$ represents the resulting ultrasound wavefield.

Consequently, USCT wave imaging can be formulated as a PDE‐constrained optimization problem that reconstructs the spatial distribution of sound speed $c(x)$ in biological tissues from transducer measurements:
\begin{equation}
\begin{aligned}
    \min_{c(x)} & \sum_{j=1}^N\sum_{k=1}^M L_k^j=\sum_{j=1}^N \sum_{k=1}^M \left\| \boldsymbol{y}_k^j - u_k^j(\boldsymbol{x}_f) \right\|_2^2 \\
    \text{s.t.}  & \left[ \nabla^{2} + \left( \frac{\omega_j}{c(x)} \right)^2 \right] u_k^j(x) = -s_k(x),
\end{aligned}
\label{eq:wave-imaging}
\end{equation}
where $k \in [1, M]$ indexes the transducers, $j \in [0, N]$ indexes the frequencies, $\boldsymbol{y}_k^j\in\mathbb{R}^{M}$ represents the measurements from all transducers when the $k$-th transducer is activated at the $j$-th frequency, and $\boldsymbol{x}_f\in\mathbb{R}^{M}$ denotes the transducer locations. $M$ and $N$ represent the number of transducers and frequencies, respectively. When a transducer is activated, it creates a point source $s_k(x)$. The total measurement for a given $c(x)$ forms a tensor  $\mathbf{Y}\in \mathbb{C}^{M \times M \times N}$. 
\subsection{Neural Operators for Wave Imaging}
Neural operators address the USCT wave imaging problem in two primary ways. In the first paradigm, one replaces the conventional numerical PDE solvers within the model-based iterative reconstruction (MBIR) framework by learned forward neural operators. By introducing Lagrange multipliers, the FWI problem in Eq.~\ref{eq:wave-imaging} can be reformulated as the unconstrained minimization of the Lagrangian
\begin{equation}
\begin{split}
\underset{c(x), u_k(x), \lambda_k(x)}{\min} \mathcal{L} = \sum_{j=1}^N\sum_{k=1}^M \mathcal{L}_k^j = \sum_{j=1}^N\sum_{k=1}^M \left\| \boldsymbol{y}_k^j - u_k^j(\boldsymbol{x}_f) \right\|_2^2 \\ 
- \sum_{j=1}^N\sum_{k=1}^M\left\langle \lambda_k^j(x),\mathcal{S}_c^j u_k^j(x)+s_k(x)\right\rangle 
\end{split}
\label{lagrange_suppl}
\end{equation}
where $\left\langle f,g \right\rangle$ denotes the real part of inner product of functions $f$ and $g$ in $L^2(\mathbb{C})$, $y_k^j$ denotes the transducer measurement for source $s_k$, $\lambda_k^j$ is the Lagrange multiplier, and $\mathcal{S}_c^j$ is the Helmholtz operator 
\begin{equation}
\mathcal{S}^j_c (\cdot) = \left[\nabla^{2}+\left(\frac{\omega_j}{c(x)}\right)^{2} \right](\cdot).
\label{S_operator_suppl}
\end{equation}	
Under the Karush–Kuhn–Tucker (KKT) conditions, optimal solutions to Eq.~\ref{lagrange_suppl} require the partial derivatives of $\mathcal{L}_k^j$ with respect to both $\lambda_k^j$ and $u_k^j$ to vanish. Setting $\frac{\partial \mathcal{L}_k^j}{\partial \lambda_k^j}(x) = 0$ recovers the forward Helmholtz equation, 
\begin{equation}
    \mathcal{S}_c^j u_k^j(x) = -s_k(x),
\label{eq:forwardprop}
\end{equation}
while enforcing $\frac{\partial \mathcal{L}_k^j}{\partial u_k^j}(x) = 0$ yields the adjoint equation,
\begin{equation}
\mathcal{S}_c^j \lambda_k^j(x)  =\sum_{i=1}^M [u_k^j(x_f^{(i)})-y_k^{j,(i)}] \delta(x_f^{(i)}),
\label{eq:backprop}
\end{equation}
in which $\delta(\cdot)$ denotes a normalized point source at each transducer location, and $i$ indexes the USCT transducers.
Substituting Eq.~\ref{S_operator_suppl} and Eq.~\ref{eq:backprop} into $\partial \mathcal{L}_k^j/\partial c$ shows that
\begin{equation}
\begin{split}
\frac{\partial \mathcal{L}^j_k}{\partial c}(x) &=\frac{\partial \mathcal{S}^j_c}{\partial c}(x)\lambda_k^{j\star}(x)u^j_k(x)\\
&=-2(\omega^j)^2 \frac{\lambda^{j\star}_k(x)u^j_k(x)}{c(x)^3},
\end{split}
\label{adjoint_suppl}
\end{equation}
so computing the gradient with respect to $c$ requires solving the Helmholtz equation twice per iteration—a major computational bottleneck in FWI reconstruction. To accelerate this, we train a forward neural operator $\mathcal{G}$ that approximates the solution map of the heterogeneous Helmholtz equation,
\begin{equation}
\mathcal{G}: \left(\omega,c\left(x\right),s\left(x\right)\right) \rightarrow u(x), x\in \Omega.
\label{eq:for-operator}
\end{equation}
Because wavefields at different frequencies exhibit distinct oscillatory behaviors, we train separate operator networks, $\mathcal{G}_{\omega_j}: \left(c\left(x\right), s\left(x\right)\right) \to u(x)$ for each frequency. These single‐frequency models are then combined in a mixture‐of‐experts (MoE) framework, $\mathcal{G}= \{\mathcal{G}_{\omega_1}, \cdots, \mathcal{G}_{\omega_N}\}$, where $N$ represents the total number of frequencies. Once the forward operator $\mathcal{G}$ is learned (denoted $\mathcal{P}(\omega_j, c, s_k)$), it can replace the Helmholtz solves in Eqs.~\ref{eq:forwardprop}–\ref{eq:backprop}, dramatically reducing the cost of each gradient computation in the iterative inversion.

In the second paradigm, one directly learns an inverse neural operator
\begin{equation} \begin{array}{cccc}
      \mathcal{G}^{-1}:& (\{\omega_j\}_{j=1}^N,\{s_k\}_{k=1}^M, \{\mathbf{y}_{k}^j\}_{j=1,k=1}^{N,M})&\rightarrow & c(x) 
\end{array}\end{equation}
which maps multi‐frequency measurements $\mathbf{Y}$ back to the tissue sound‐speed distribution $c(x)$. By bypassing the iterative adjoint method entirely, $\mathcal{G}_{\theta}^{-1}$ delivers a reconstructed image via a single forward pass, offering further acceleration over the neural-operator-enhanced MBIR approach.

In OpenWaves dataset, each entry comprises four components— $c(x)$, $\omega$, $s(x)$ and $u(x)$ — supporting the training of both forward and inverse neural operators for wave-imaging tasks.
\begin{table*}[!hbtp]
    \centering
    \begin{tabular}{c c c c c}
        \hline
        \multicolumn{5}{c}{\textbf{Data Statistics}} \\  % 添加第一部分的标题
        \hline
        \textbf{Breast Type} & \textbf{Frequency} & \textbf{\#Train/\#Test} & \textbf{\# Source} & \textbf{Storage}  \\ \hline
               Heterogeneous (HET) & 300$\sim$650 kHz & 1800/200 & 256 & 7.2TB\\ 
               Fibroglandular (FIB) & 300$\sim$650 kHz & 2700/300 & 256 & 10.8TB\\ 
       Fatty (FAT) & 300$\sim$650 kHz & 1800/200 & 256 & 7.2TB\\ 
       Extremely dense (EXD) & 300$\sim$650 kHz & 900/100 & 256 & 3.6TB\\ 
        \hline
        \multicolumn{5}{c}{\textbf{Physical Settings}} \\  % 添加第二部分的标题
        \hline
        \textbf{Grid Spacing} & \textbf{Resolution} & \textbf{Ring Diameter} & \textbf{Source Spacing} & \textbf{Source Value} \\ \hline
       0.5 mm & $480\times480$ & 220 mm & $\frac{2\pi}{256}$ rad & $0.195-0.0275i$\\ \hline
    \end{tabular}
     \caption{\textbf{Overview of OpenBreastUS.}  Dataset composition and physical settings for data generation.}
    \label{tab:data_descript}
\end{table*}

\subsection{Overview of the Dataset} \label{sebsec:overview_dataset}
\subsubsection{Physical Settings and Statistics}
OpenBreastUS includes 8,000 breast phantoms designed to represent the distribution of diverse human breast types in the population. As shown in Table~\ref{tab:data_descript}, the dataset is divided into four groups, each corresponding to a specific breast density type: \textbf{heterogeneous (HET), fibroglandular (FIB), all fatty (FAT), and extremely dense (EXD)}. The proportions of the four breast categories were derived from \cite{li20213} and slightly adjusted to account for the higher breast density in Asian populations. The wavefields are simulated using parameters from a real annular USCT system, which consists of 256 transducers arranged in a 220 mm diameter ring. The system operates at frequencies ($\omega/2 \pi$) ranging from 300 kHz to 1500 kHz, corresponding to acoustic wavelengths between 1 mm and 5 mm. We focus on 8 frequencies between 300 kHz and 650 kHz, sampled at 50 kHz intervals, resulting in ROIs with approximately 50 to 100 wavenumbers. For each breast phantom, wavefields are simulated by activating each transducer at all frequencies, generating a total of $8,000 \times 256 \times 8 = 16,384,000$ data entries.
\subsubsection{Data Generation}
The data generation involves two steps: 1) generating anatomically accurate breast phantoms, and 2) simulating the corresponding wavefields using real USCT parameters.\\
\textbf{Phantom Generation} The breast phantoms are generated using a medical simulation tool developed by the Virtual Imaging Clinical Trial for Regulatory Evaluation (VICTRE) project at the US Food and Drug Administration (FDA). \cite{li_3-d_2022} This tool produces 3D models of various breast anatomies, categorized into the four density types mentioned earlier (see details in Appendix~\ref{app:VICTRE}). These models are sliced into 2D tissue maps, and then scaled by a random factor to simulate breasts of varying sizes.  Distinct breast tissues (for example, skin, adipose tissue and muscle) are then segmented, and we assign a physically realistic sound speed with small random perturbations for each region. To replicate real experimental conditions, the area surrounding the breast models is filled with water.\\
\textbf{Wavefield Simulation} After generating the breast phantoms, we simulate the resulting wavefields at the USCT system’s source locations and operating frequencies using numerical solvers. We employ the Convergent Born Series (CBS) algorithm \cite{osnabrugge_convergent_2016}, an iterative solver for simulating the Helmholtz equation. CBS incorporates a preconditioner into the Born series to ensure convergence, making it reliable for simulating complex media with strong scattering properties.

% \begin{figure*}[htbp!]
% \centering
% \subfloat[]{\includegraphics[width=2.5in]{fig/gt.pdf}%
% \label{fig:gt_ma}}
% \hfil
% \subfloat[]{\includegraphics[width=2.5in]{fig/init.pdf}%
% \label{fig:init_ma}}
% \caption{(a) True velocity, (b) inital velocity for full Marmousi model, and (c) Noise-free/10dB Measured signals for 5Hz}
% \end{figure*}

\section{Experiments} \label{sec:experiments}
In this section, we present the evaluation results of baseline methods on the OpenBreastUS dataset. Section~\ref{subsec:setting} describes the evaluation metrics of neural operators for wave imaging. In Section~\ref{subsec:baselines}, we introduce existing baseline models for forward wave simulation and inverse imaging. Section~\ref{subsec:forward-experiment}  discusses the baseline performance on forward wave simulation and inverse wave imaging tasks, respectively. In Section~\ref{subsec:clinical data} ,we validate the alignment between OpenBreastUS data and real-world data by reconstructing \textit{in vivo} clinical breast USCT dataset with models trained on OpenBreastUS dataset. In Section~\ref{subsec:analysis}, we provide additional analysis on the complexities introduced by different breast types and frequencies, as well as examine the scalability and generalization capabilities of the baselines.
\subsection{Experimental Setting}\label{subsec:setting}
Deep learning models here may serve either as surrogate models for forward wave simulation in PDE‐constrained iterative optimization or as approximators of direct imaging maps. Accordingly, we employ distinct metrics to evaluate their performance in the USCT imaging problem. In particular, baseline models for forward simulation surrogates are assessed using the relative root mean square error (RRMSE) between the ground-truth physics field $u$ and the model predicted field $\hat{u}$ and the Max Error (maximum RRMSE across samples). For the inverse wave imaging results, the quality of the reconstructed sound speed map are assessed using the Structural Similarity Index Measure (SSIM) and the Peak Signal‐to‐Noise Ratio (PSNR). The dataset and source codes are available at \url{https://ai4scientificimaging.org/OpenBreastUS/}.  

\begin{table*}[!hbp]
\centering
\begin{tabular}{c|c|ccccc}
\hline
\multirow{2}{*}{\textbf{Frequency(kHz)}} & \multicolumn{1}{c|}{\multirow{2}{*}{\textbf{Metric}}}              & \multicolumn{5}{c}{\textbf{Models}}                                          \\ \cline{3-7} 
                                    & \multicolumn{1}{c|}{}                                              & \textbf{UNet} & \textbf{FNO} &\textbf{AFNO}         & \textbf{BFNO}          & \textbf{MgNO}         \\ \hline
\multirow{2}{*}{300}               & RRMSE$\downarrow$                                                  &  0.1236       &  0.0269      &  0.0165              & $\underline{0.0113}$   & \textbf{0.0028}       \\
                                   & Max Error$\downarrow$                                              &  0.2551       &  0.0617      & $\underline{0.0293}$ &    0.0519              & \textbf{0.0092}       \\ \hline
\multirow{2}{*}{400}               & RRMSE$\downarrow$                                                  &  0.1503       &  0.0426      &  0.0242              &   $\underline{0.0148}$ & \textbf{0.0036}       \\
                                   & Max Error$\downarrow$                                              &  0.3017       &  0.1172      & $\underline{0.0464}$ &    0.0840              & \textbf{0.0178}       \\ \hline
\multirow{2}{*}{500}               & RRMSE$\downarrow$                                                  &  0.1798       &  0.0490      &  0.0276              &  $\underline{0.0209}$  & \textbf{0.0049}       \\
                                   & Max Error$\downarrow$                                              &  0.3571       &  0.1432      & $\underline{0.0639}$ &    0.0838              & \textbf{0.0262}       \\ \hline
\multirow{2}{*}{600}               & RRMSE$\downarrow$                                                  &  0.1969       &  0.0644      &  0.0383              &  $\underline{0.0285}$  & \textbf{0.0092}       \\
                                   & Max Error$\downarrow$                                              &  0.3882       &  0.2007      & $\underline{0.0931}$ &    0.1249              & \textbf{0.0524}       \\ \hline
\end{tabular}
\caption{\textbf{Quantitative evaluation of forward simulation baselines.} Performance was evaluated on the test set using RRMSE and Max Error. \textbf{Bold}:Best, \underline{Underlined}:Second Best}
\label{tab:forward-baselines}
\end{table*}

\subsection{Existing Baselines for Forward and Inverse Neural Operators} \label{subsec:baselines}
We benchmark several existing methods for both wave simulation and wave imaging tasks on OpenBreastUS. All baselines are implemented in PyTorch, with detailed architectures provided in Appendix~\ref{app:implementation}. The model sizes and corresponding inference times are summarized in Appendix Table~\ref{tab:baseline-overview}.

\subsubsection{Baselines for Forward Wave Simulation}
For forward modeling, we include UNet, FNO, BFNO, AFNO, and MgNO as baseline methods:\\
\textbf{UNet} \cite{ronneberger2015u} is a convolutional neural network with an encoder-decoder architecture and skip connections, effective for capturing multiscale features in images\\
\textbf{Fourier Neural Operator (FNO)} \cite{li_fno_2021} uses Fourier transforms to parameterize integral operators, efficiently learning mappings between function spaces for solving PDEs.\\
\textbf{Adaptive Fourier Neural Operator (AFNO)} \cite{guibas_adaptive_2022} enhances FNO by adaptively selecting Fourier modes through an attention mechanism, improving performance on high-resolution inputs and discontinuities.\\
\textbf{Born Fourier Neural Operator (BFNO)} \cite{zhao_deep_2023}  modifies FNO by incorporating the iterative Born approximation, sharing parameters across layers to better model wave scattering.\\
\textbf{Multigrid Neural Operator (MgNO)} \cite{he2023mgno} integrates multigrid techniques with neural operators for efficient and accurate modeling of multiscale phenomena.
\subsubsection{Baselines for Inverse Wave Imaging}
For inverse imaging, we benchmark DeepONet, InversionNet, and NIO for direct inversion, and also evaluate optimization-based imaging (MBIR framework) with forward neural surrogates:\\
\textbf{DeepONet} \cite{lu_DeepONet_2019} employs a branch-trunk architecture to map observations to PDE parameters.\\
\textbf{InversionNet} \cite{zeng_inversionnet3d_2022} proposes a CNN-based network, leveraging the exceptional capability of CNNs in handling image-related tasks.\\
\textbf{Neural Inverse Operator (NIO)} \cite{molinaro2023neural} combines DeepONet and FNO, with an added bagging mechanism to improve inversion accuracy and generalizability.\\
\textbf{Gradient-based Optimization} \cite{zeng2023neural} solves the inverse problem using conventional gradient-based adjoint method (Eq.~\ref{adjoint_suppl}) but replaces traditional numerical wave equation solvers with the more efficient neural operators (Eq.~\ref{eq:for-operator}).

\subsection{Benchmarks for Forward and Inverse Baselines} \label{subsec:forward-experiment}
We first evaluated five forward simulation baselines — UNet, FNO, BFNO, AFNO, and MgNO — using a subset of OpenBreastUS dataset comprising wavefields at three frequencies (300, 400, and 500 kHz) from 64 uniformly sampled sources out of 256. All models were trained with relative L2 loss on four NVIDIA A800 PCIe 80 GB GPUs. Further implementation details are provided in the Appendix ~\ref{app:implementation}. 

\begin{figure*}[t]
    \centering % Center the entire figure content
    % ROW 1: HET
    \noindent % Prevent potential paragraph indentation for the line of minipages
    \begin{minipage}[b]{0.05\textwidth}
        \centering
        \rotatebox{90}{\quad\quad\quad HET}
    \end{minipage}%
    \begin{minipage}[b]{0.1285\textwidth}
        \centering
        Sound Speed \\ \vspace{1mm} % Caption text
        \includegraphics[width=\linewidth]{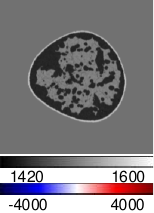}
    \end{minipage}%
    \begin{minipage}[b]{0.12\textwidth}
        \centering
        CBS \\ \vspace{1mm} % Caption text
        \includegraphics[width=\linewidth]{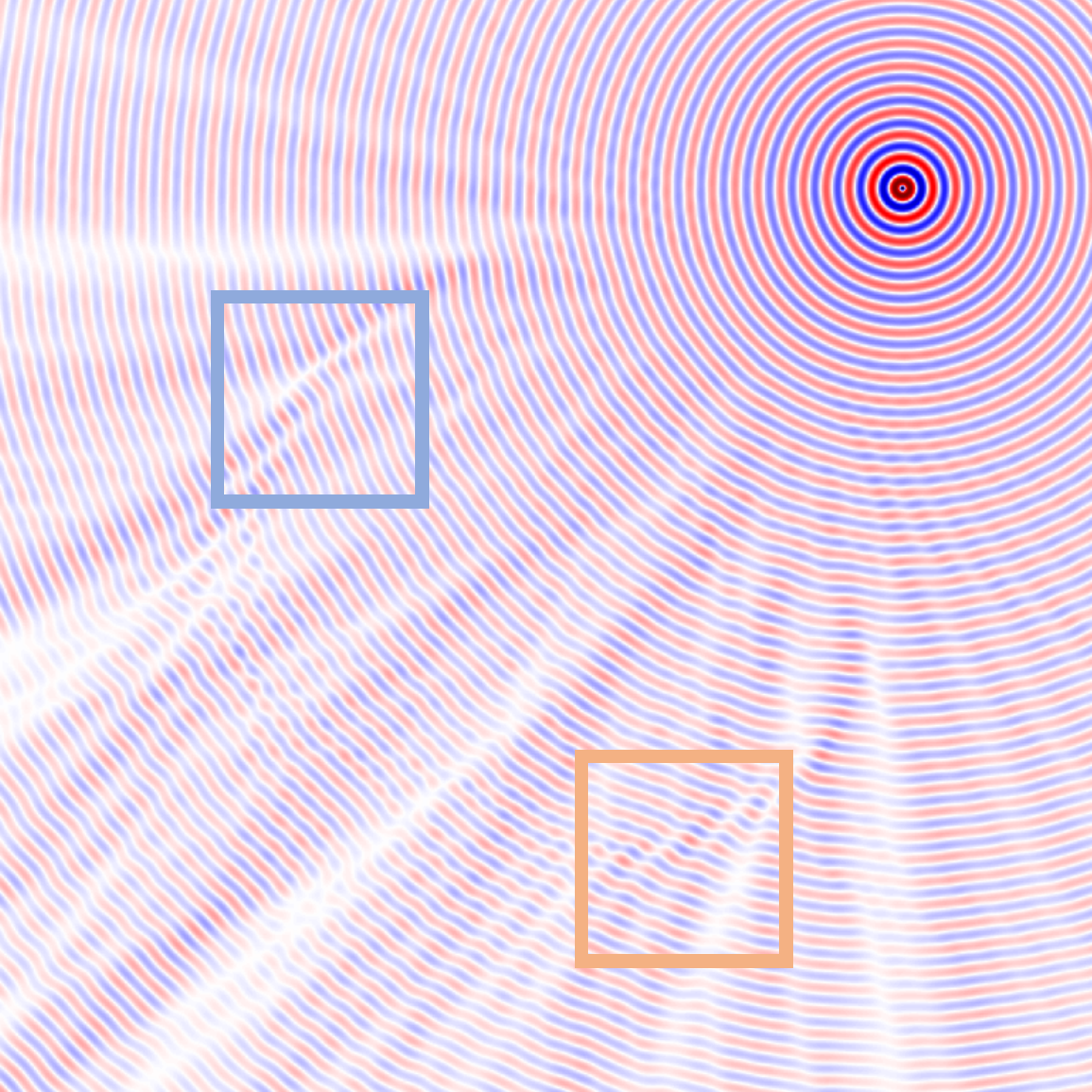} \\ \vspace{0.5mm}
        \includegraphics[width=0.48\linewidth]{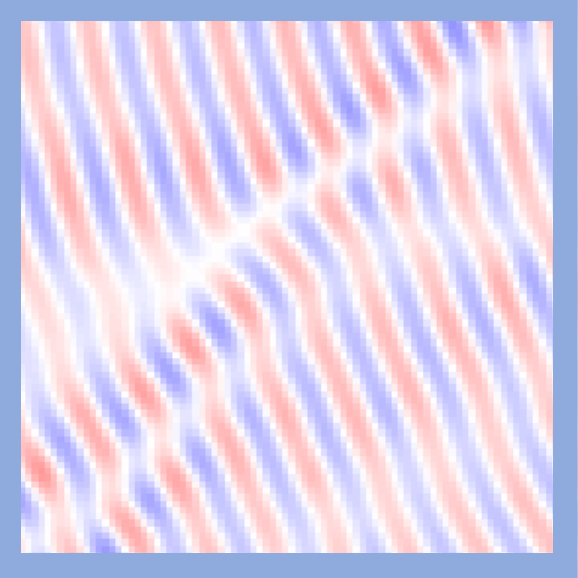}\hfill%
        \includegraphics[width=0.48\linewidth]{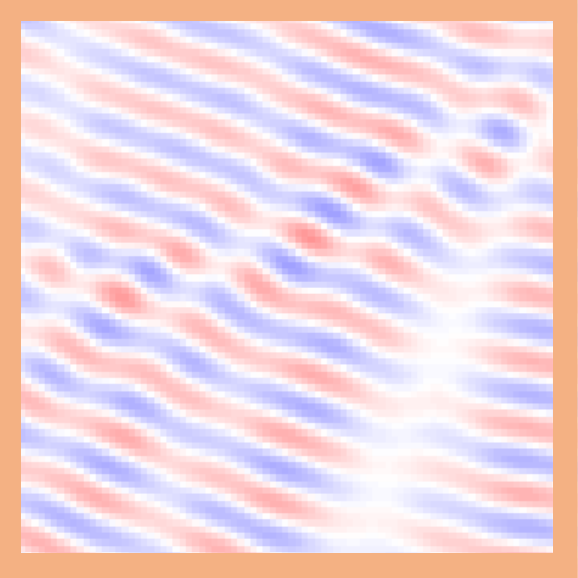}
    \end{minipage}%
    \begin{minipage}[b]{0.12\textwidth}
        \centering
        UNet \\ \vspace{1mm} % Caption text
        \includegraphics[width=\linewidth]{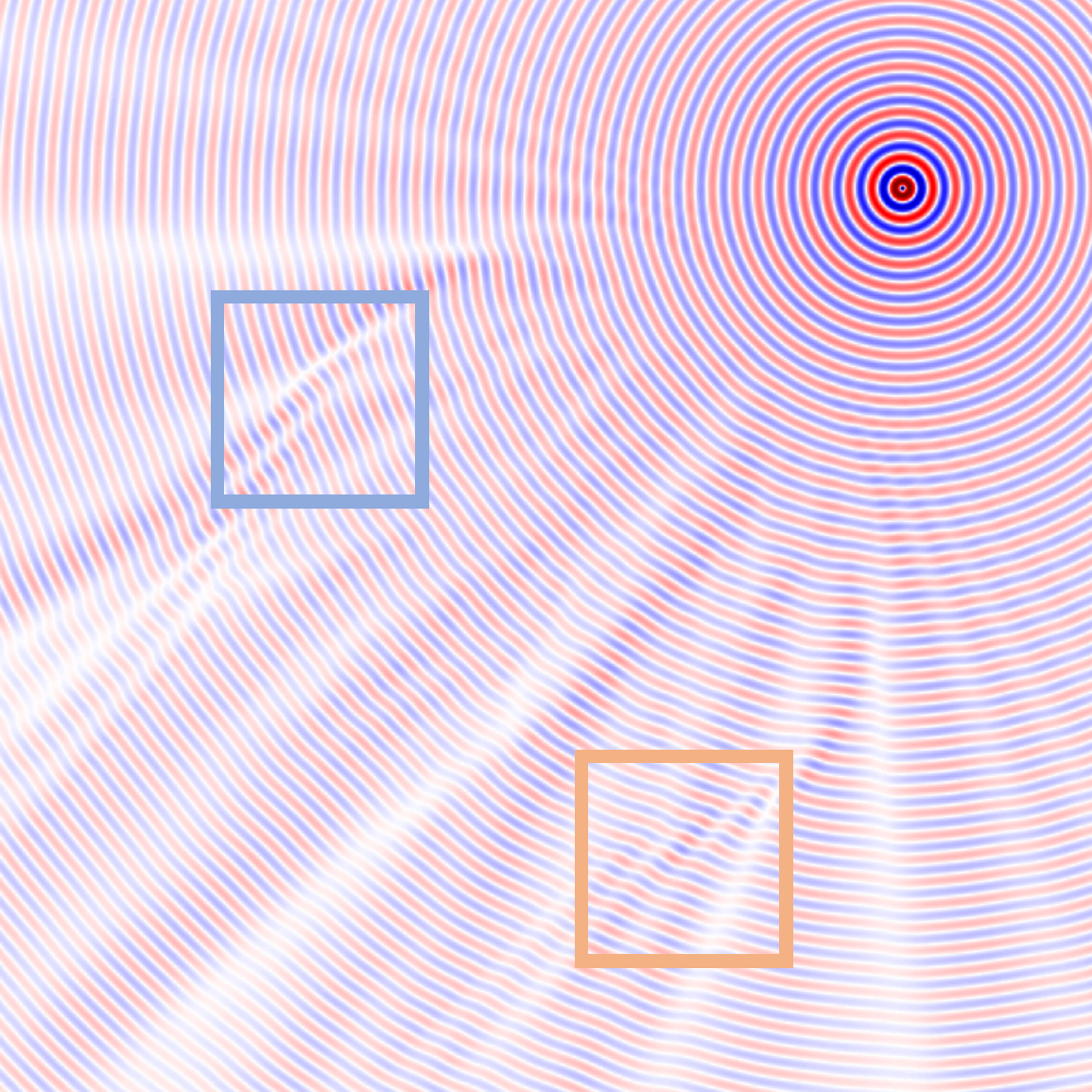} \\ \vspace{0.5mm}
        \includegraphics[width=0.48\linewidth]{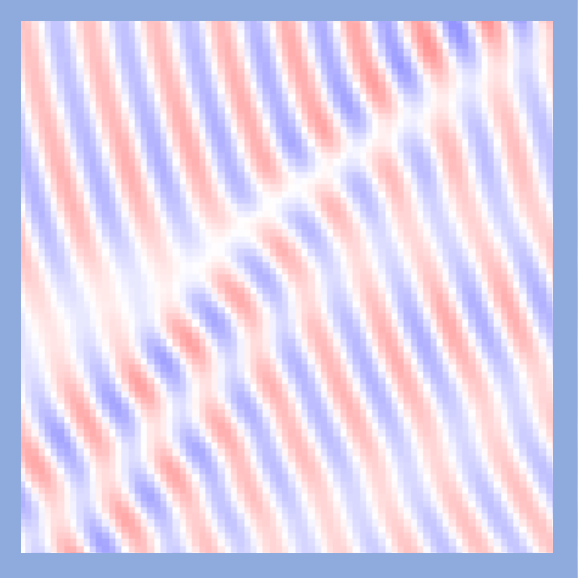}\hfill%
        \includegraphics[width=0.48\linewidth]{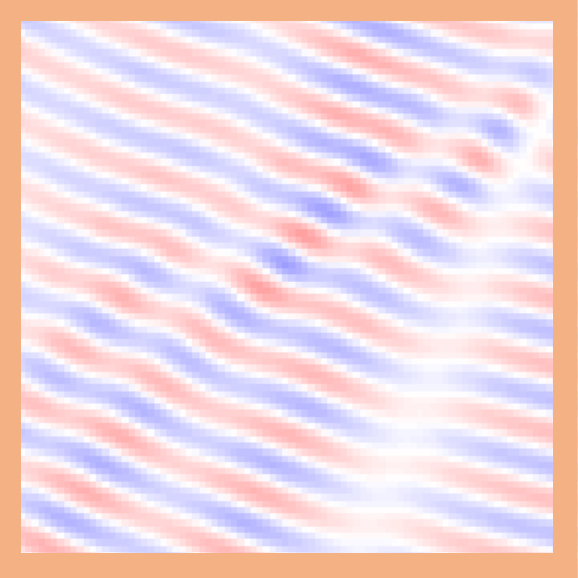}
    \end{minipage}%
    \begin{minipage}[b]{0.12\textwidth}
        \centering
        FNO \\ \vspace{1mm} % Caption text
        \includegraphics[width=\linewidth]{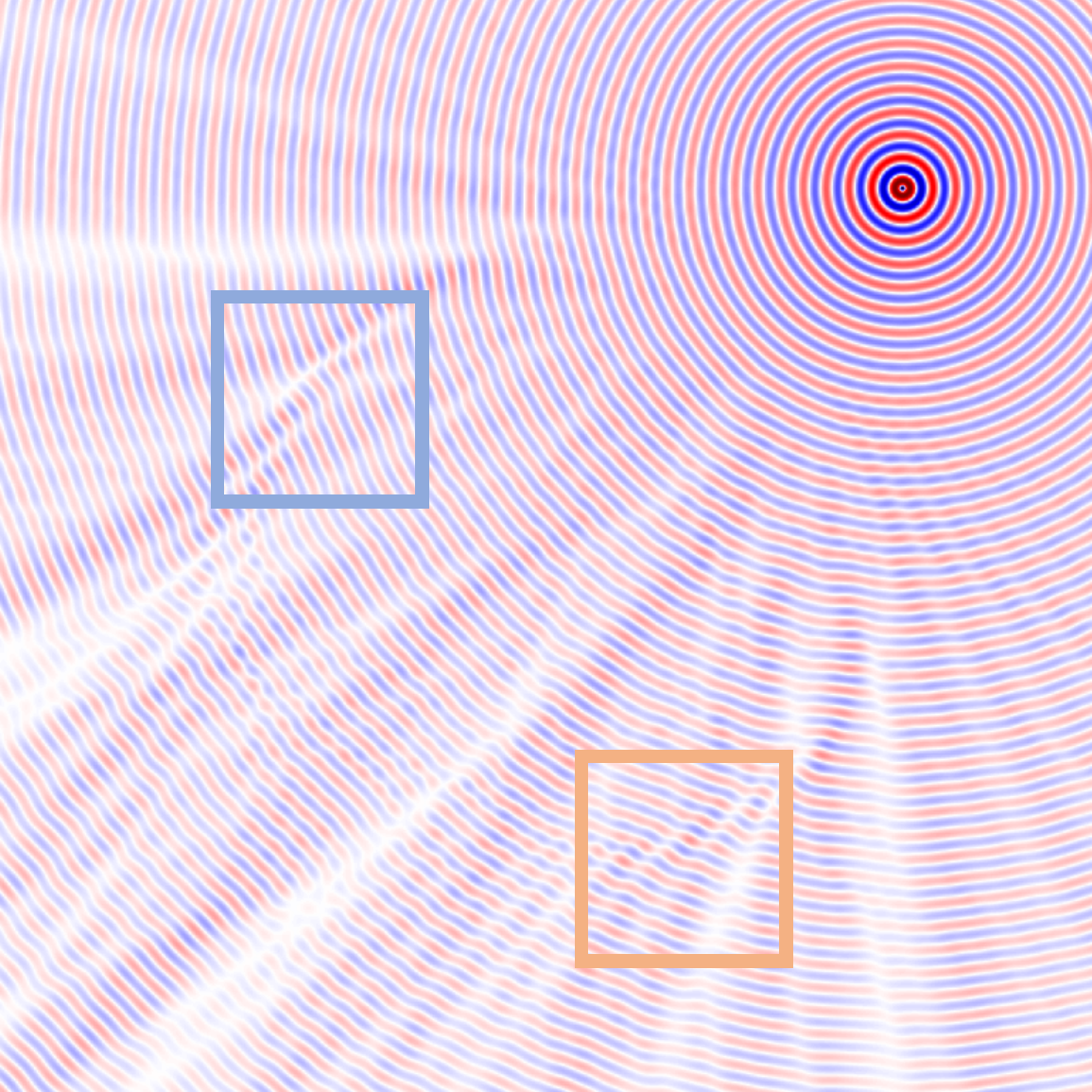} \\ \vspace{0.5mm}
        \includegraphics[width=0.48\linewidth]{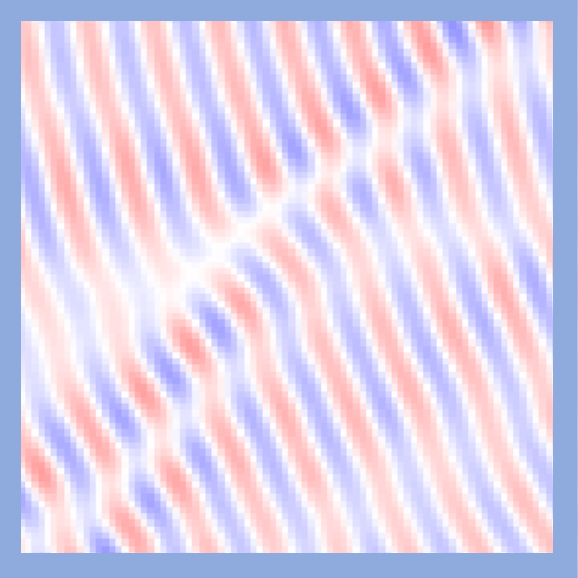}\hfill%
        \includegraphics[width=0.48\linewidth]{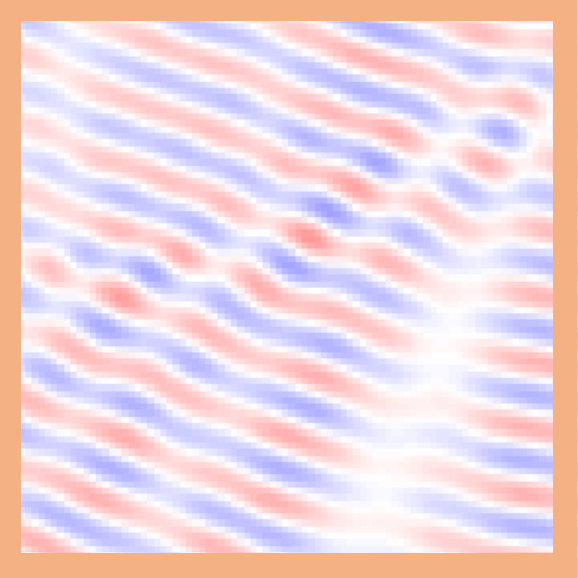}
    \end{minipage}%
    \begin{minipage}[b]{0.12\textwidth}
        \centering
        AFNO \\ \vspace{1mm} % Caption text
        \includegraphics[width=\linewidth]{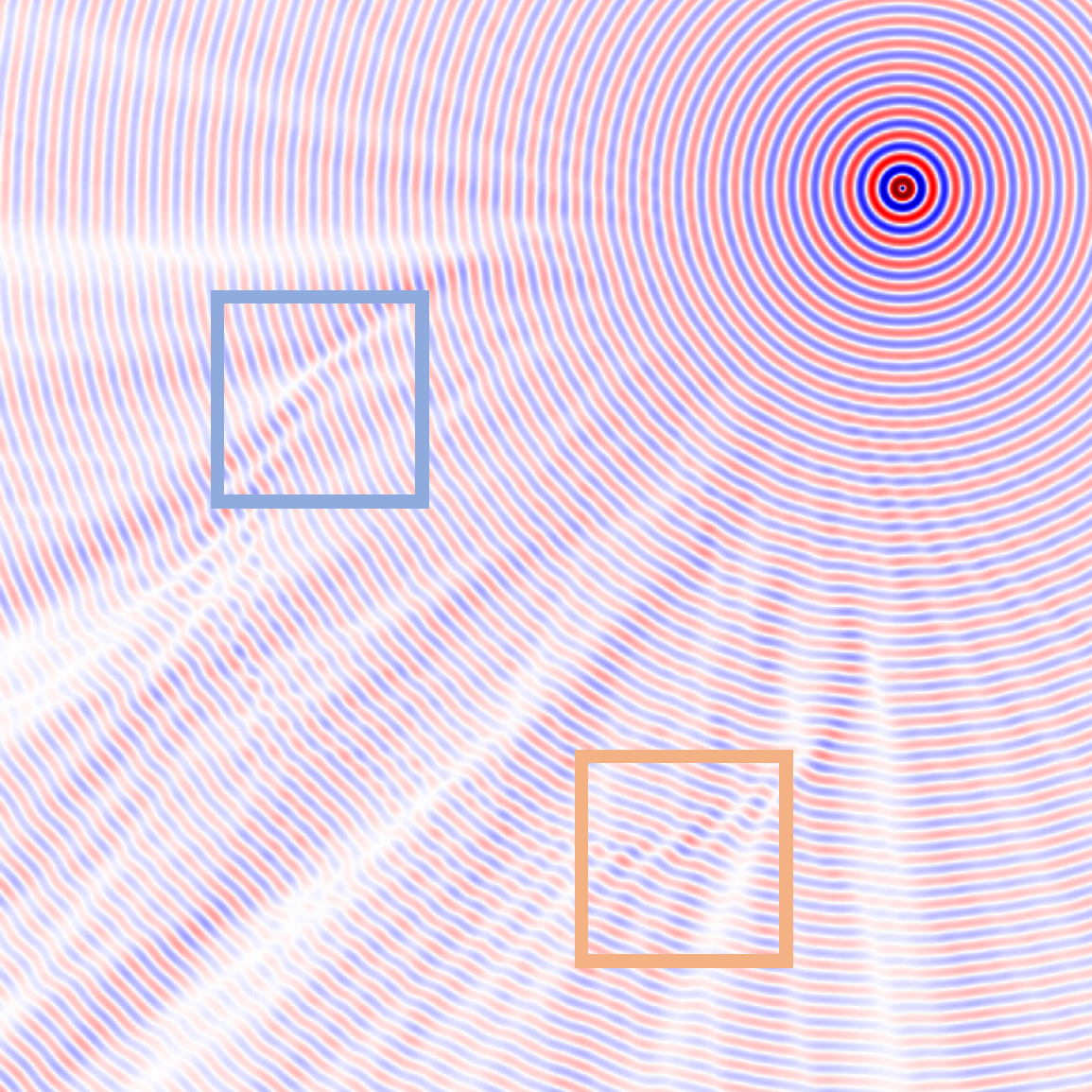} \\ \vspace{0.5mm}
        \includegraphics[width=0.48\linewidth]{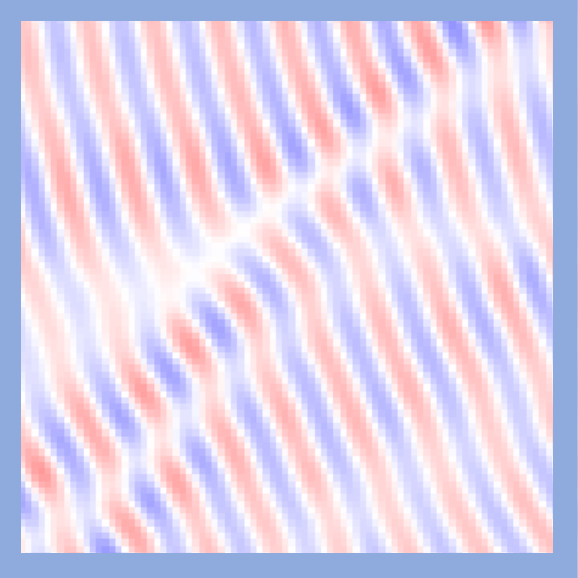}\hfill%
        \includegraphics[width=0.48\linewidth]{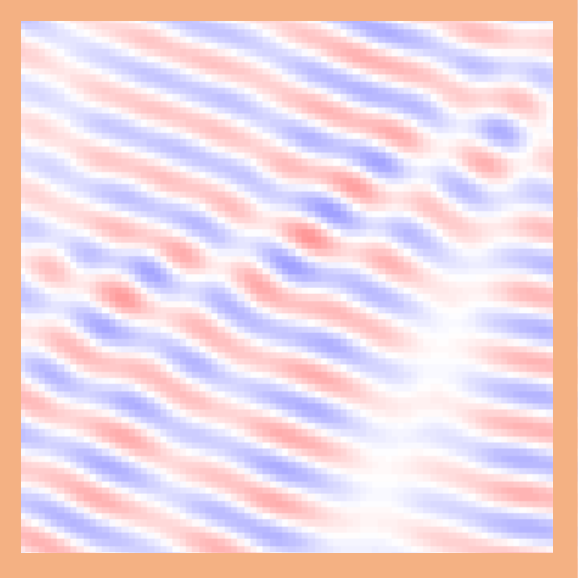}
    \end{minipage}%
    \begin{minipage}[b]{0.12\textwidth}
        \centering
        BFNO \\ \vspace{1mm} % Caption text
        \includegraphics[width=\linewidth]{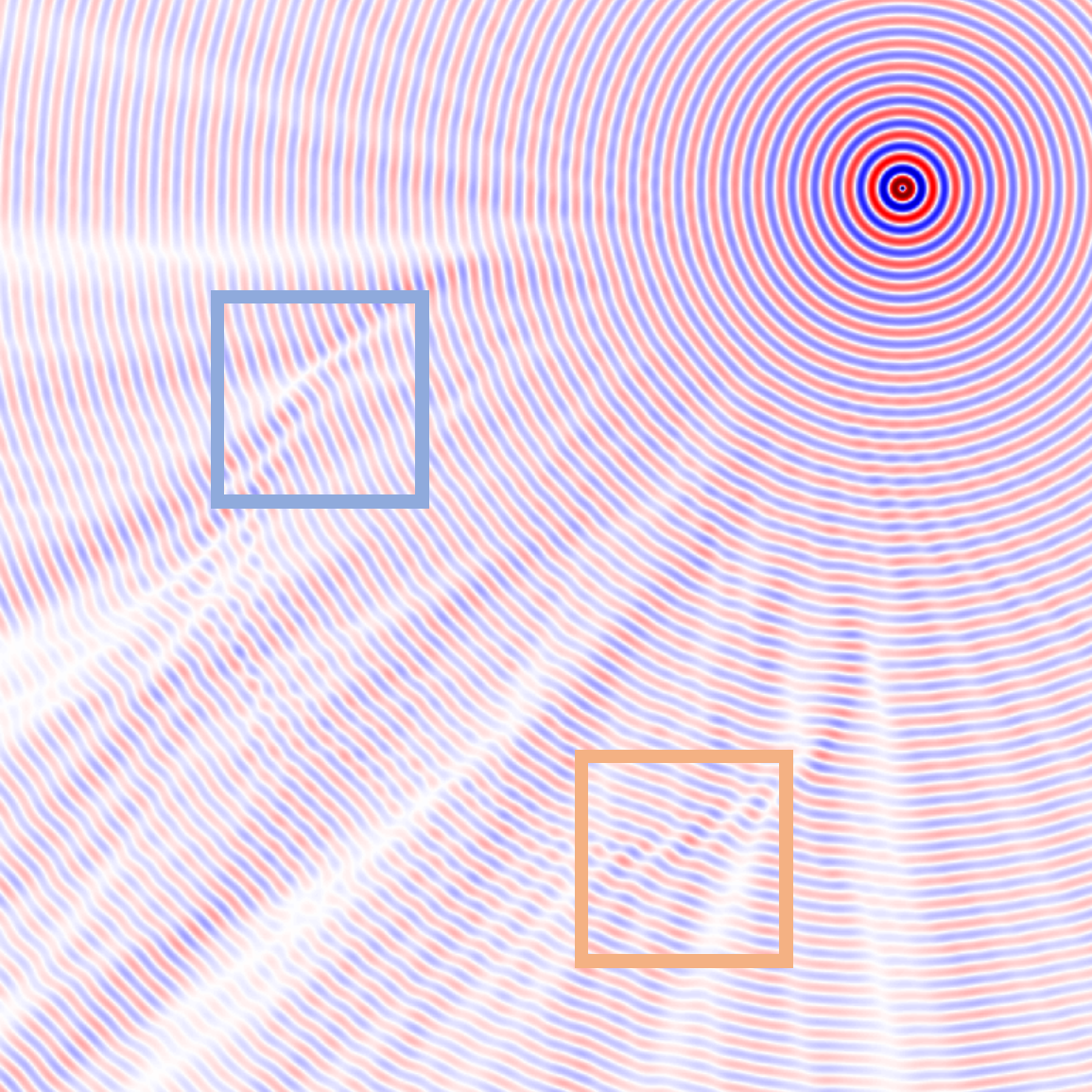} \\ \vspace{0.5mm}
        \includegraphics[width=0.48\linewidth]{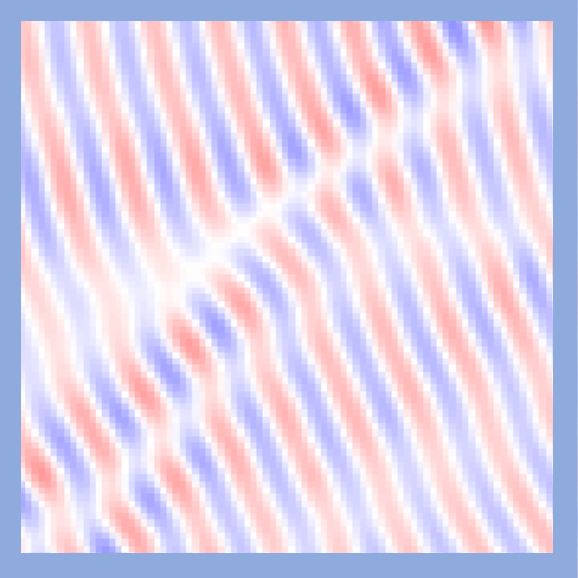}\hfill%
        \includegraphics[width=0.48\linewidth]{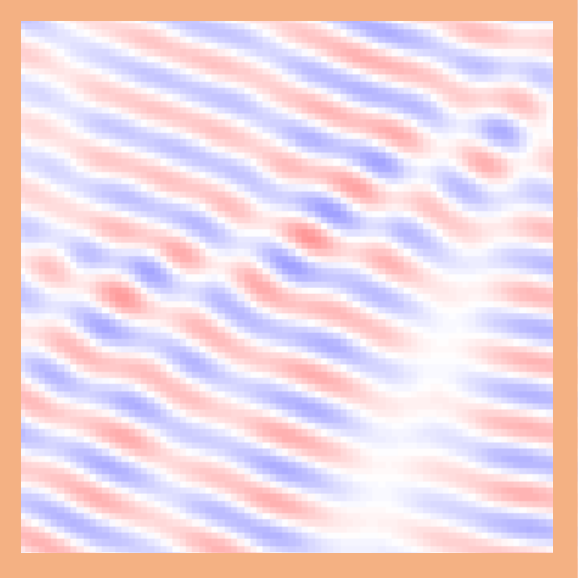}
    \end{minipage}%
    \begin{minipage}[b]{0.12\textwidth}
        \centering
        MgNO \\ \vspace{1mm} % Caption text
        \includegraphics[width=\linewidth]{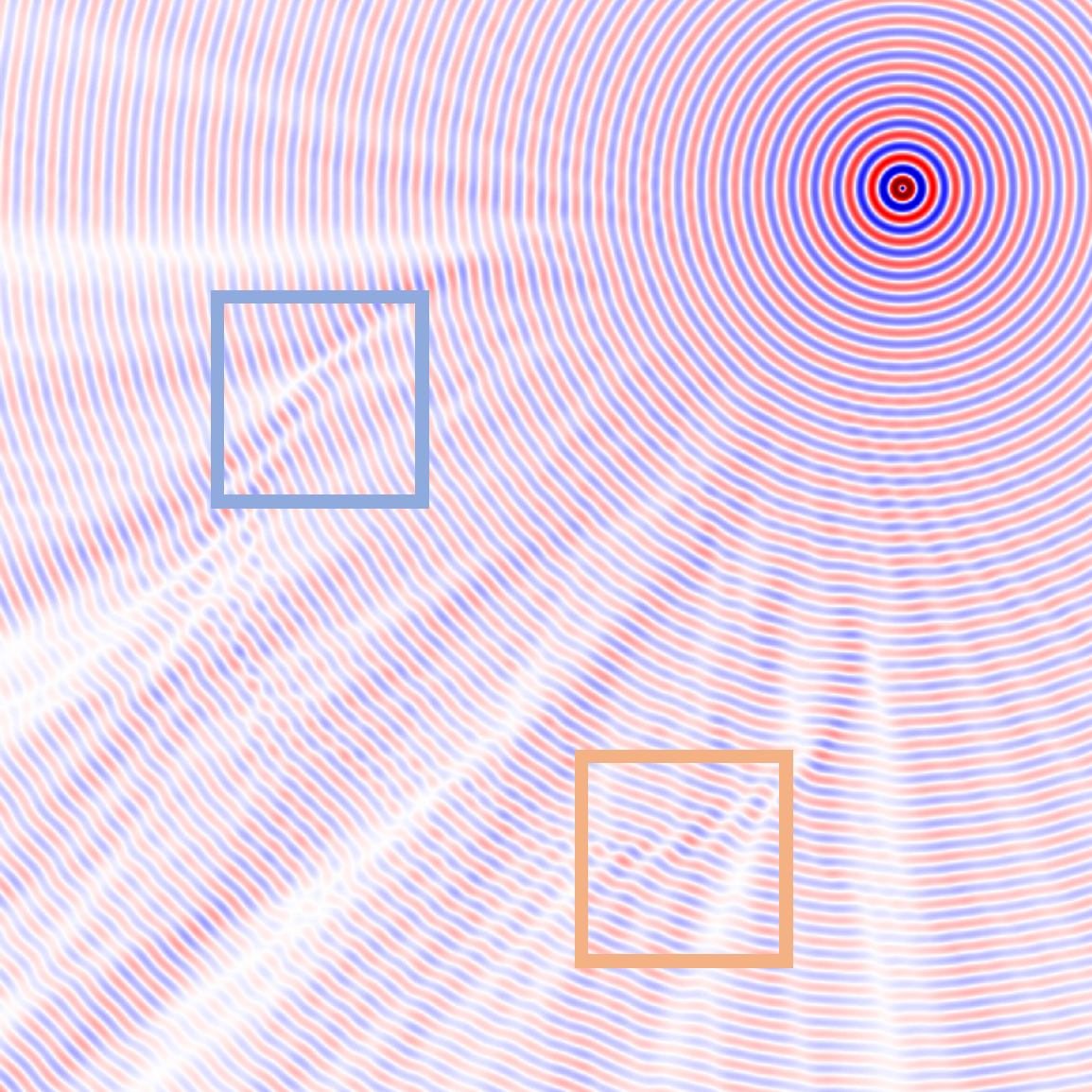} \\ \vspace{0.5mm}
        \includegraphics[width=0.48\linewidth]{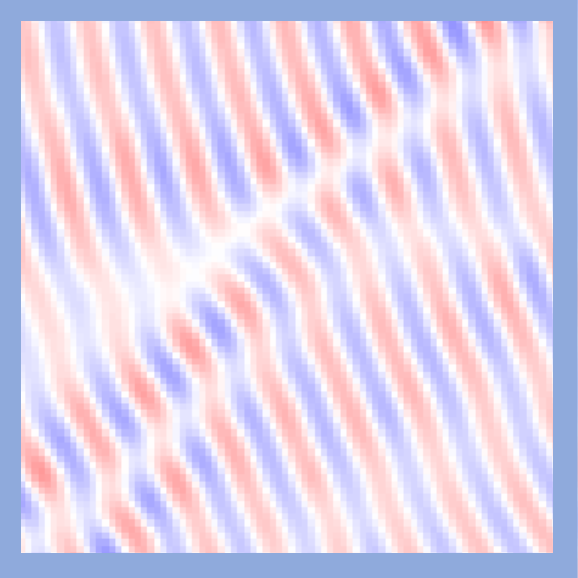}\hfill%
        \includegraphics[width=0.48\linewidth]{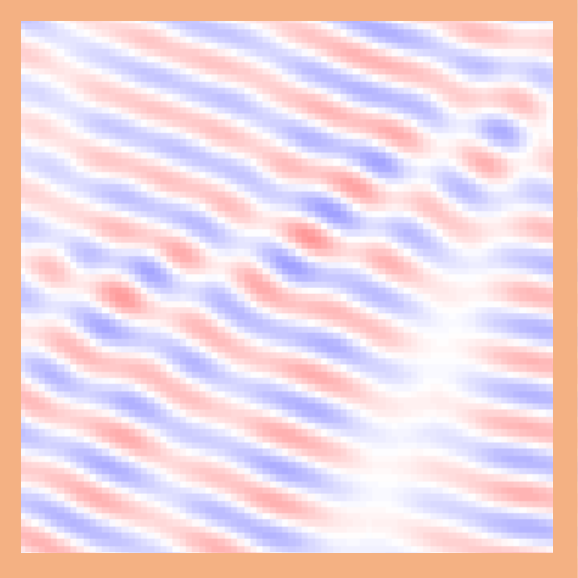}
    \end{minipage}%
    \begin{minipage}[b]{0.07\textwidth} % Spacer
        \hspace{0pt} % Ensures it takes up space
    \end{minipage}
    \\ \vspace{2mm} % Vertical space between rows

    % ROW 2: FIB
    \noindent
    \begin{minipage}[b]{0.05\textwidth}
        \centering
        \rotatebox{90}{\quad\quad\quad FIB}
    \end{minipage}%
    \begin{minipage}[b]{0.1285\textwidth}
        \centering
        \includegraphics[width=\linewidth]{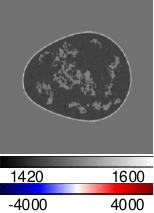}
    \end{minipage}%
    \begin{minipage}[b]{0.12\textwidth}
        \centering
        \includegraphics[width=\linewidth]{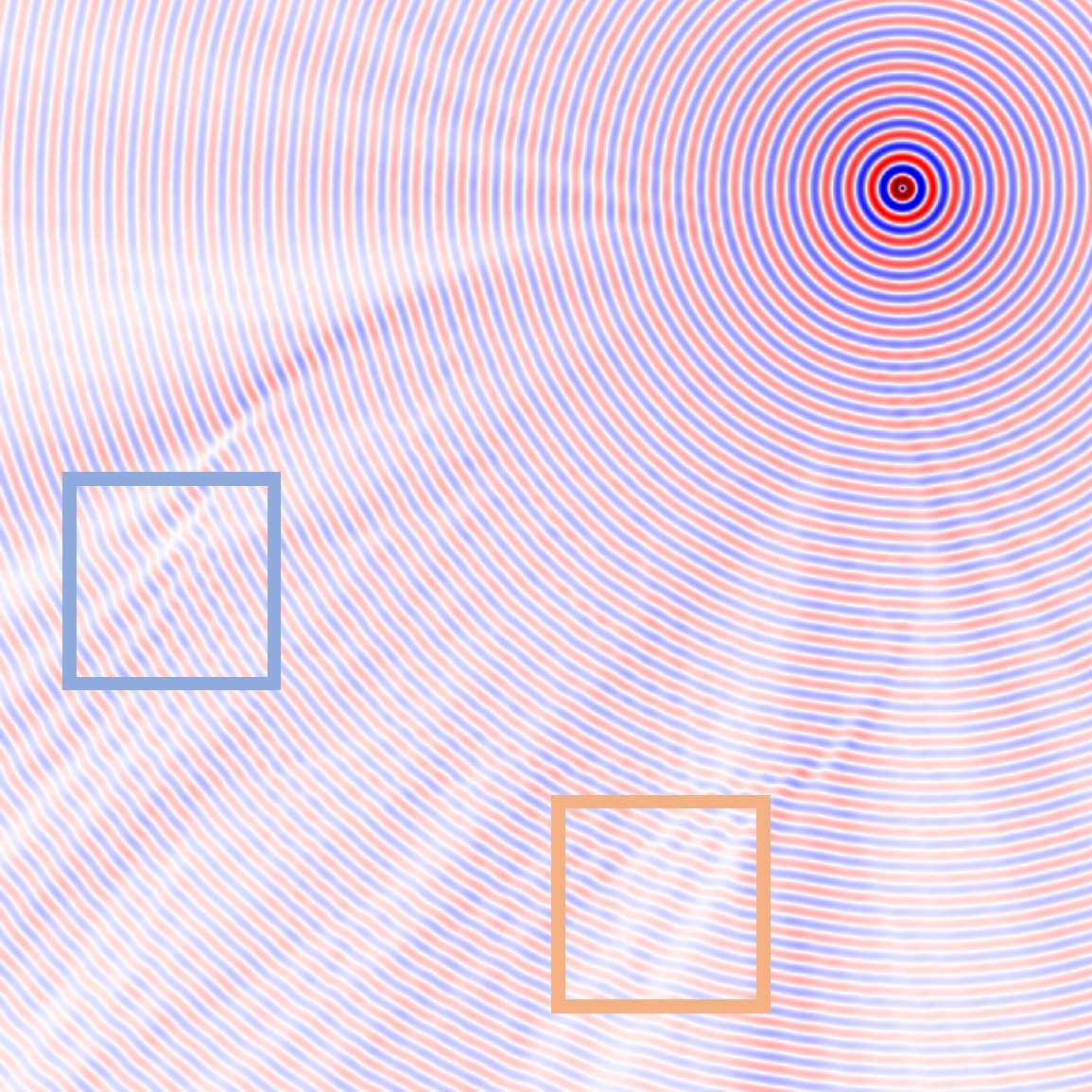} \\ \vspace{0.5mm}
        \includegraphics[width=0.48\linewidth]{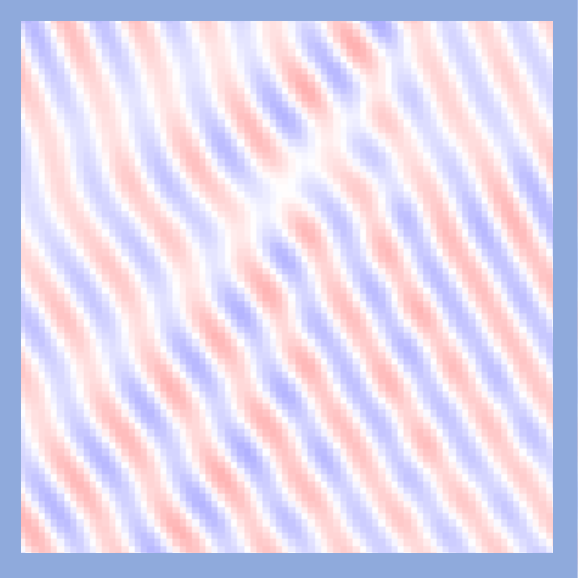}\hfill%
        \includegraphics[width=0.48\linewidth]{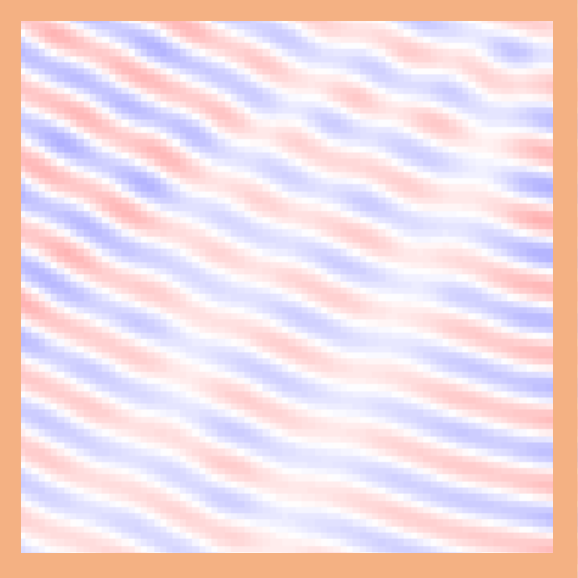}
    \end{minipage}%
    \begin{minipage}[b]{0.12\textwidth}
        \centering
        \includegraphics[width=\linewidth]{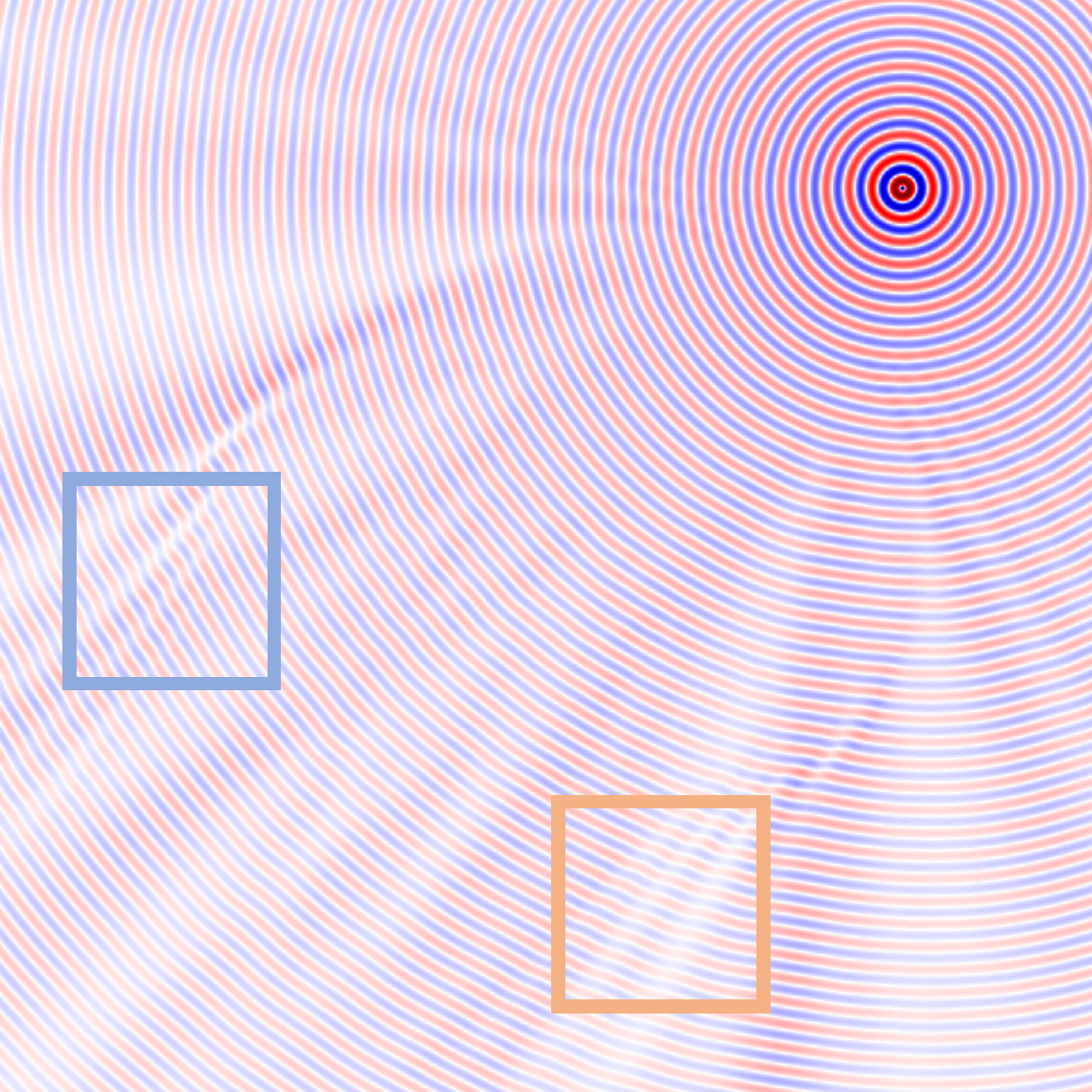} \\ \vspace{0.5mm}
        \includegraphics[width=0.48\linewidth]{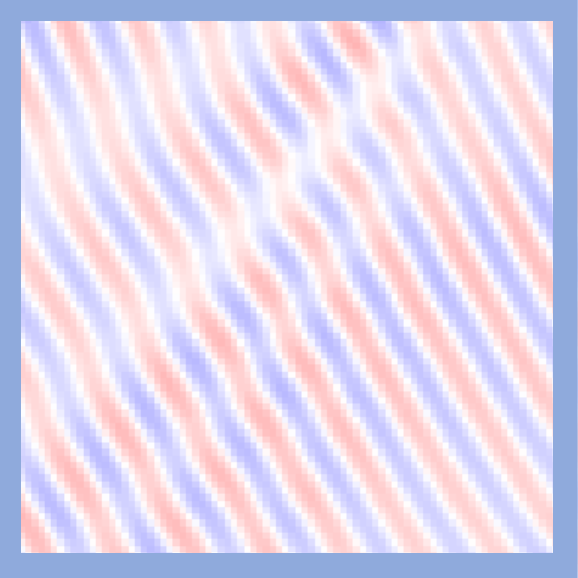}\hfill%
        \includegraphics[width=0.48\linewidth]{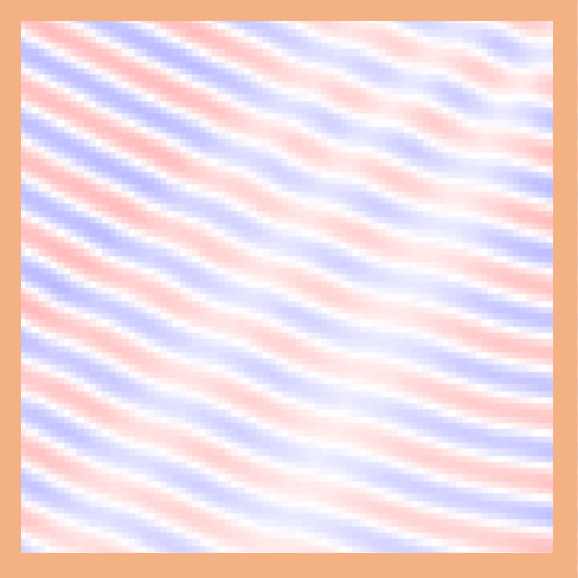}
    \end{minipage}%
    \begin{minipage}[b]{0.12\textwidth}
        \centering
        \includegraphics[width=\linewidth]{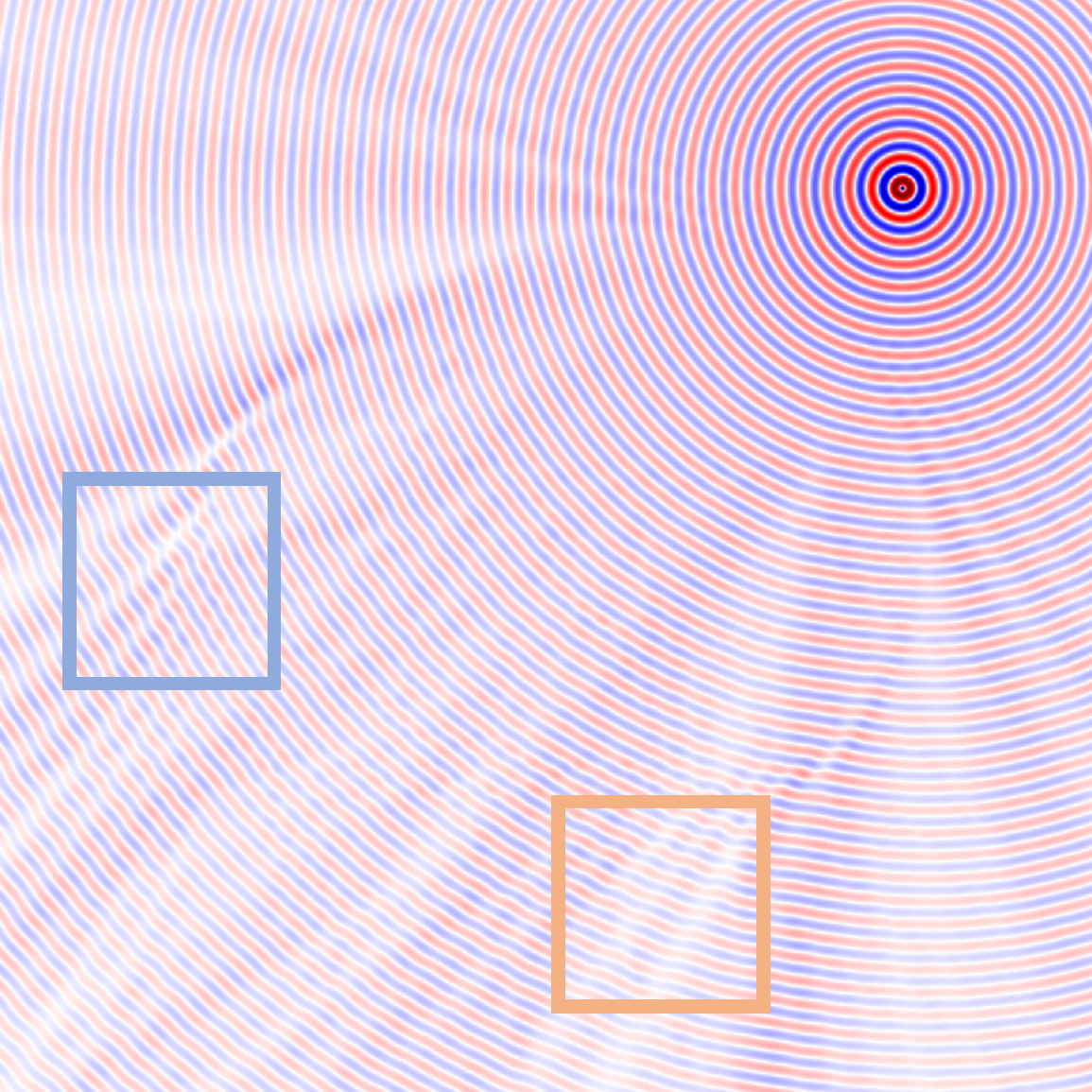} \\ \vspace{0.5mm}
        \includegraphics[width=0.48\linewidth]{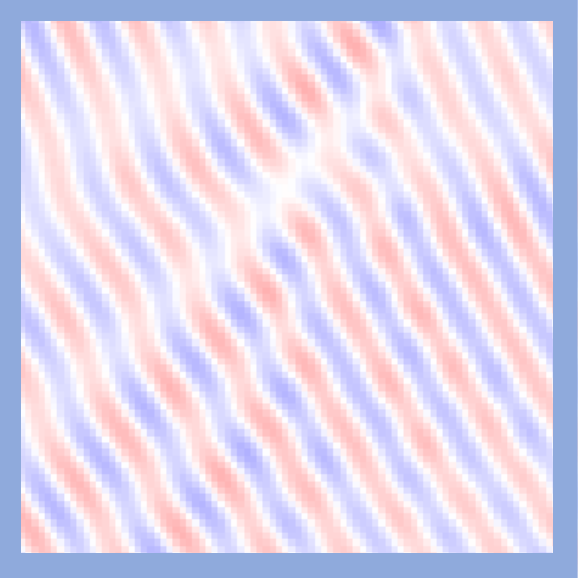}\hfill%
        \includegraphics[width=0.48\linewidth]{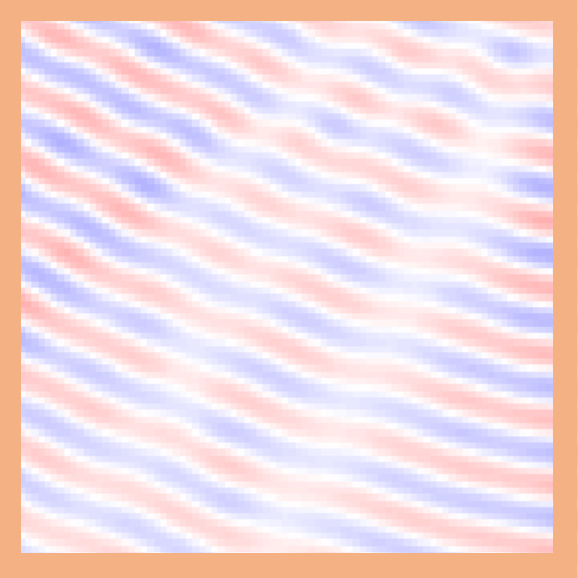}
    \end{minipage}%
    \begin{minipage}[b]{0.12\textwidth}
        \centering
        \includegraphics[width=\linewidth]{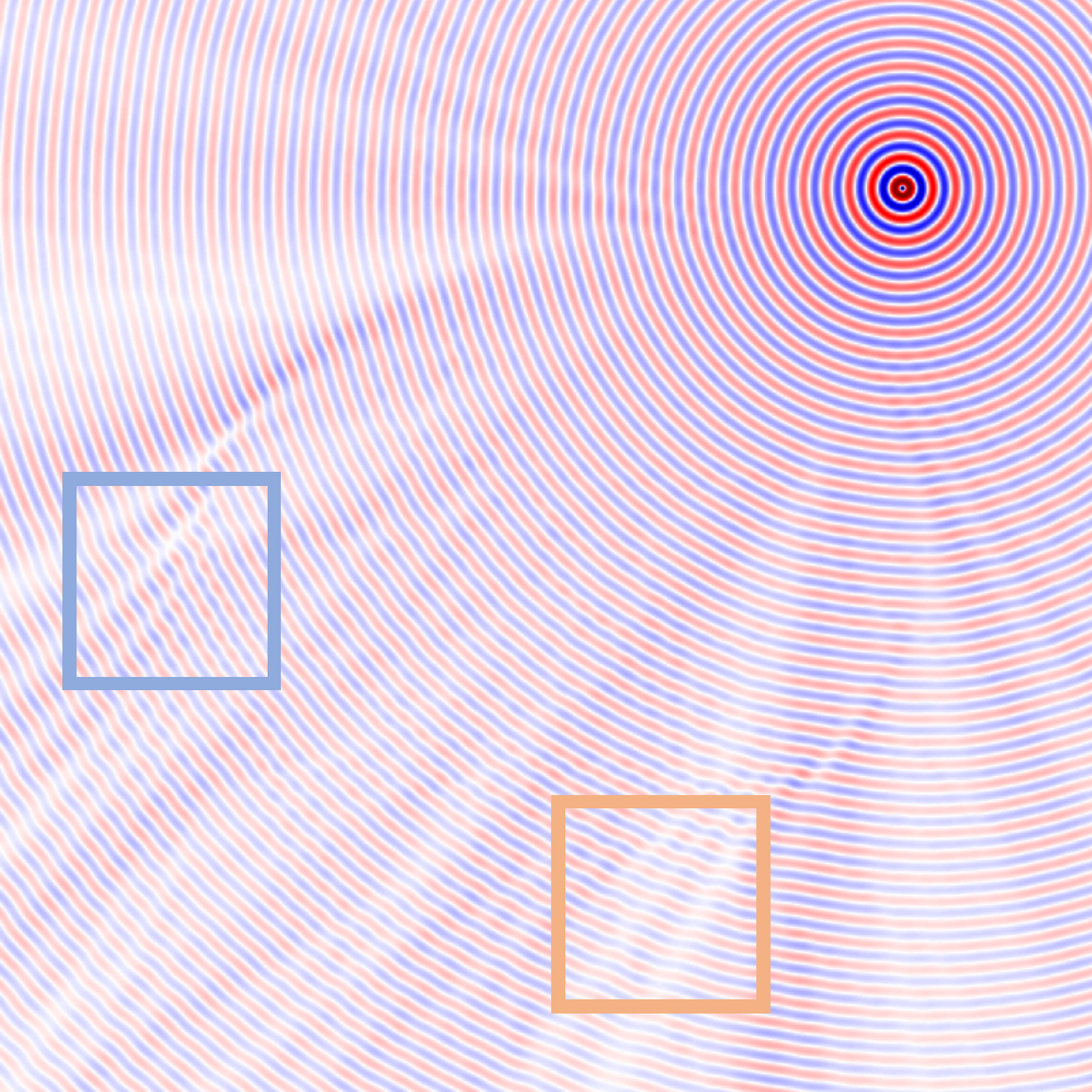} \\ \vspace{0.5mm}
        \includegraphics[width=0.48\linewidth]{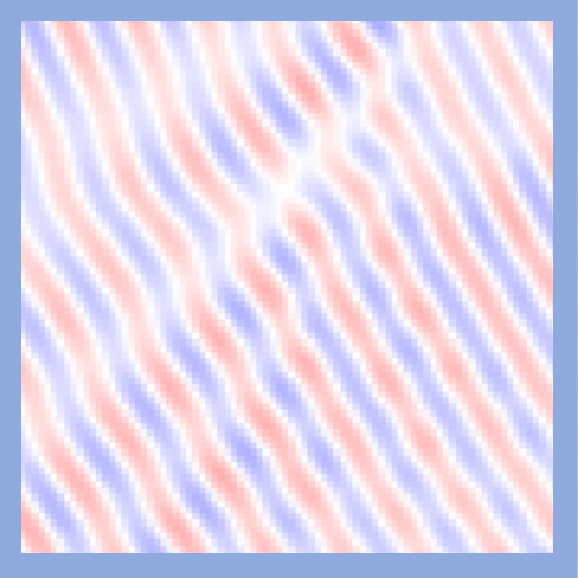}\hfill%
        \includegraphics[width=0.48\linewidth]{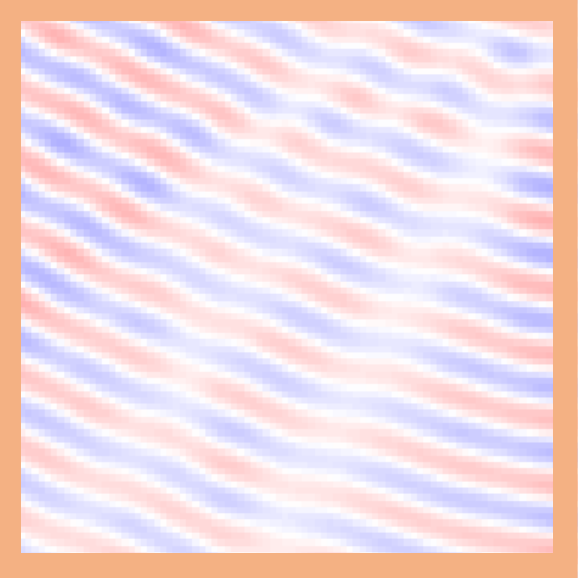}
    \end{minipage}%
    \begin{minipage}[b]{0.12\textwidth}
        \centering
        \includegraphics[width=\linewidth]{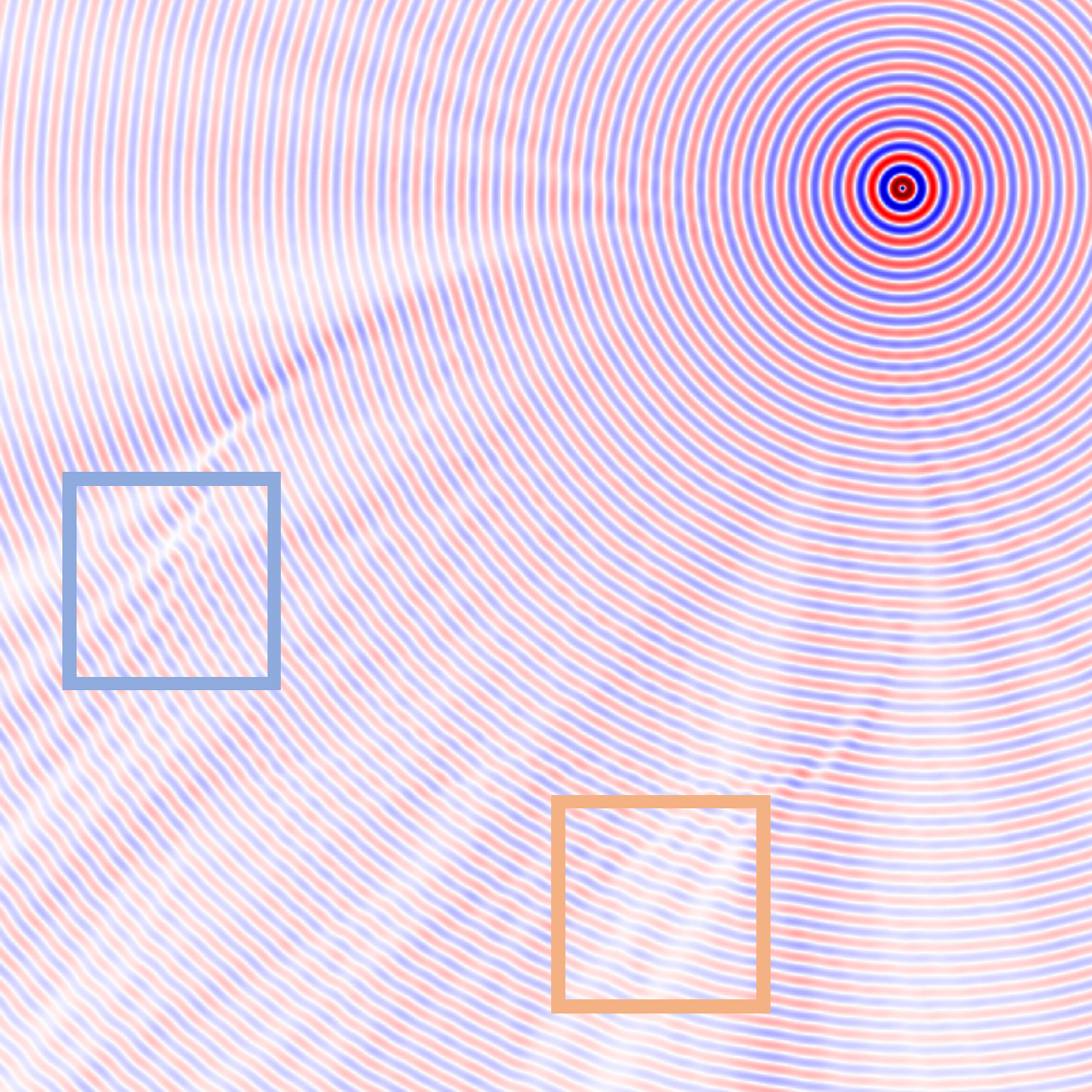} \\ \vspace{0.5mm}
        \includegraphics[width=0.48\linewidth]{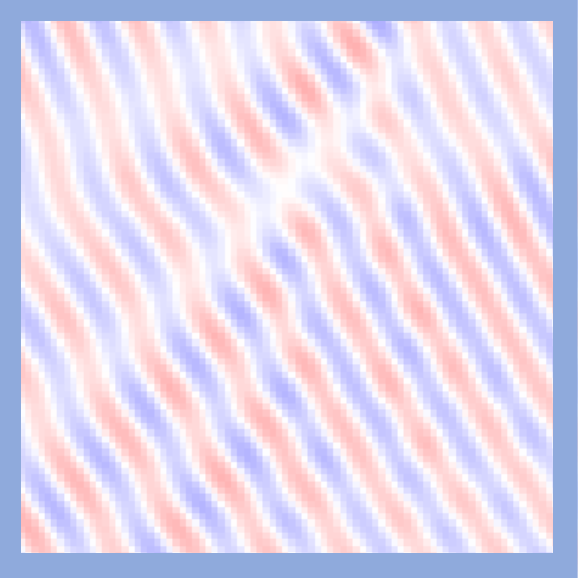}\hfill%
        \includegraphics[width=0.48\linewidth]{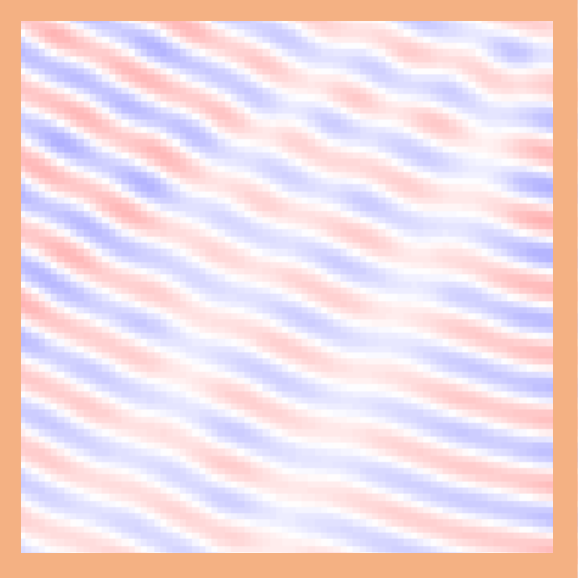}
    \end{minipage}%
    \begin{minipage}[b]{0.12\textwidth}
        \centering
        \includegraphics[width=\linewidth]{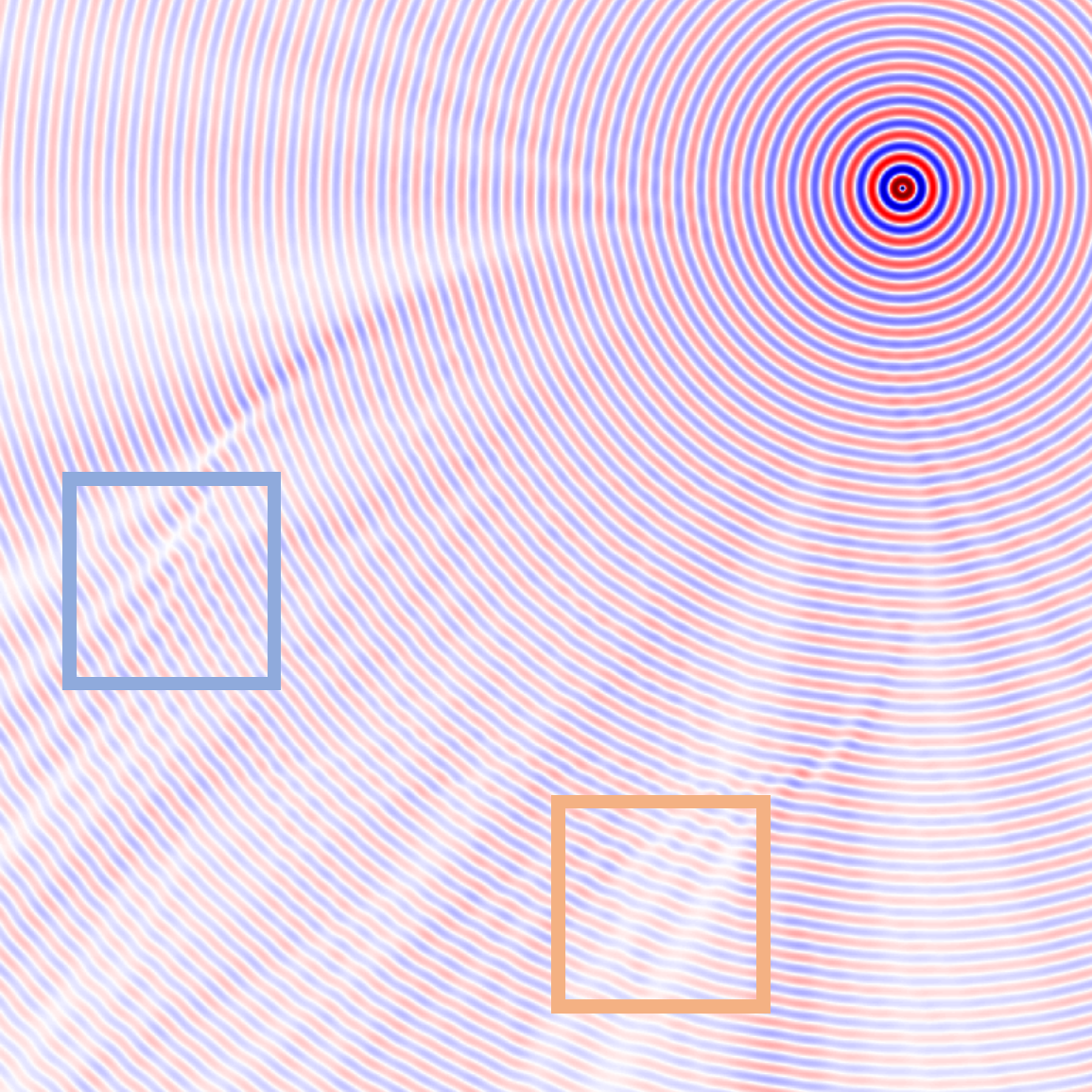} \\ \vspace{0.5mm}
        \includegraphics[width=0.48\linewidth]{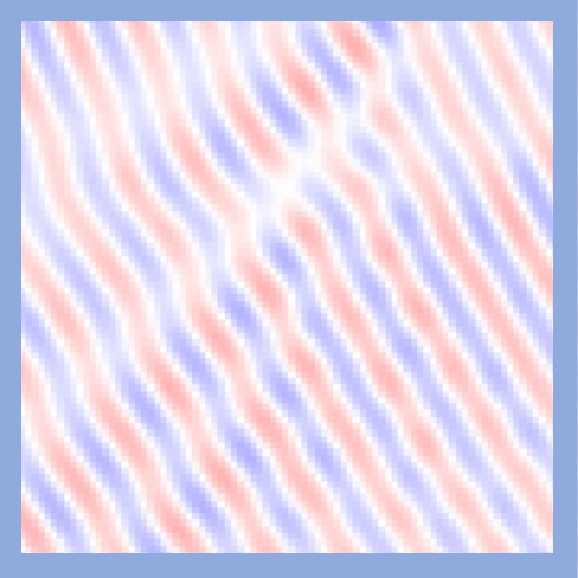}\hfill%
        \includegraphics[width=0.48\linewidth]{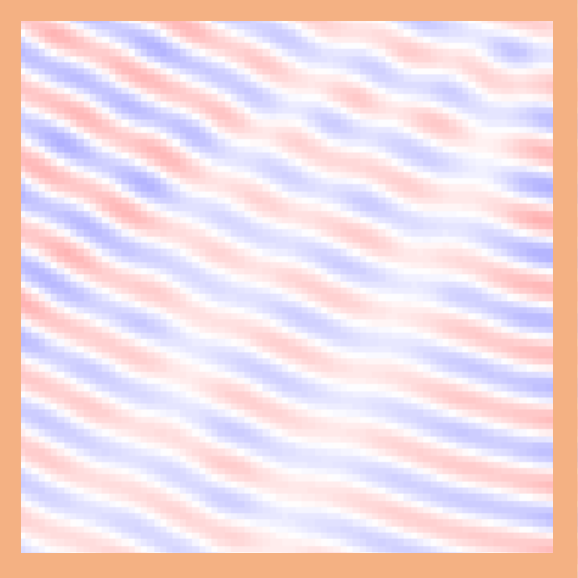}
    \end{minipage}%
    \begin{minipage}[b]{0.07\textwidth} % Spacer
        \hspace{0pt}
    \end{minipage}
    \\ \vspace{2mm}

    % ROW 3: FAT
    \noindent
    \begin{minipage}[b]{0.05\textwidth}
        \centering
        \rotatebox{90}{\quad\quad\quad FAT}
    \end{minipage}%
    \begin{minipage}[b]{0.1285\textwidth}
        \centering
        \includegraphics[width=\linewidth]{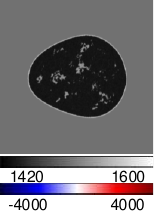}
    \end{minipage}%
    \begin{minipage}[b]{0.12\textwidth}
        \centering
        \includegraphics[width=\linewidth]{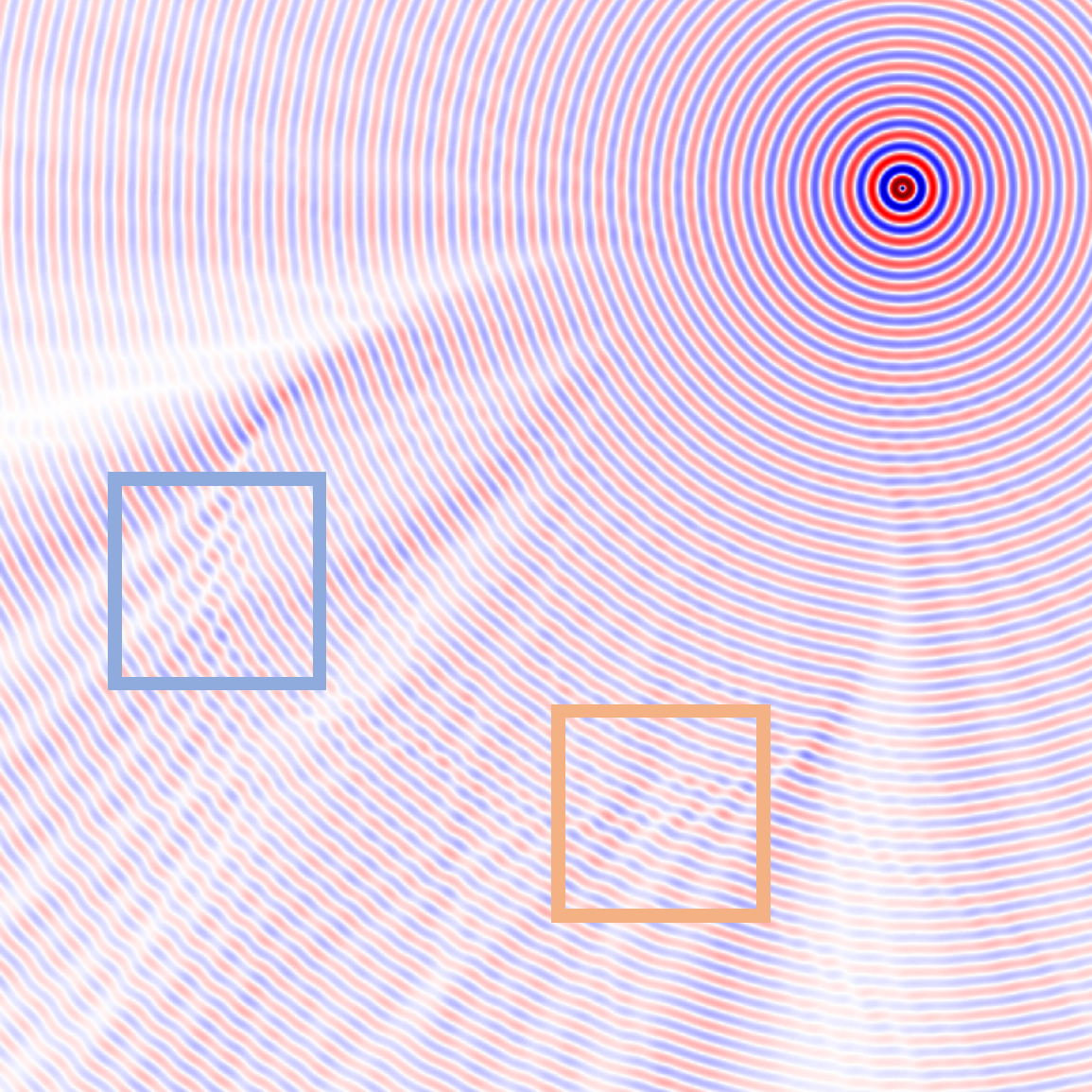} \\ \vspace{0.5mm}
        \includegraphics[width=0.48\linewidth]{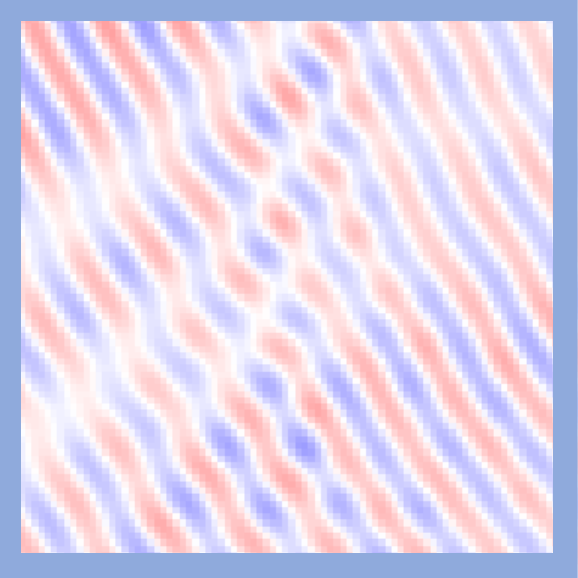}\hfill%
        \includegraphics[width=0.48\linewidth]{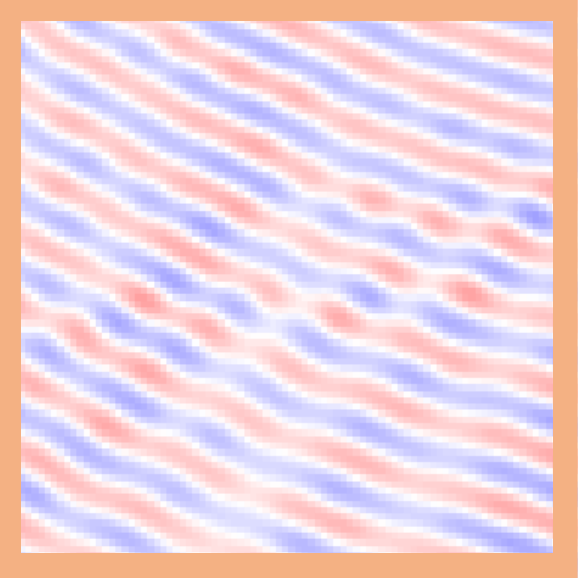}
    \end{minipage}%
    \begin{minipage}[b]{0.12\textwidth}
        \centering
        \includegraphics[width=\linewidth]{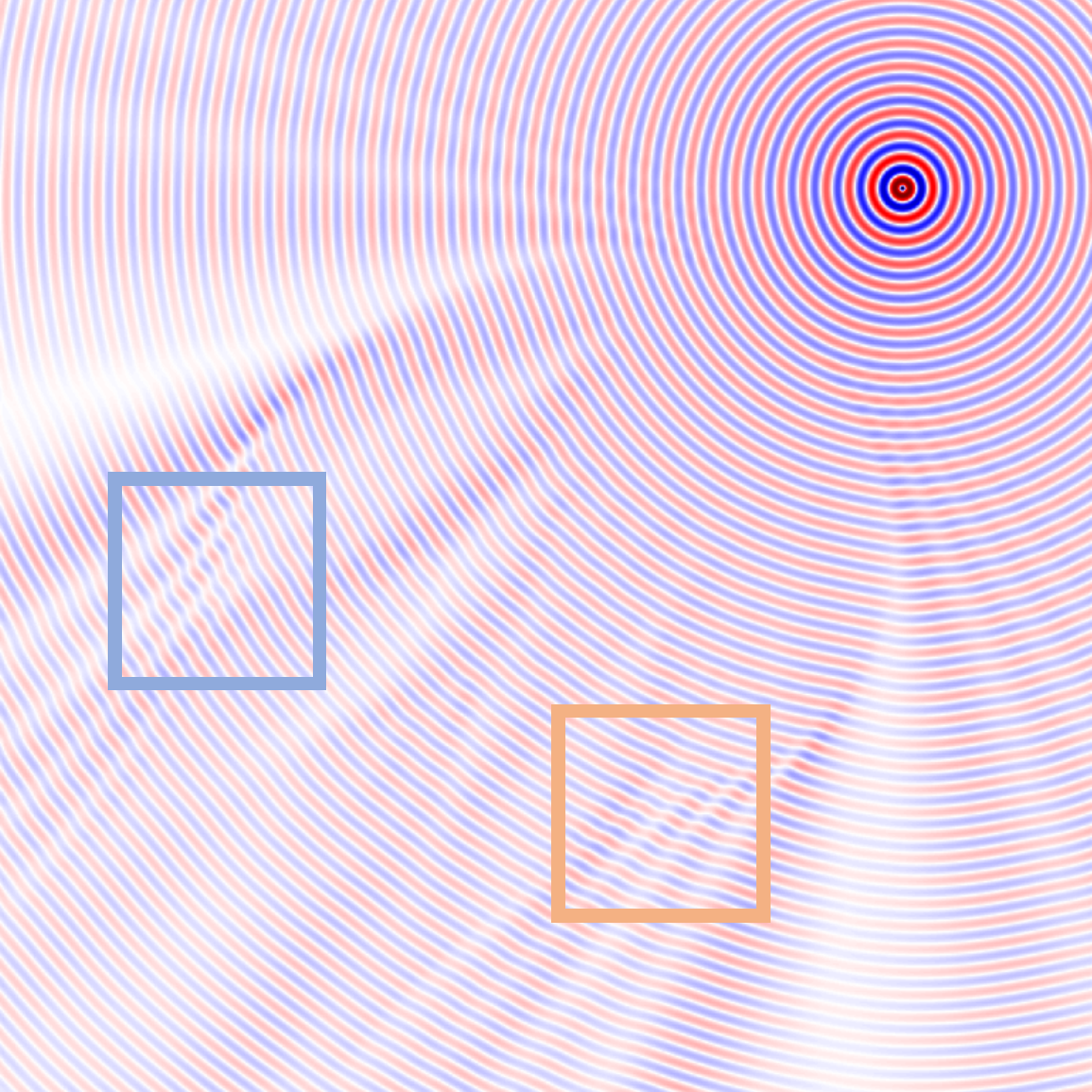} \\ \vspace{0.5mm}
        \includegraphics[width=0.48\linewidth]{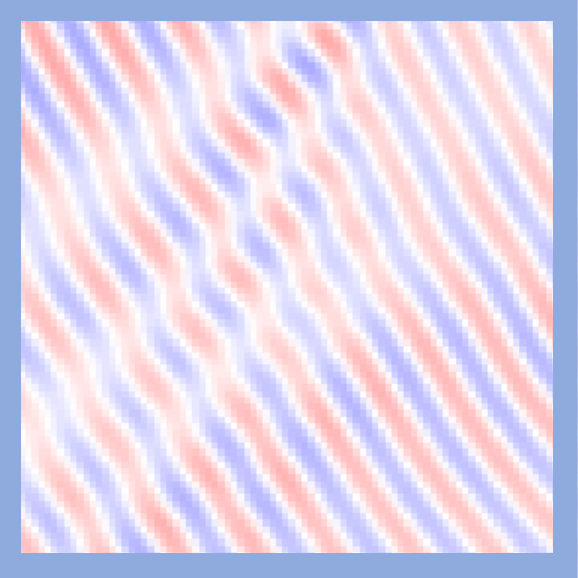}\hfill%
        \includegraphics[width=0.48\linewidth]{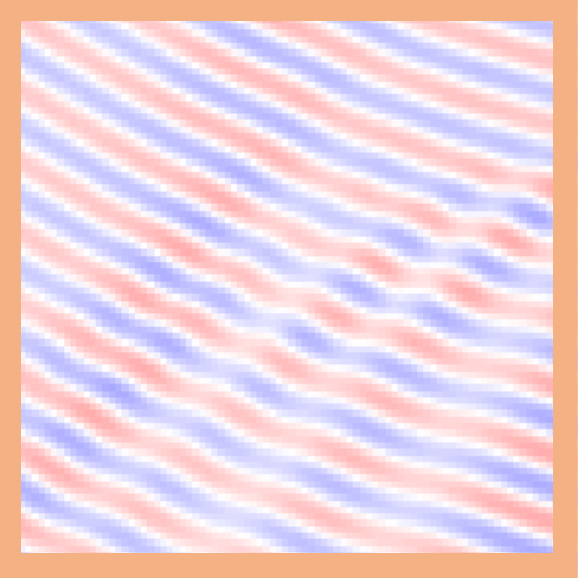}
    \end{minipage}%
    \begin{minipage}[b]{0.12\textwidth}
        \centering
        \includegraphics[width=\linewidth]{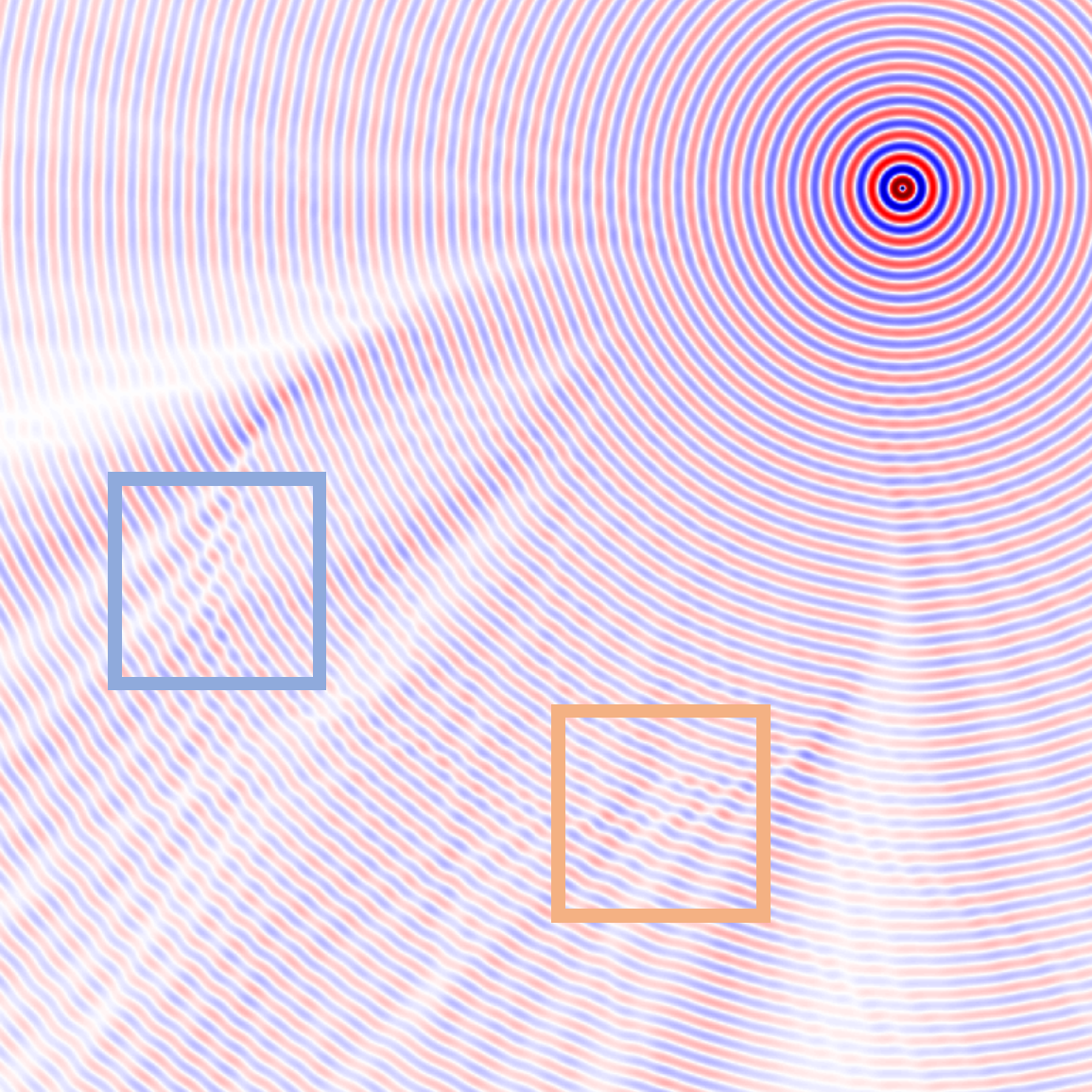} \\ \vspace{0.5mm}
        \includegraphics[width=0.48\linewidth]{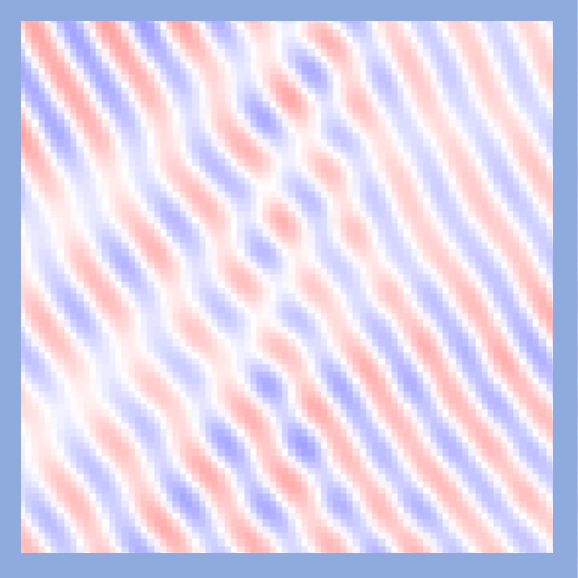}\hfill%
        \includegraphics[width=0.48\linewidth]{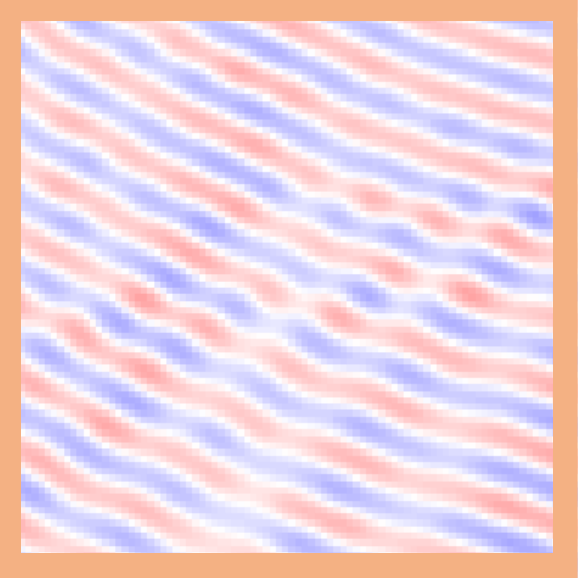}
    \end{minipage}%
    \begin{minipage}[b]{0.12\textwidth}
        \centering
        \includegraphics[width=\linewidth]{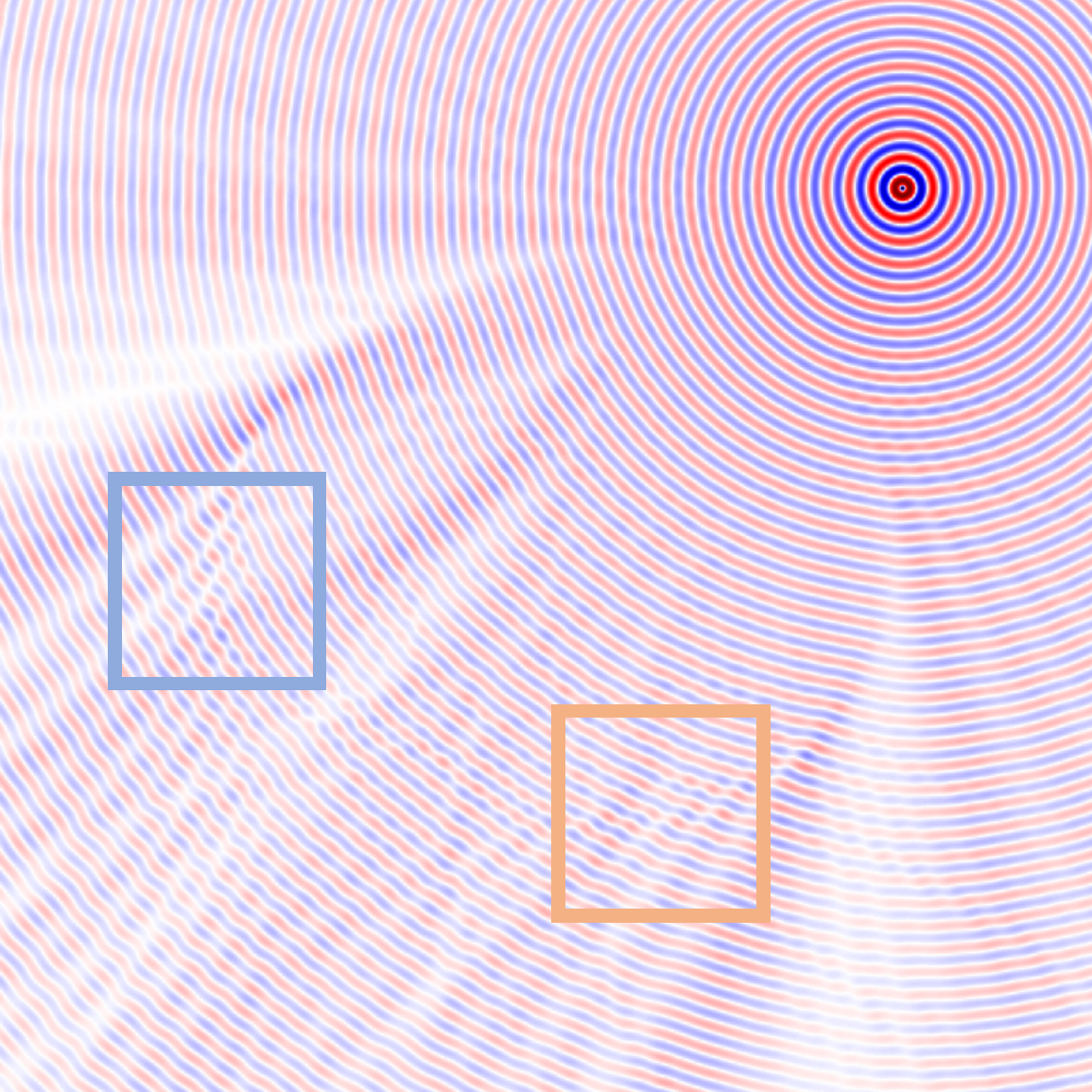} \\ \vspace{0.5mm}
        \includegraphics[width=0.48\linewidth]{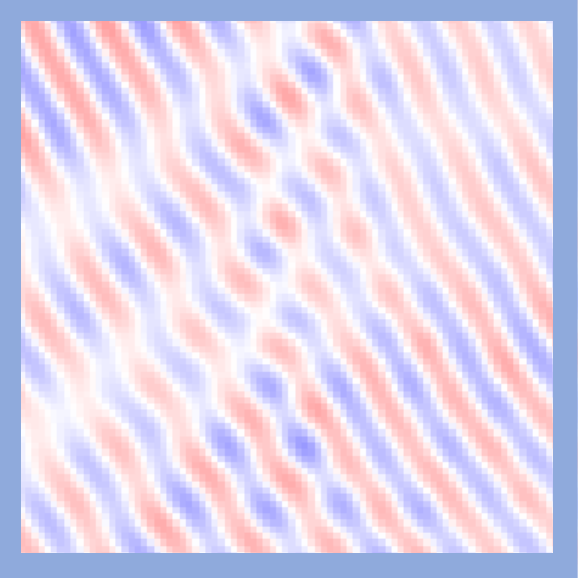}\hfill%
        \includegraphics[width=0.48\linewidth]{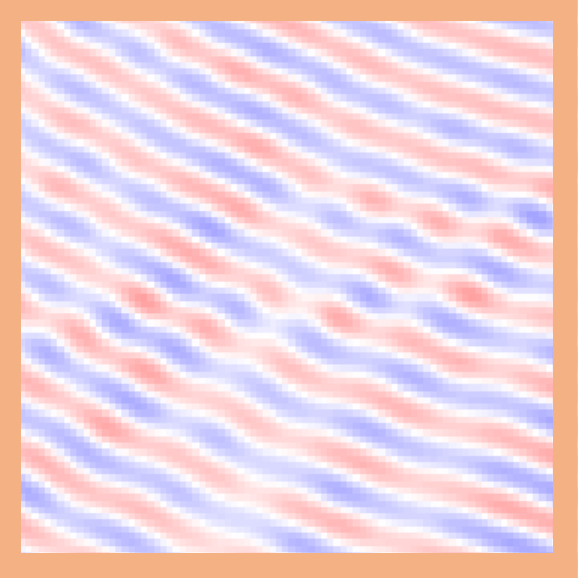}
    \end{minipage}%
    \begin{minipage}[b]{0.12\textwidth}
        \centering
        \includegraphics[width=\linewidth]{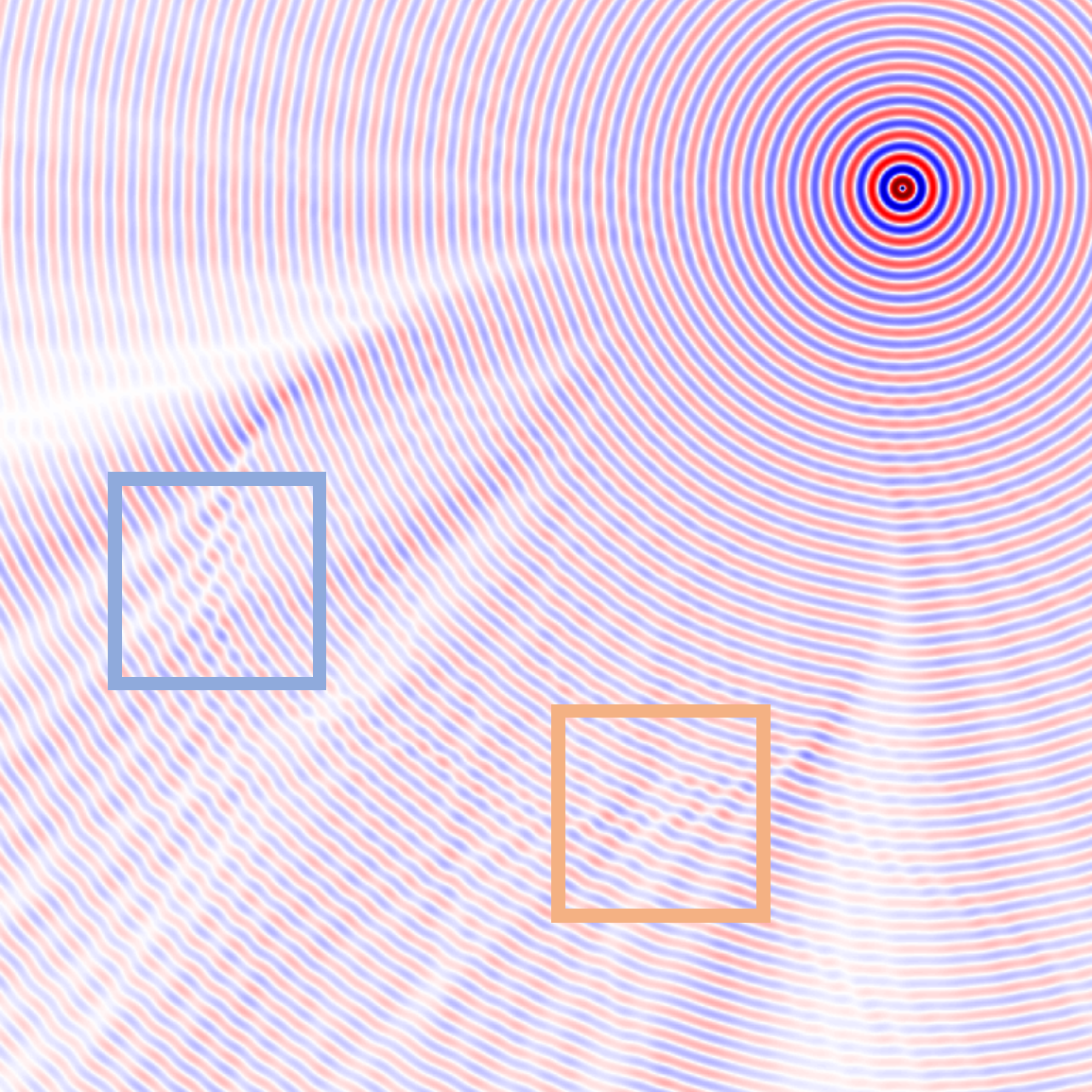} \\ \vspace{0.5mm}
        \includegraphics[width=0.48\linewidth]{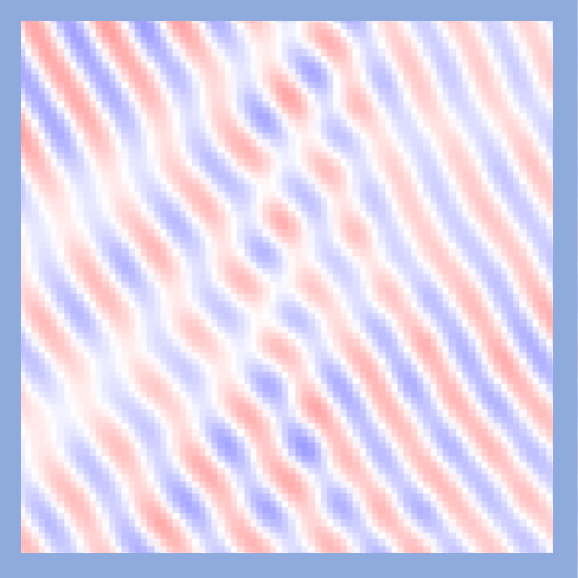}\hfill%
        \includegraphics[width=0.48\linewidth]{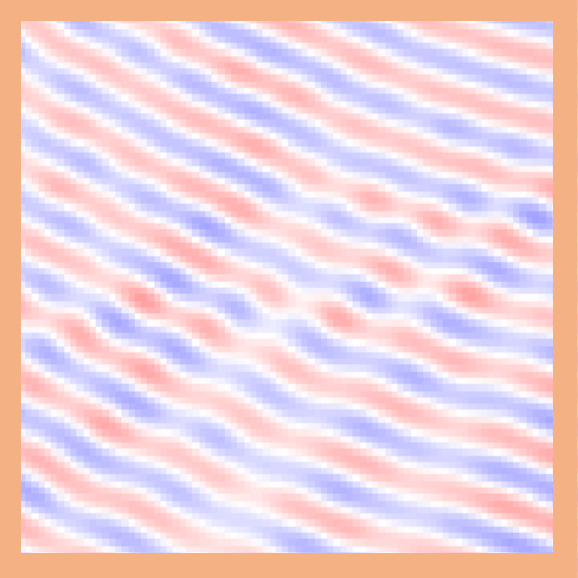}
    \end{minipage}%
    \begin{minipage}[b]{0.12\textwidth}
        \centering
        \includegraphics[width=\linewidth]{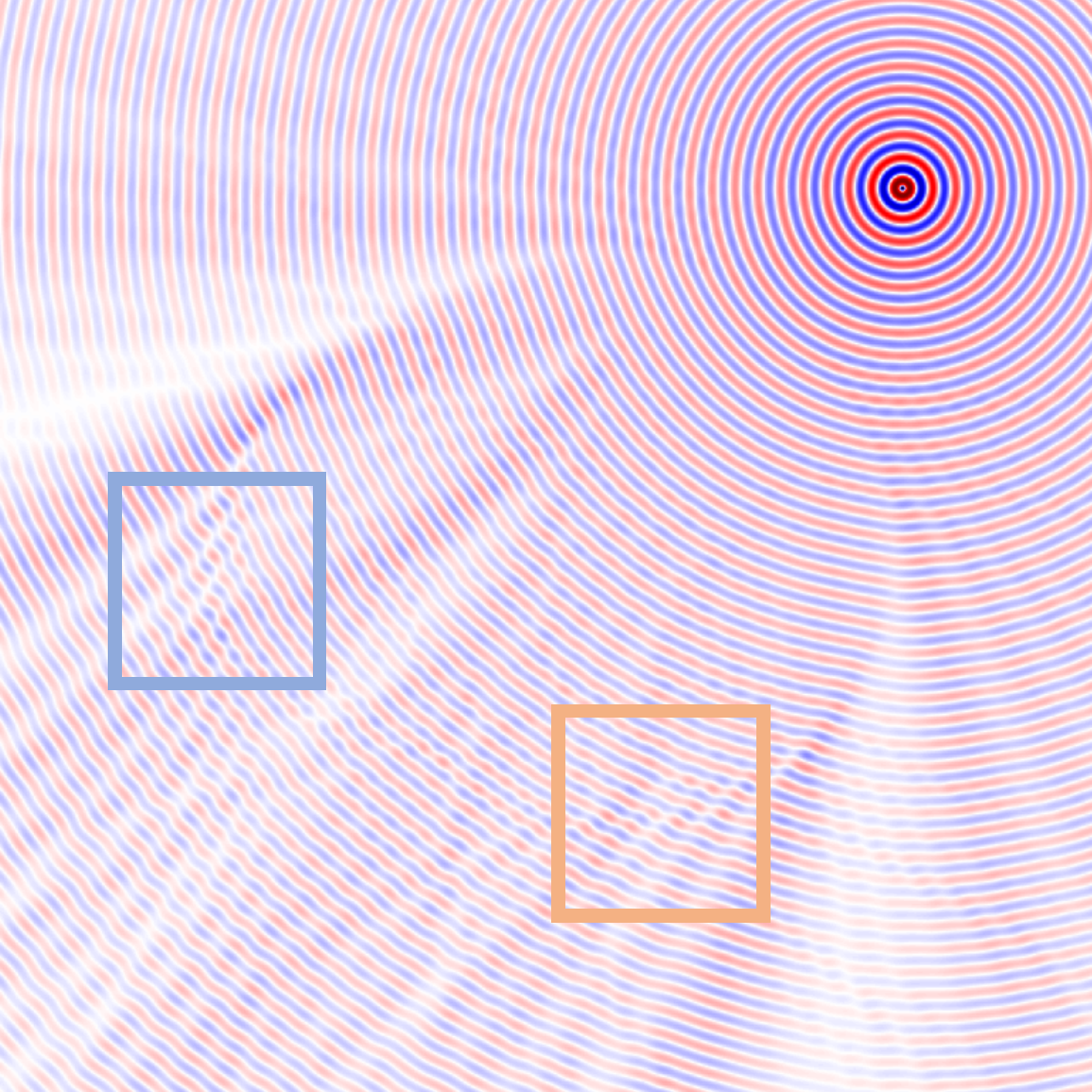} \\ \vspace{0.5mm}
        \includegraphics[width=0.48\linewidth]{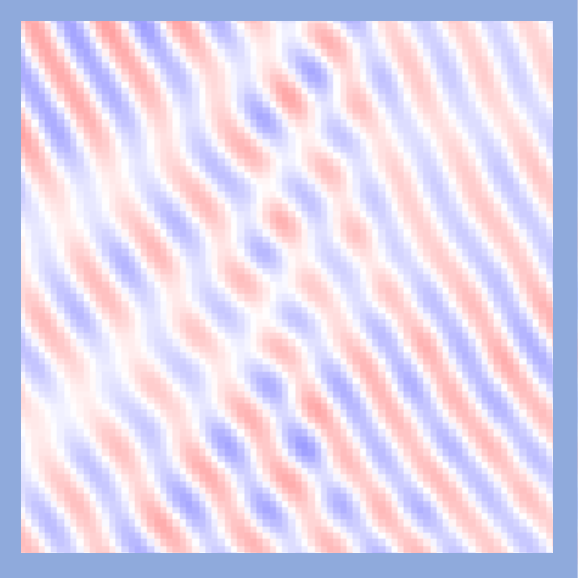}\hfill%
        \includegraphics[width=0.48\linewidth]{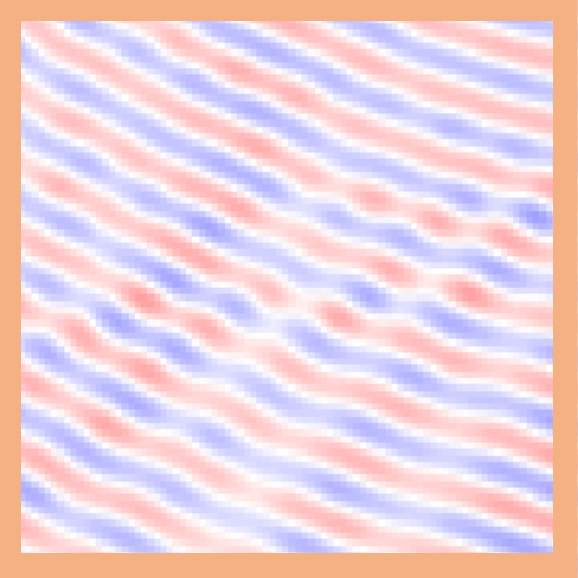}
    \end{minipage}%
    \begin{minipage}[b]{0.07\textwidth} % Spacer
        \hspace{0pt}
    \end{minipage}
    \\ \vspace{2mm}

    % ROW 4: EXD
    \noindent
    \begin{minipage}[b]{0.05\textwidth}
        \centering
        \rotatebox{90}{\quad\quad \quad EXD} % Adjusted spacing slightly to match original if needed
    \end{minipage}%
    \begin{minipage}[b]{0.1285\textwidth}
        \centering
        \includegraphics[width=\linewidth]{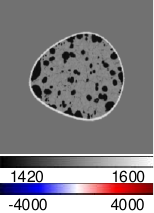}
    \end{minipage}%
    \begin{minipage}[b]{0.12\textwidth}
        \centering
        \includegraphics[width=\linewidth]{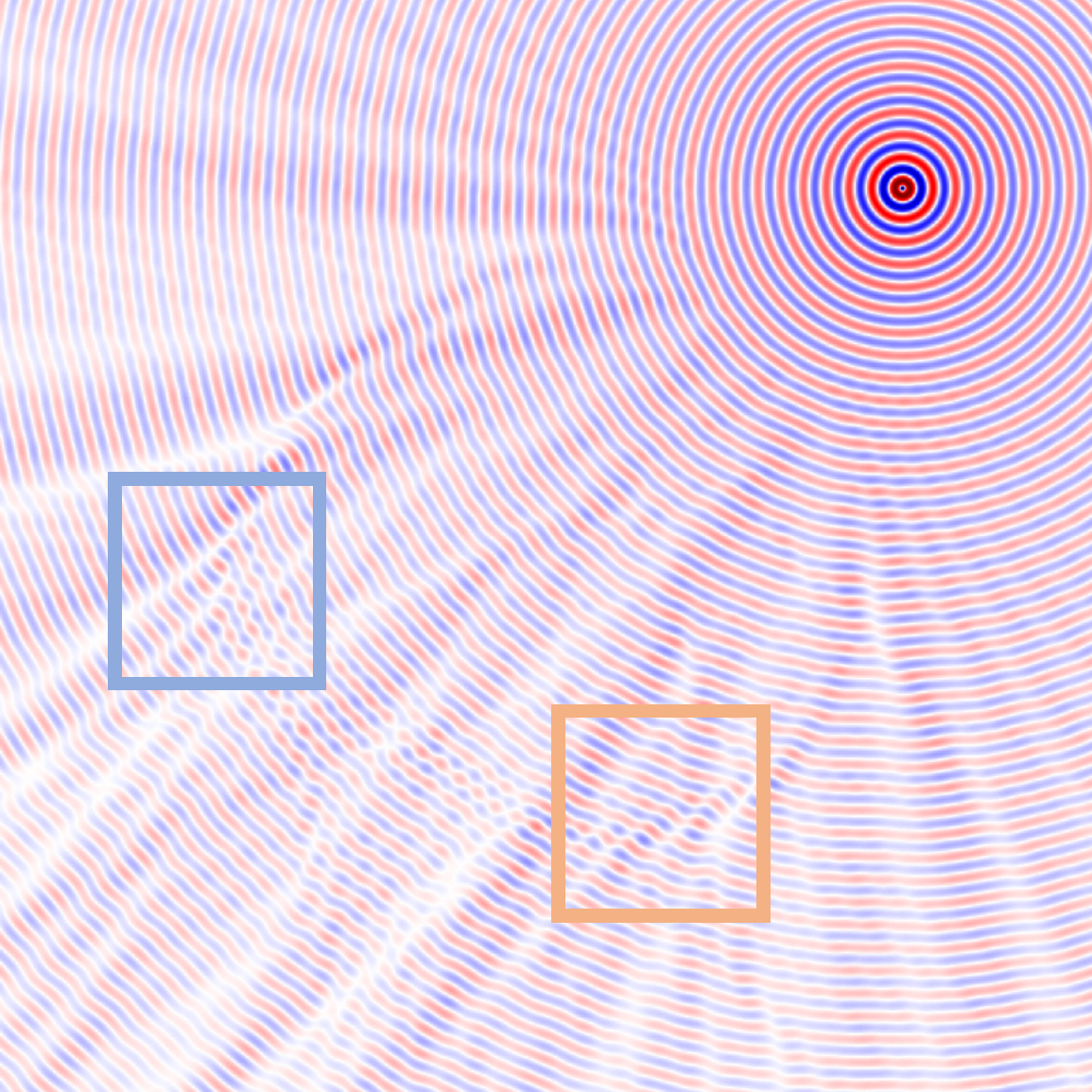} \\ \vspace{0.5mm}
        \includegraphics[width=0.48\linewidth]{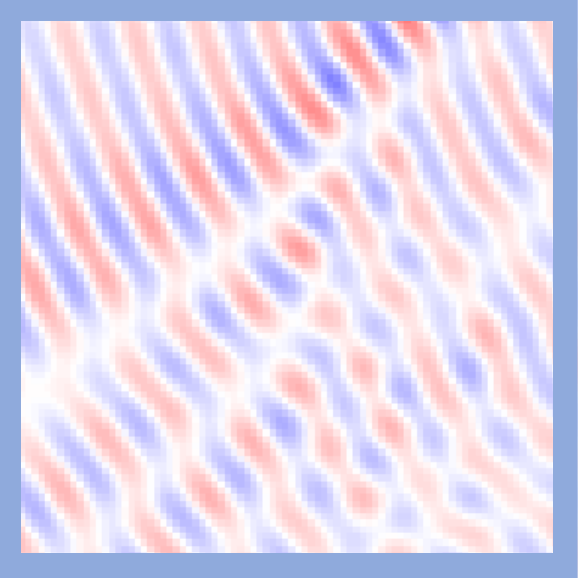}\hfill%
        \includegraphics[width=0.48\linewidth]{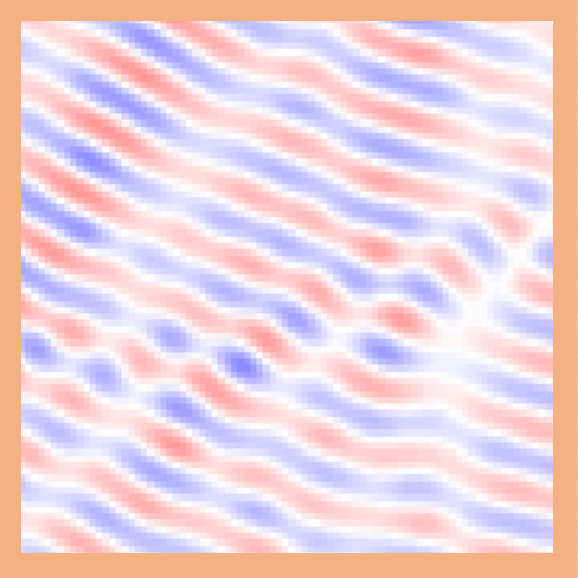}
    \end{minipage}%
    \begin{minipage}[b]{0.12\textwidth}
        \centering
        \includegraphics[width=\linewidth]{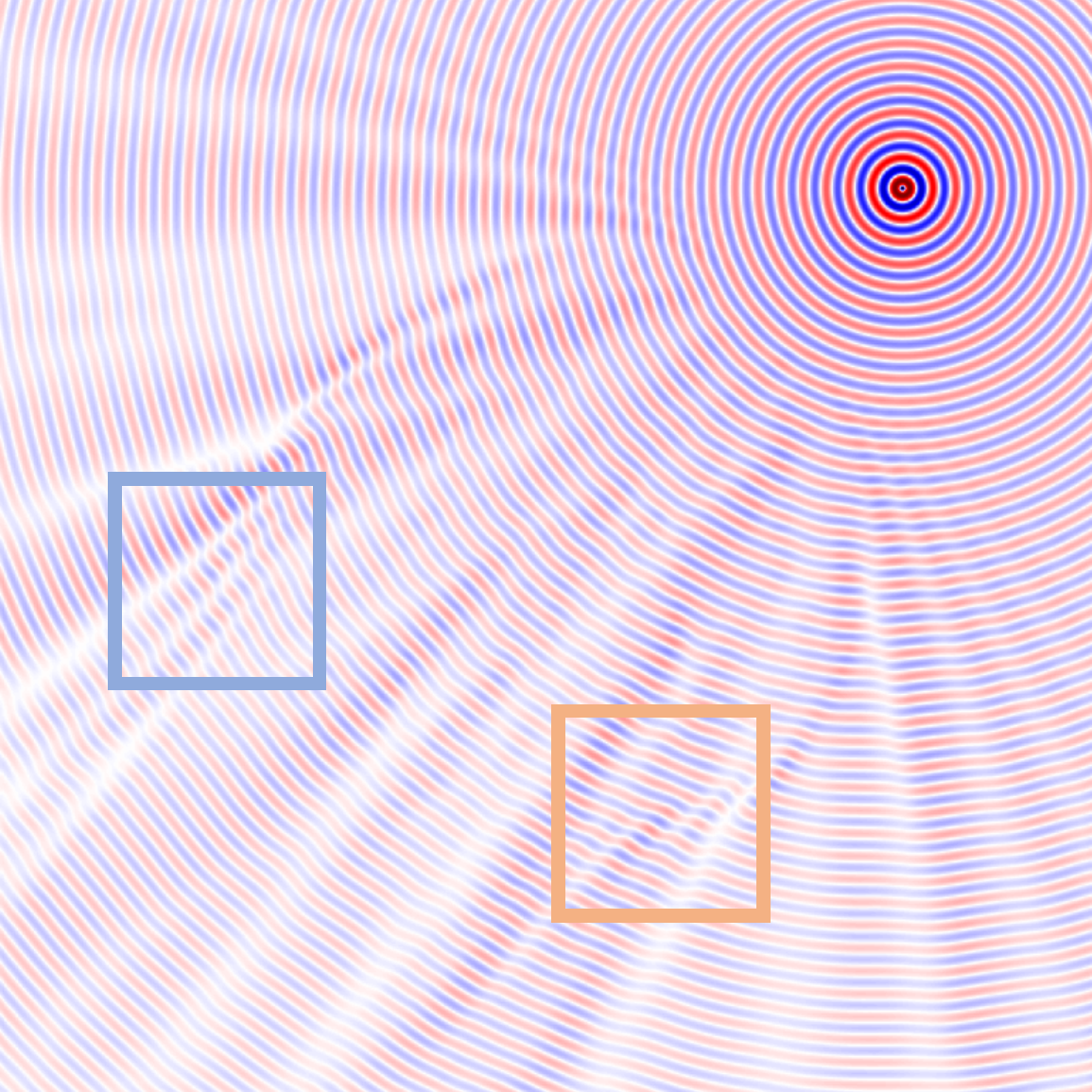} \\ \vspace{0.5mm}
        \includegraphics[width=0.48\linewidth]{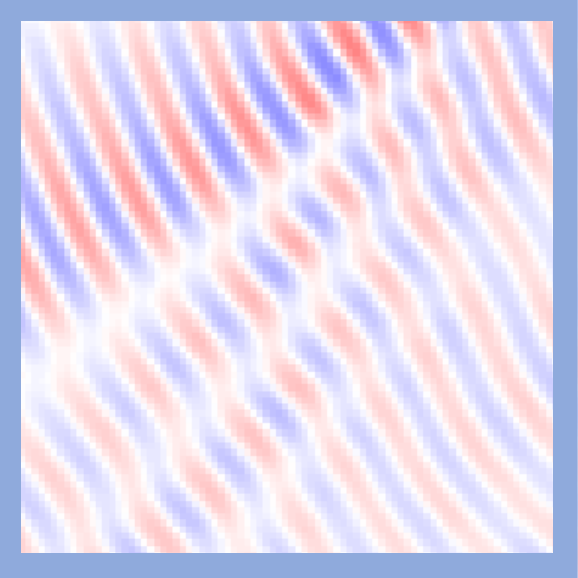}\hfill%
        \includegraphics[width=0.48\linewidth]{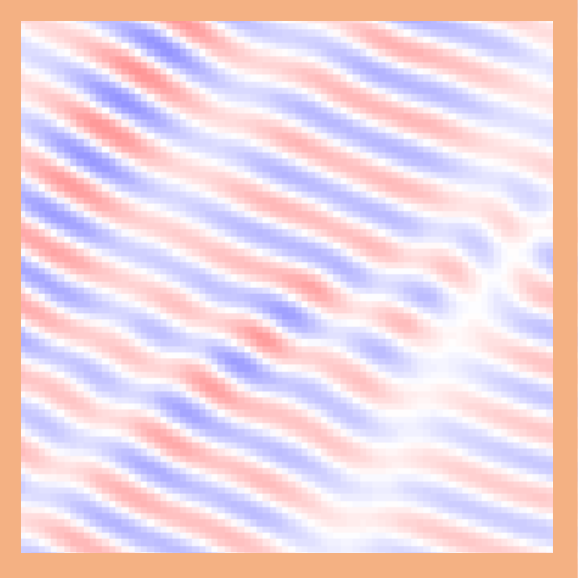}
    \end{minipage}%
    \begin{minipage}[b]{0.12\textwidth}
        \centering
        \includegraphics[width=\linewidth]{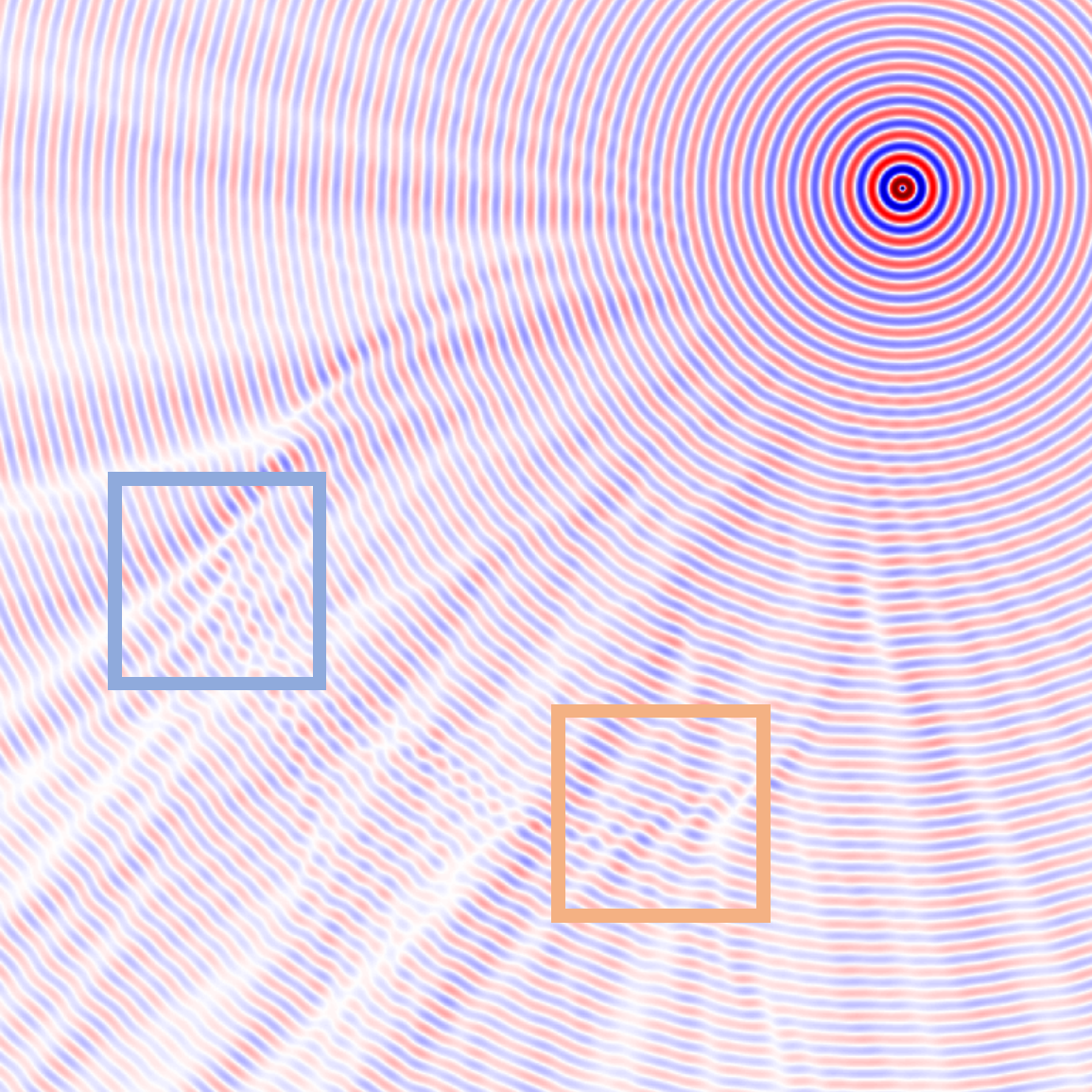} \\ \vspace{0.5mm}
        \includegraphics[width=0.48\linewidth]{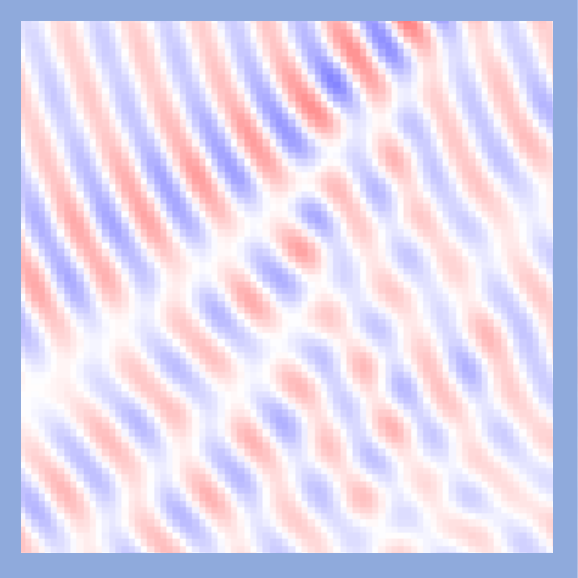}\hfill%
        \includegraphics[width=0.48\linewidth]{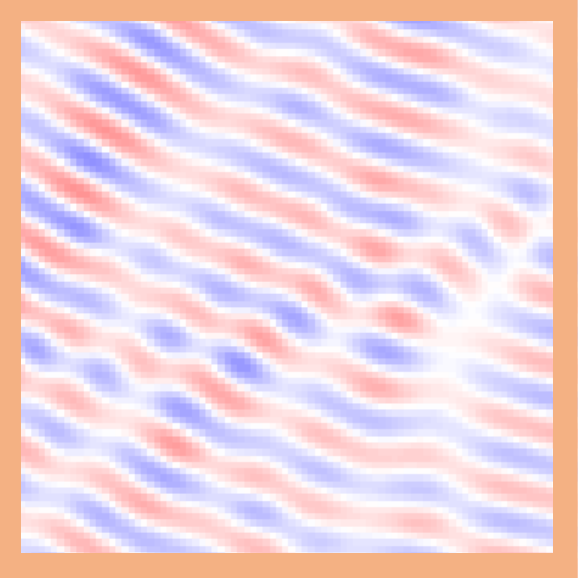}
    \end{minipage}%
    \begin{minipage}[b]{0.12\textwidth}
        \centering
        \includegraphics[width=\linewidth]{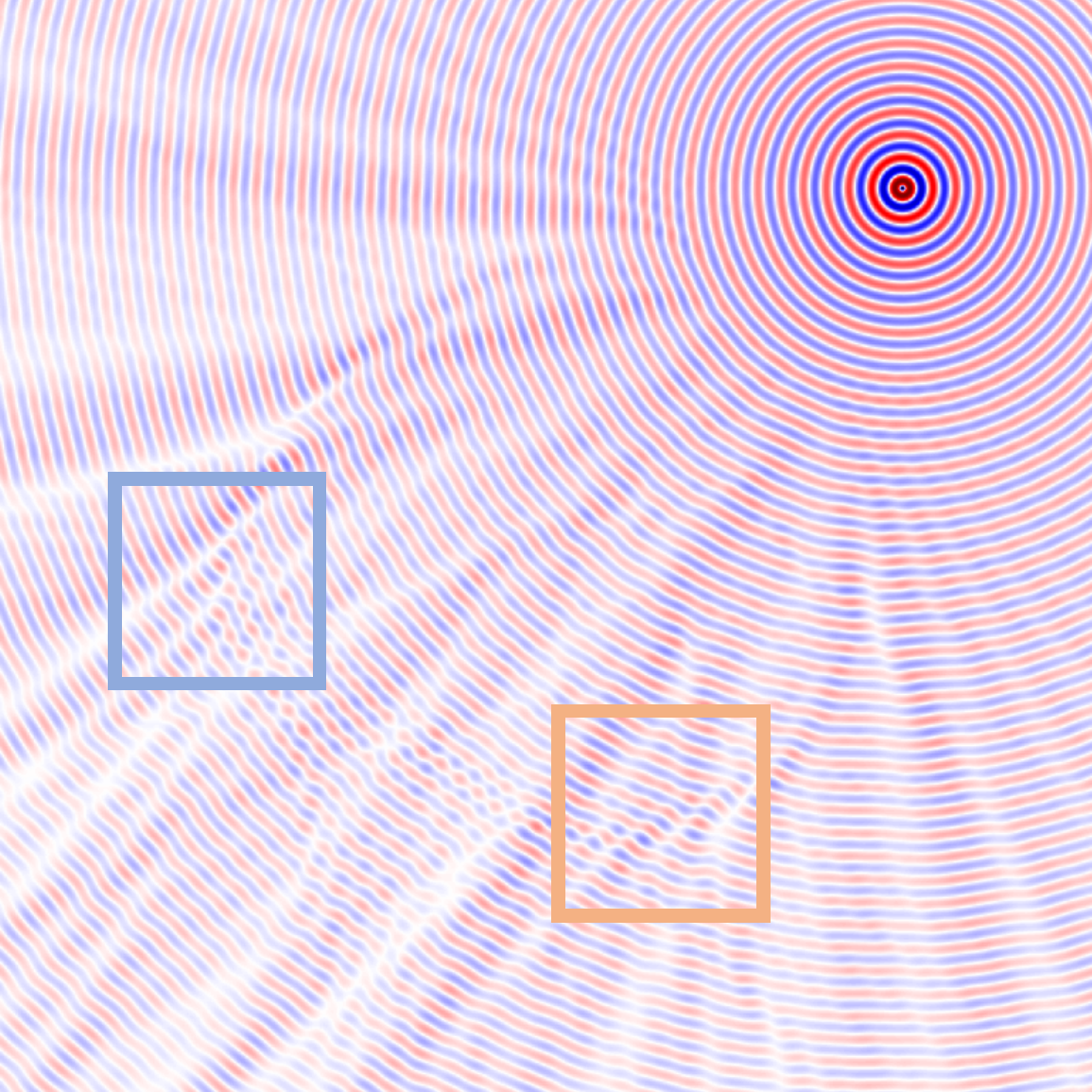} \\ \vspace{0.5mm}
        \includegraphics[width=0.48\linewidth]{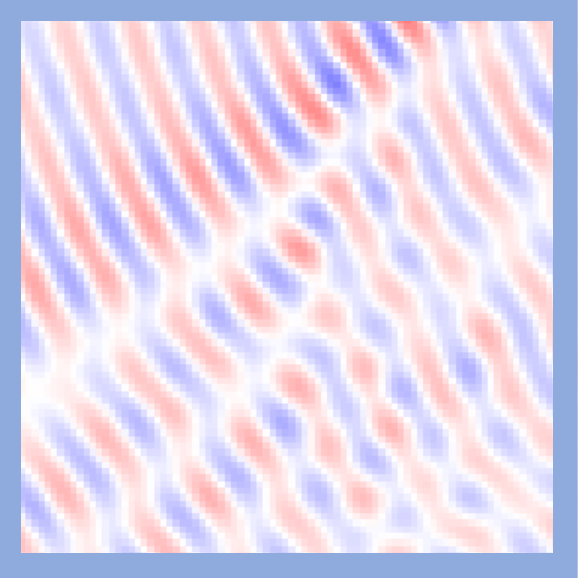}\hfill%
        \includegraphics[width=0.48\linewidth]{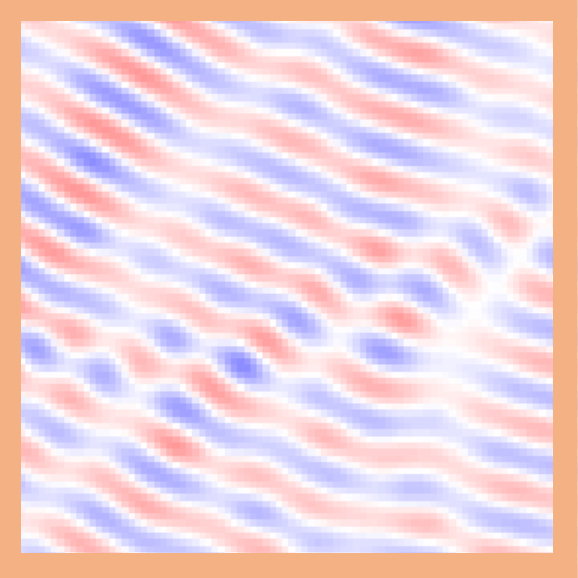}
    \end{minipage}%
    \begin{minipage}[b]{0.12\textwidth}
        \centering
        \includegraphics[width=\linewidth]{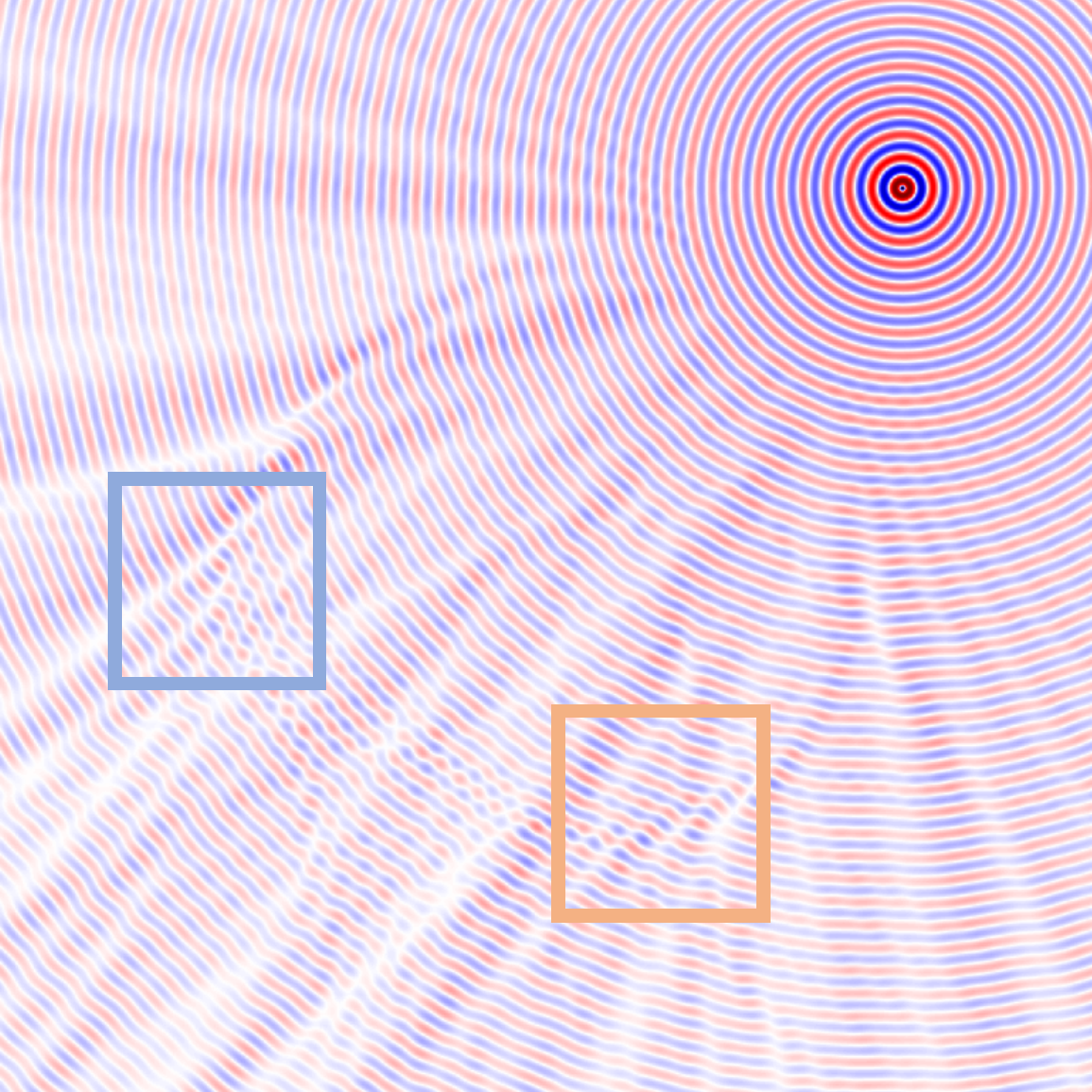} \\ \vspace{0.5mm}
        \includegraphics[width=0.48\linewidth]{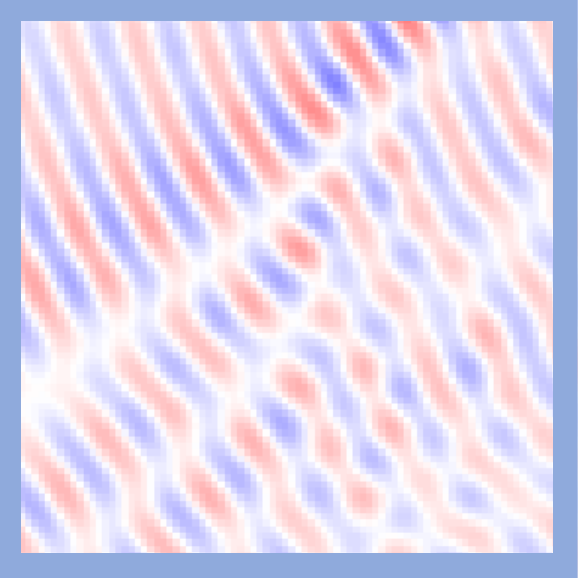}\hfill%
        \includegraphics[width=0.48\linewidth]{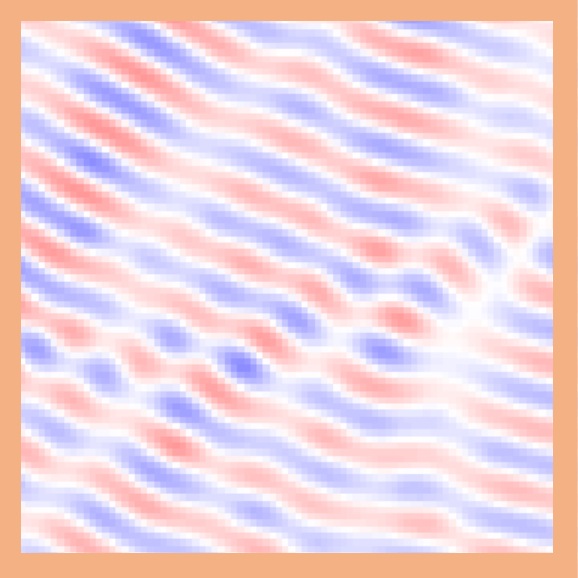}
    \end{minipage}%
    \begin{minipage}[b]{0.12\textwidth}
        \centering
        \includegraphics[width=\linewidth]{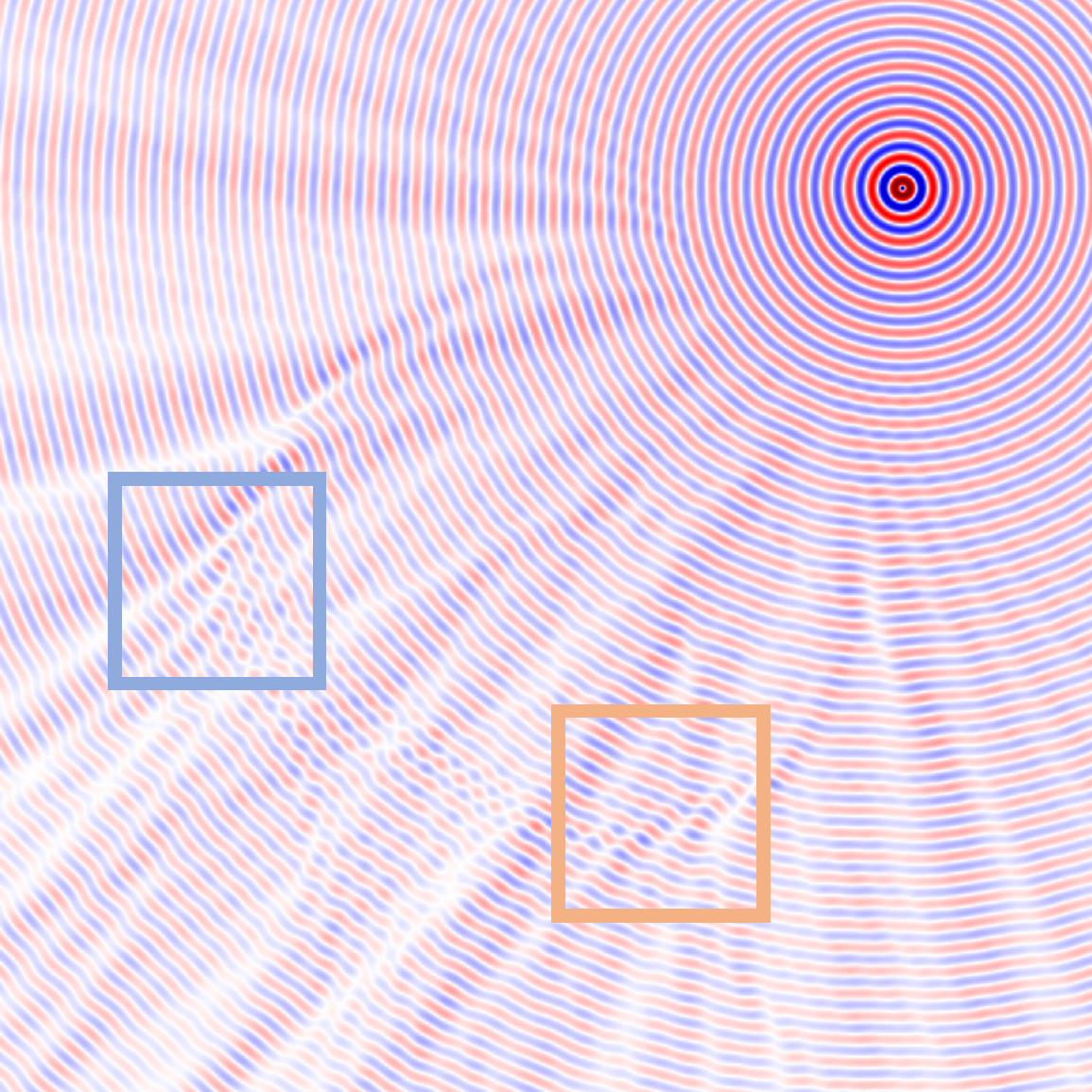} \\ \vspace{0.5mm}
        \includegraphics[width=0.48\linewidth]{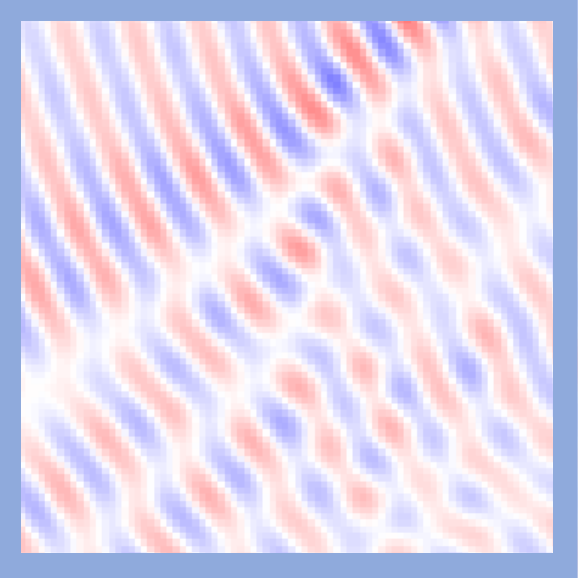}\hfill%
        \includegraphics[width=0.48\linewidth]{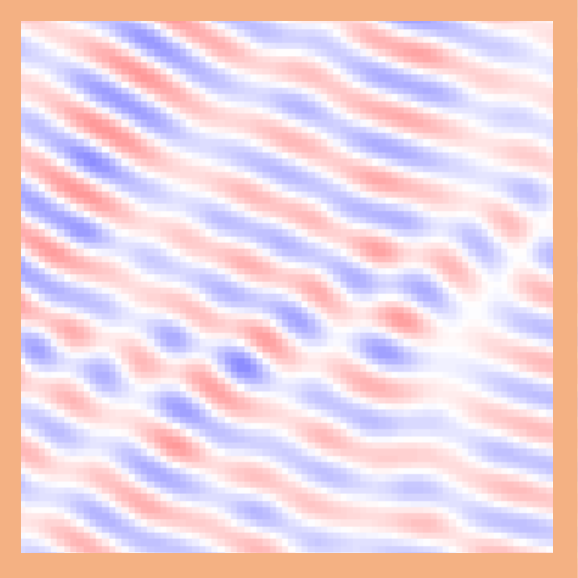}
    \end{minipage}%
    \begin{minipage}[b]{0.07\textwidth} % Spacer
        \hspace{0pt}
    \end{minipage}
    % No \\ or \vspace after the last row of minipages before the main caption
    \caption{\textbf{Forward simulation results at 300 kHz.} Comparison of wavefield predictions for four breast types using a numerical solver (CBS) and five baseline neural operators.}
    \label{fig:forward-300k}
\end{figure*}
Figure~\ref{fig:forward-300k} compares wavefield predictions at 300 kHz for representative phantoms using various neural operators, while results for 400 and 500 kHz are provided in the Supplementary Information. Table~\ref{tab:forward-baselines} further summarizes the performance of the baselines across all breast categories and frequencies.  Quantitative analysis indicates that MgNO consistently achieved the lowest prediction errors; BFNO yielded the lowest statistical error among all FNO variants, whereas AFNO attained the lowest maximum error among all FNO variants. However, UNet architectures cannot capture the medium’s scattering behavior, resulting in the highest errors.

Next, we compared the performance of DeepONet, InversionNet, NIO and optimization-based FWI baselines using neural operators in USCT inverse imaging task. All methods were trained and tested using three frequencies (300, 400, and 500 kHz) data. All baselines were trained on a single NVIDIA A800 PCIe 80 GB GPU, with measurements as input ($3 \times 256 \times 256$) and ground-truth images as output ($480 \times 480$). The optimization-based FWI performed gradient descent reconstruction, where gradients were calculated using the adjoint method with above five forward simulation baselines. 
\begin{figure*}[htbp]
	\centering
	\setlength{\tabcolsep}{0.5pt}
	\setlength{\fboxrule}{1pt}
	%\vspace*{1.5cm}
	\begin{tabular}{c}
		\begin{tabular}{ccccccccc}
			&\small{GT}  & 
			\small{Optim (CBS)} & 
			\small{Optim (FNO)} &
			\small{NIO}& 
			\small{InversionNet}  & 
			\small{DeepONet}
			\\ 
			\begin{turn}{90}   \quad\quad \small{HET} \end{turn} & 
			  \multicolumn{1}{c}{
			  	\begin{overpic}[width=0.15\linewidth]{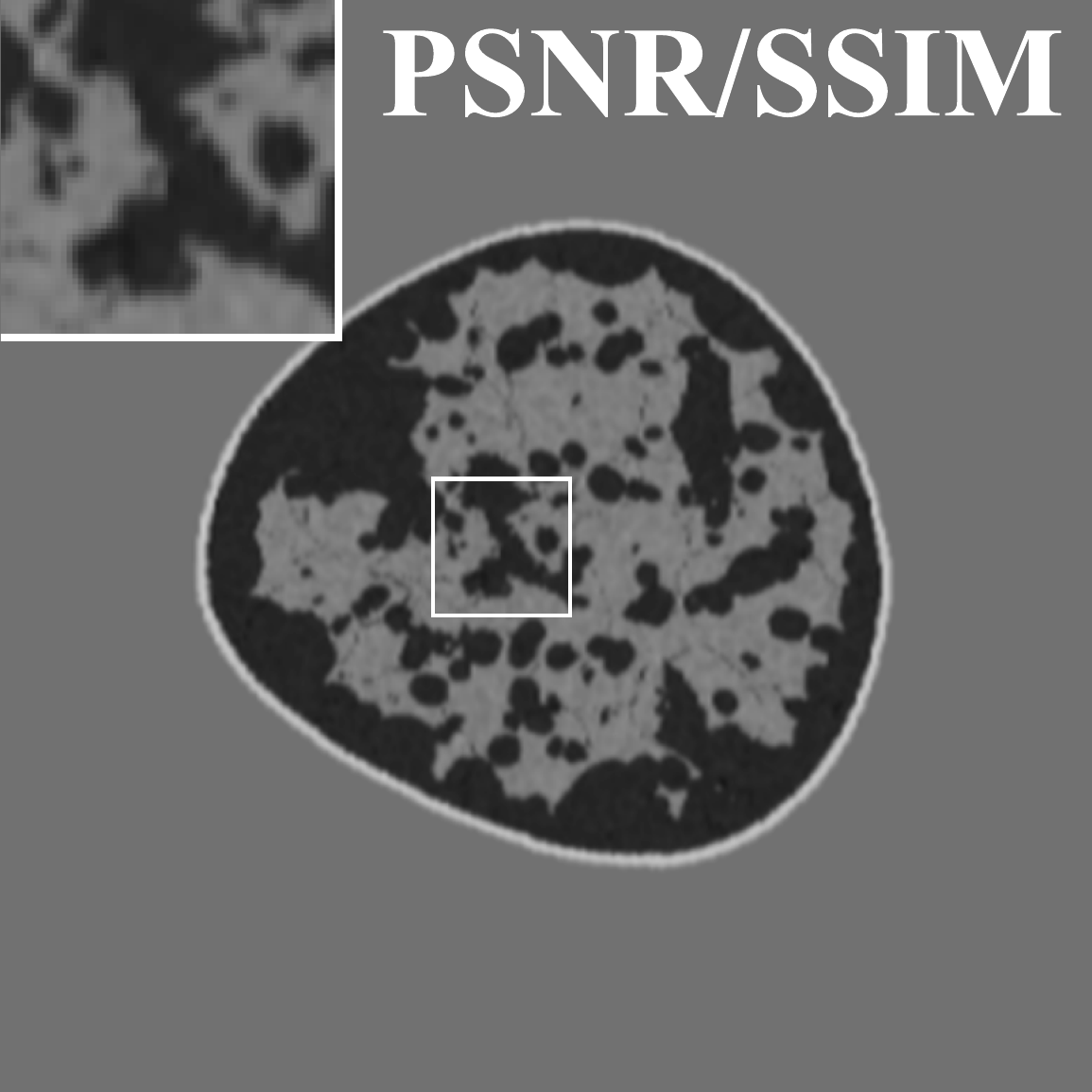}
			  	\end{overpic}
			  }  &
			  \multicolumn{1}{c}{
			  	\begin{overpic}[width=0.15\linewidth]{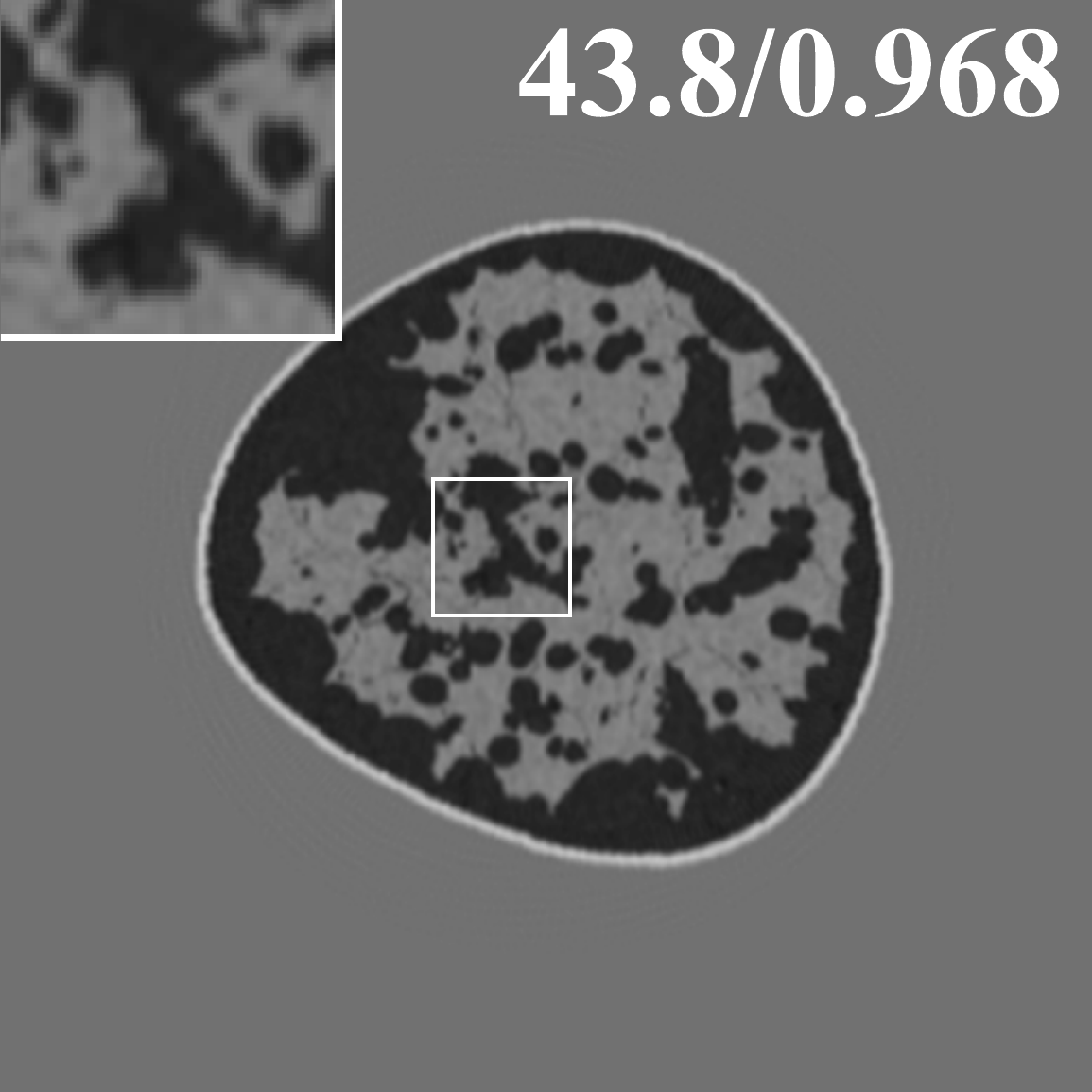}
			  	\end{overpic}
			  }  &
			  \multicolumn{1}{c}{
			  	\begin{overpic}[width=0.15\linewidth]{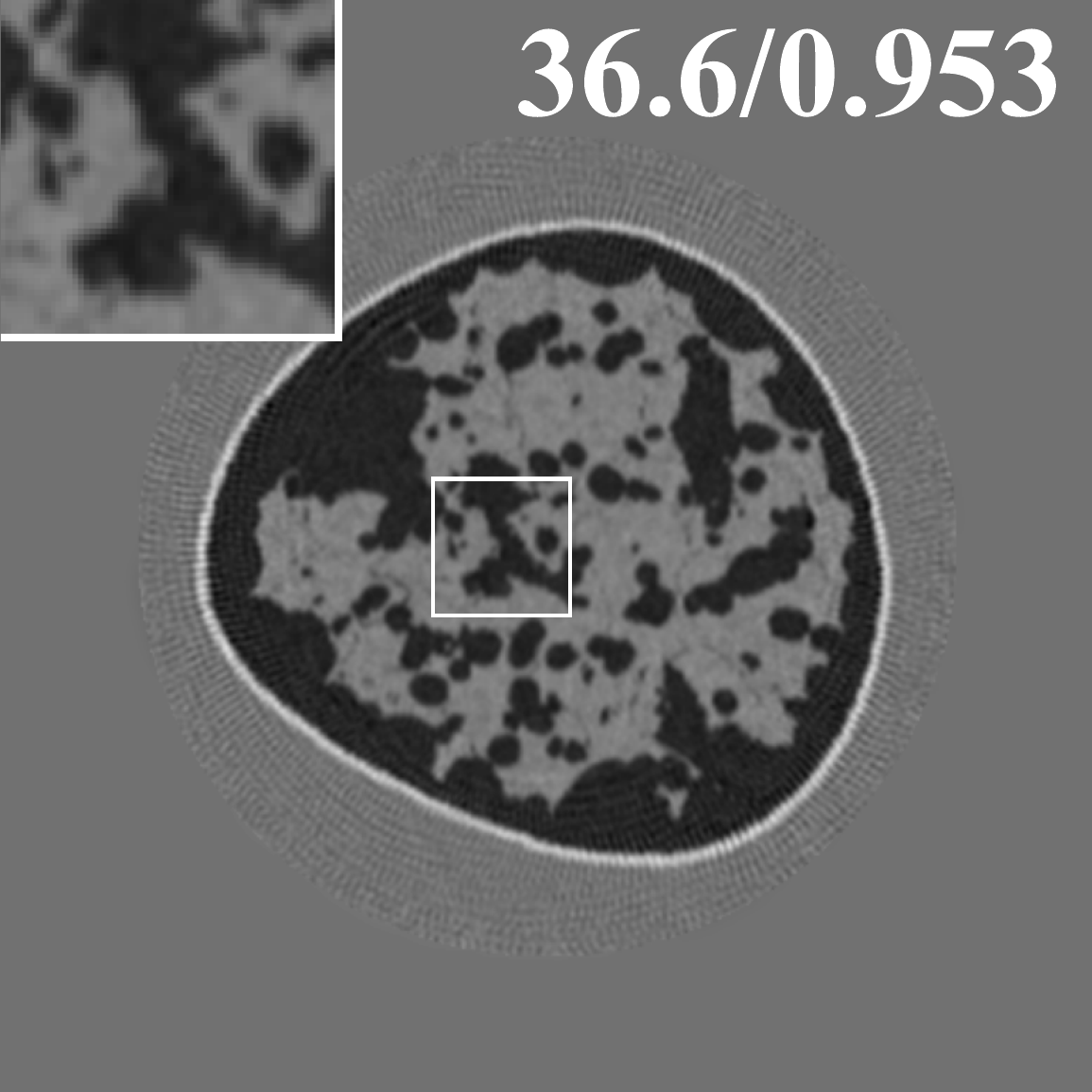}
			  	\end{overpic}
			  }  &
			  \multicolumn{1}{c}{
			  	\begin{overpic}[width=0.15\linewidth]{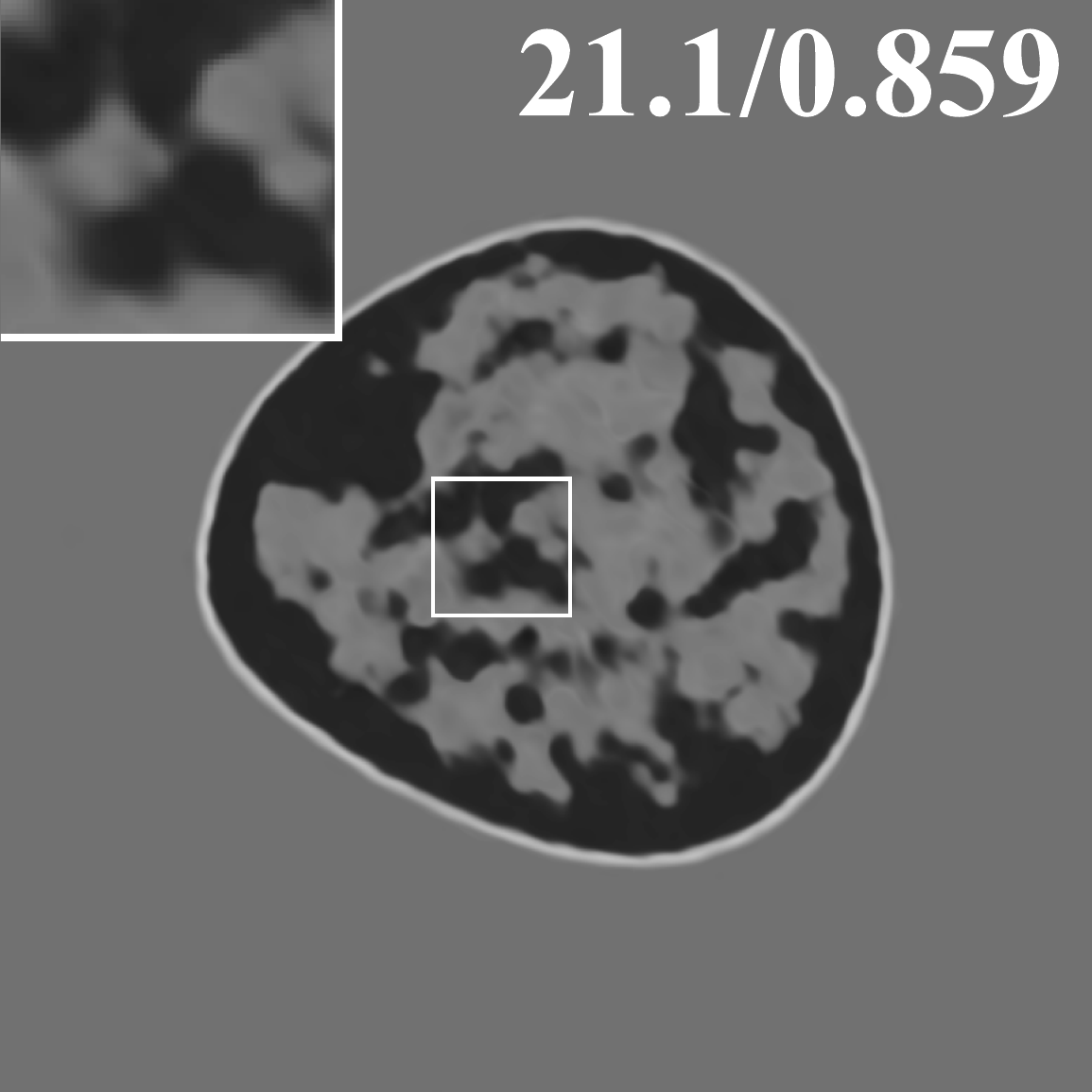}
			  		
			  	\end{overpic}
			  }  &
			  \multicolumn{1}{c}{
			  	\begin{overpic}[width=0.15\linewidth]{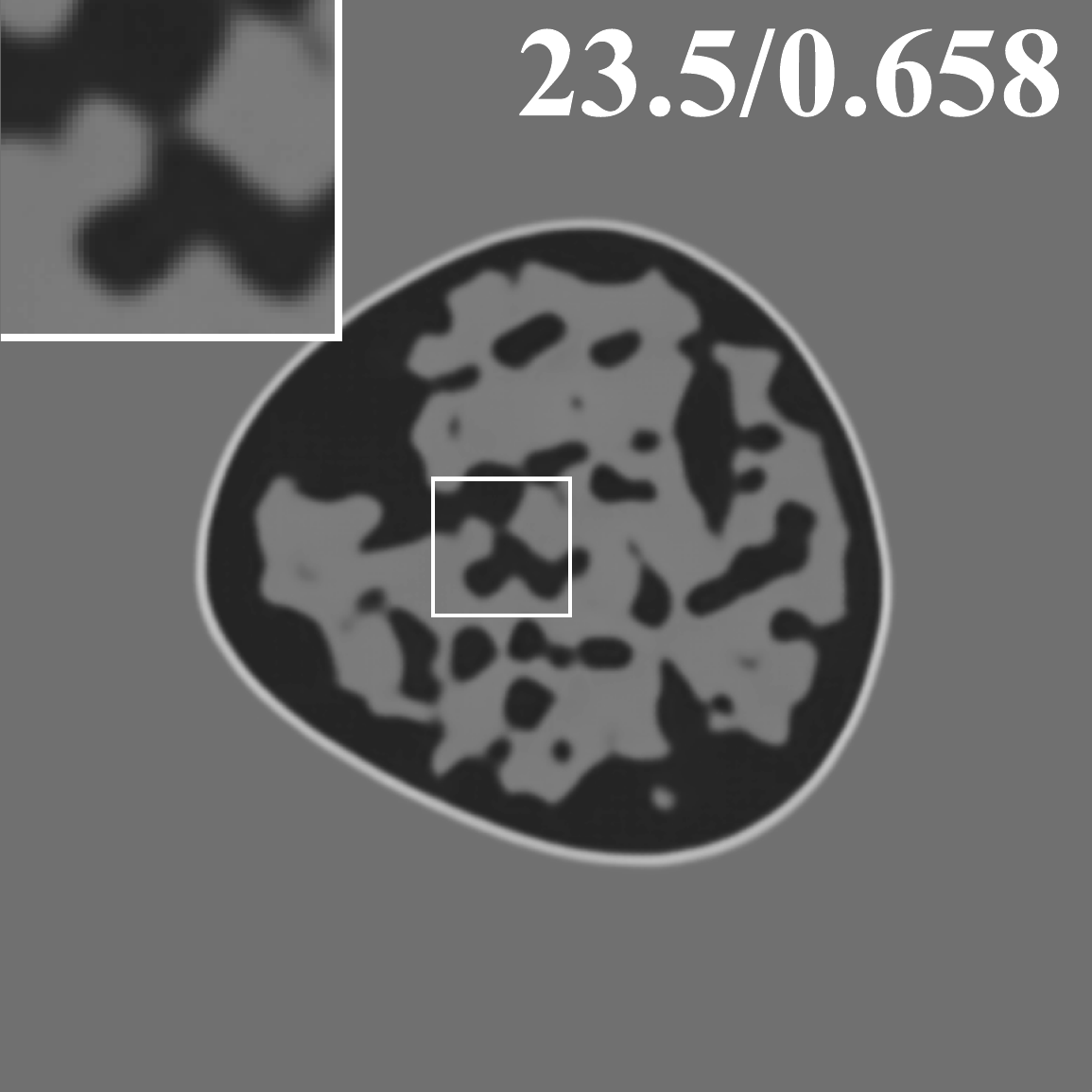}
			  		
			  	\end{overpic}
			  }  &
			  \multicolumn{1}{c}{
			  	\begin{overpic}[width=0.15\linewidth]{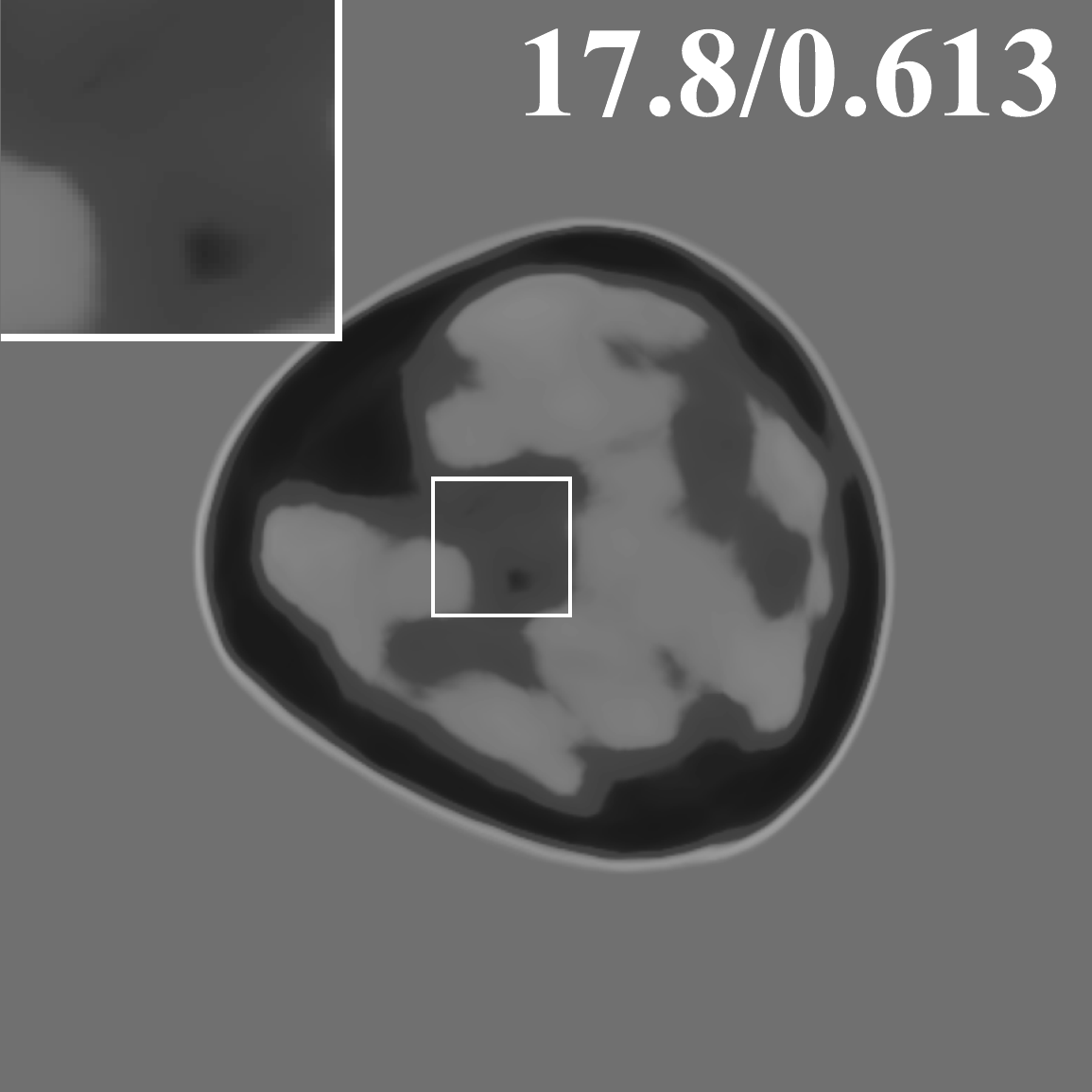}
			  		
			  	\end{overpic}
			  } 
			  \\
			\begin{turn}{90}  \quad\quad\small{FIB} \end{turn} & 
			  \multicolumn{1}{c}{
			  	\begin{overpic}[width=0.15\linewidth]{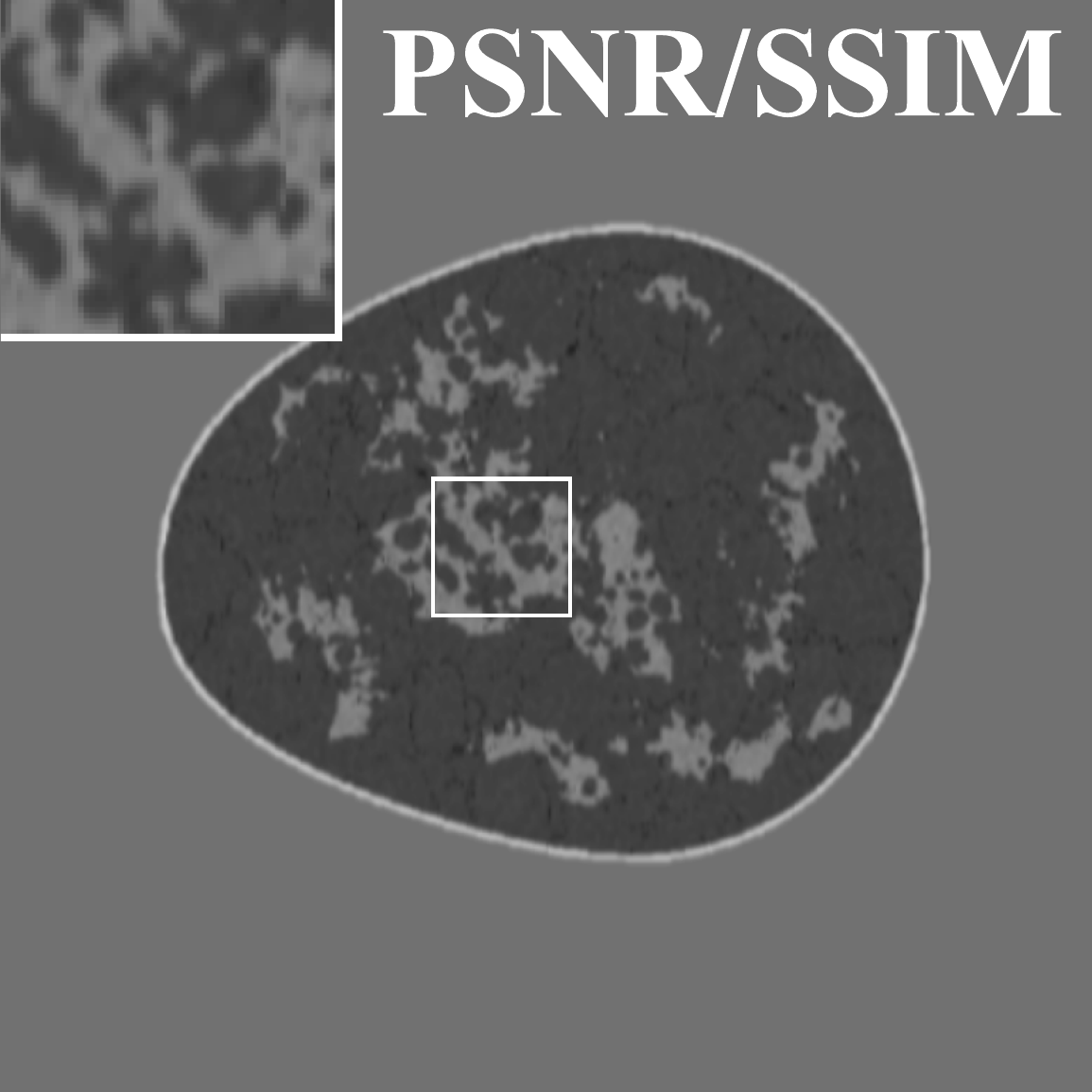}
			  	\end{overpic}
			  }  &
			  \multicolumn{1}{c}{
			  	\begin{overpic}[width=0.15\linewidth]{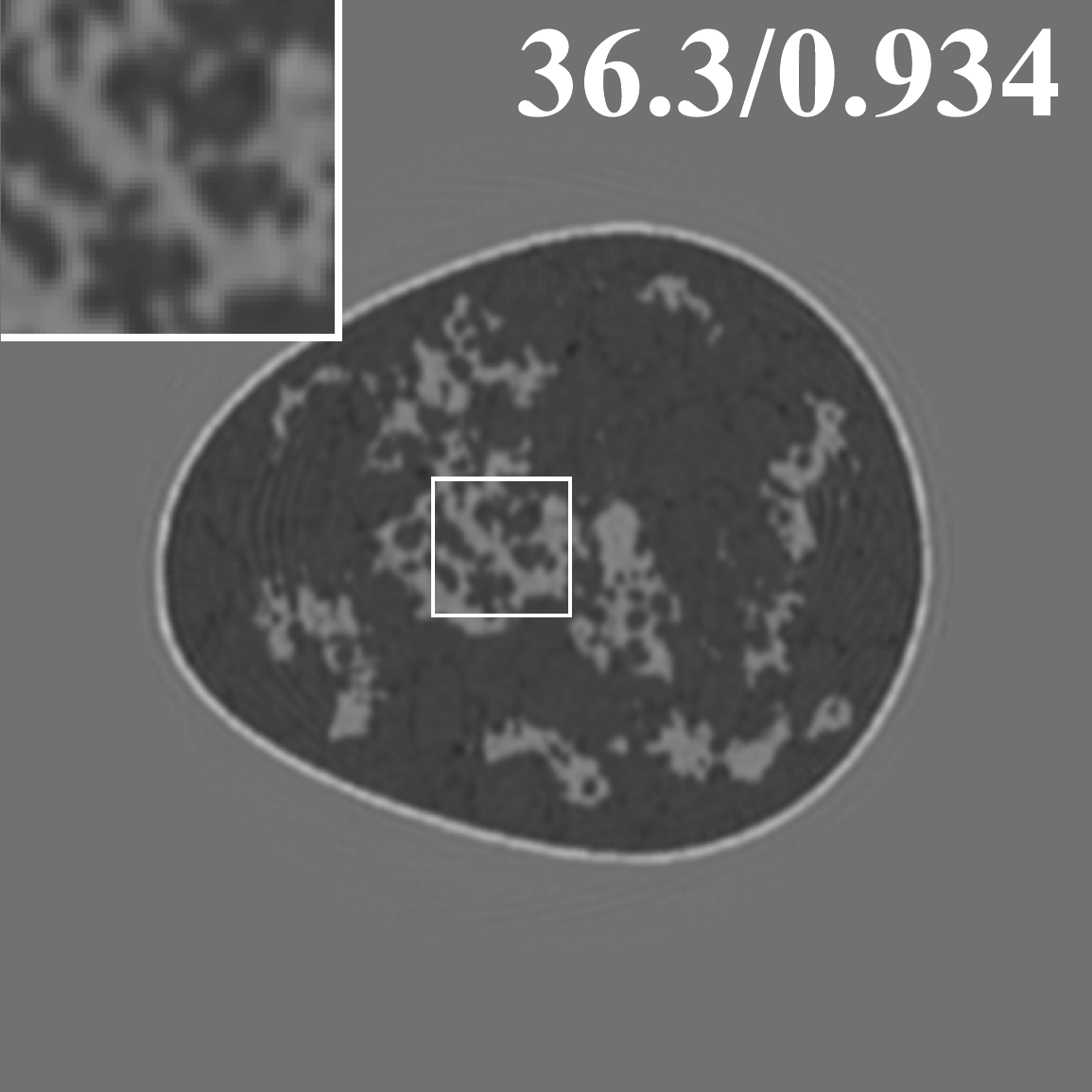}
			  	\end{overpic}
			  }  &
			  \multicolumn{1}{c}{
			  	\begin{overpic}[width=0.15\linewidth]{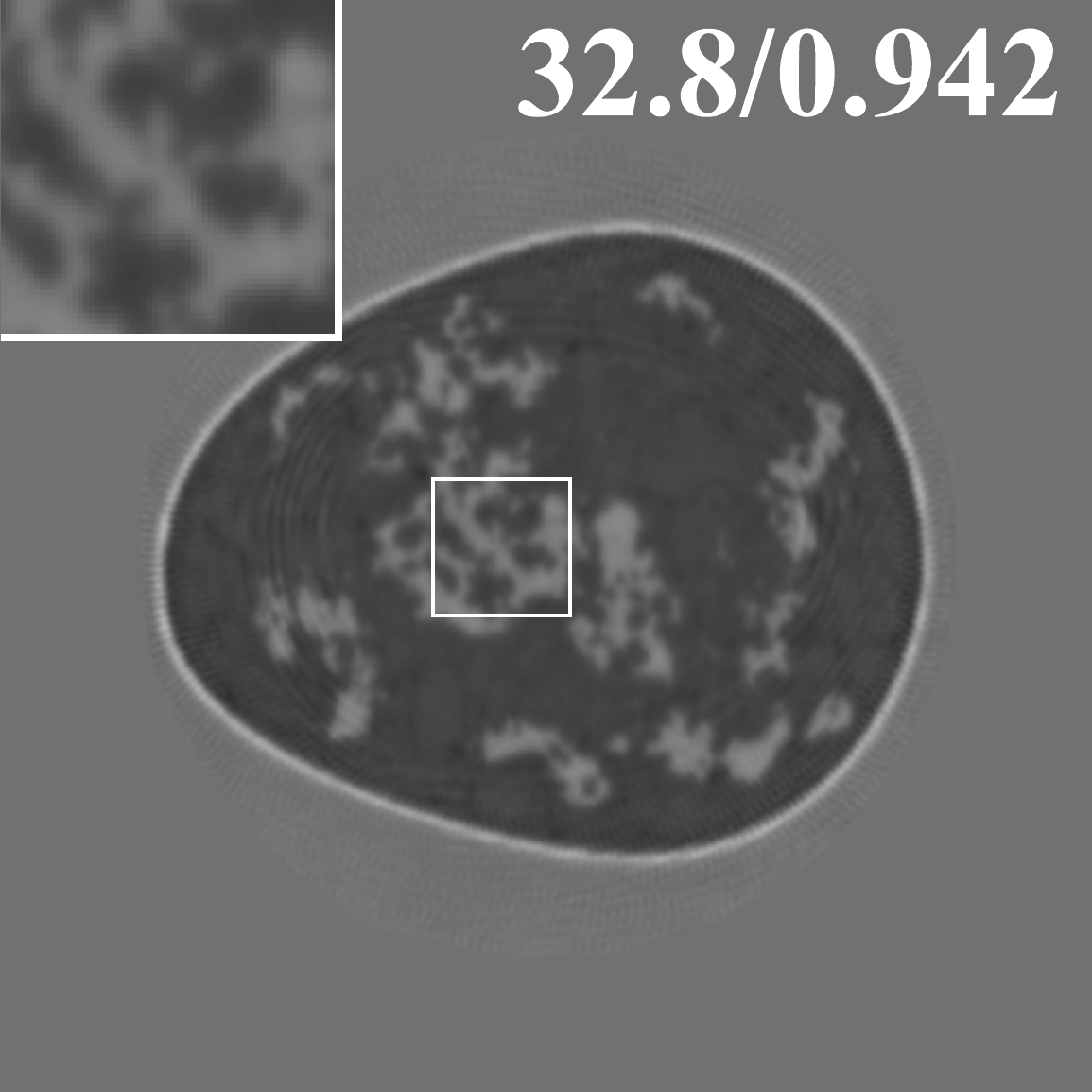}
			  	\end{overpic}
			  }  &
			  \multicolumn{1}{c}{
			  	\begin{overpic}[width=0.15\linewidth]{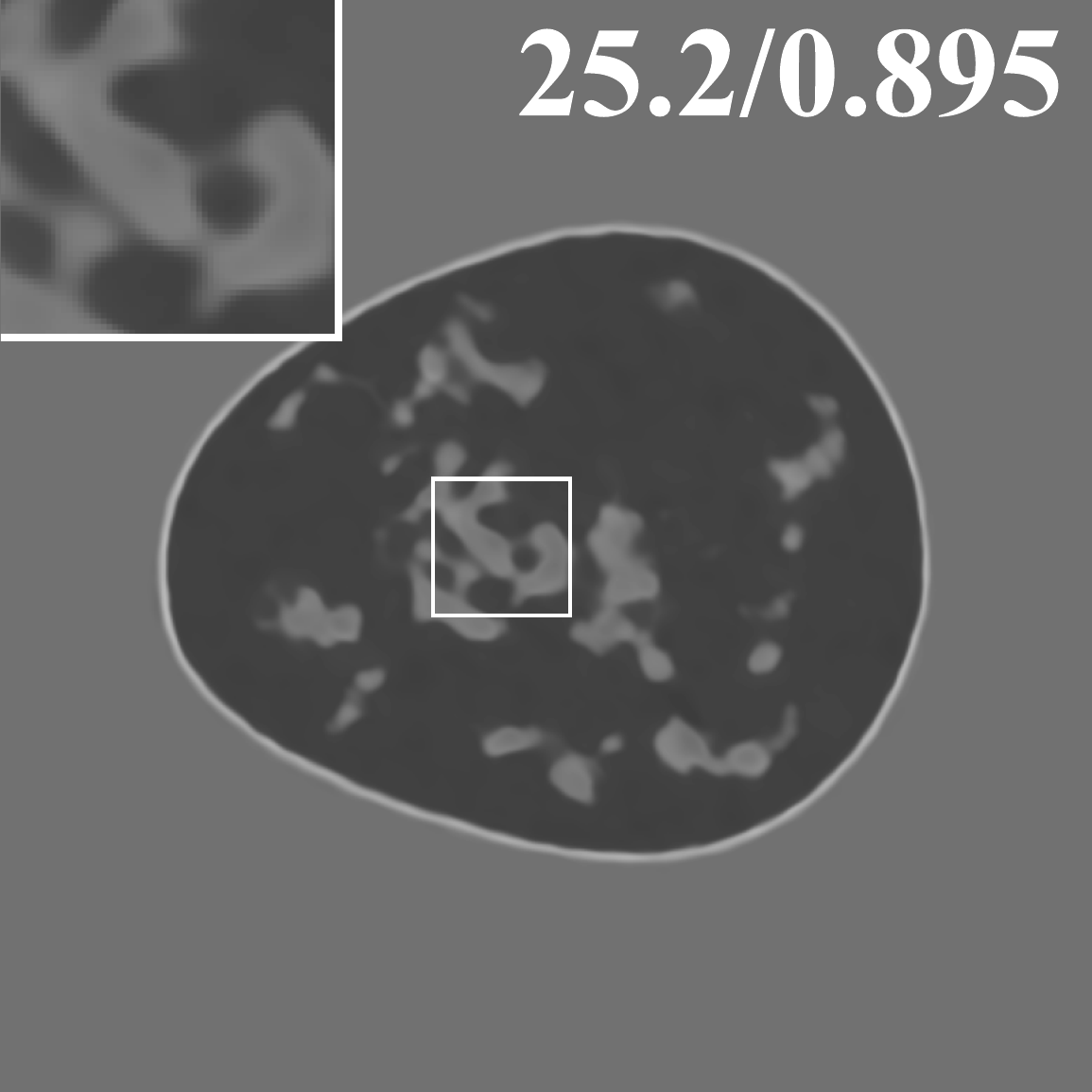}
			  		
			  	\end{overpic}
			  }  &
			  \multicolumn{1}{c}{
			  	\begin{overpic}[width=0.15\linewidth]{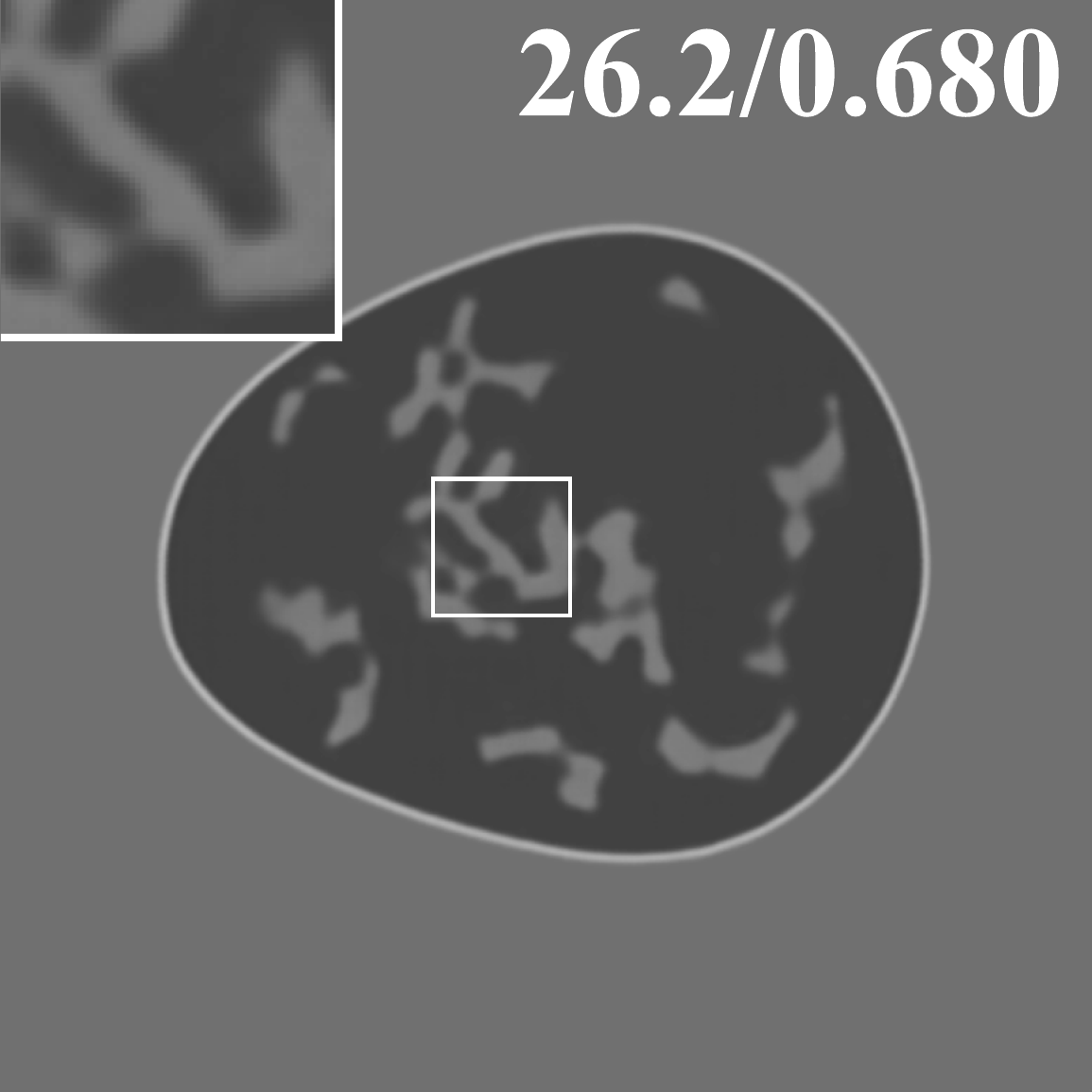}
			  		
			  	\end{overpic}
			  }  &
			  \multicolumn{1}{c}{
			  	\begin{overpic}[width=0.15\linewidth]{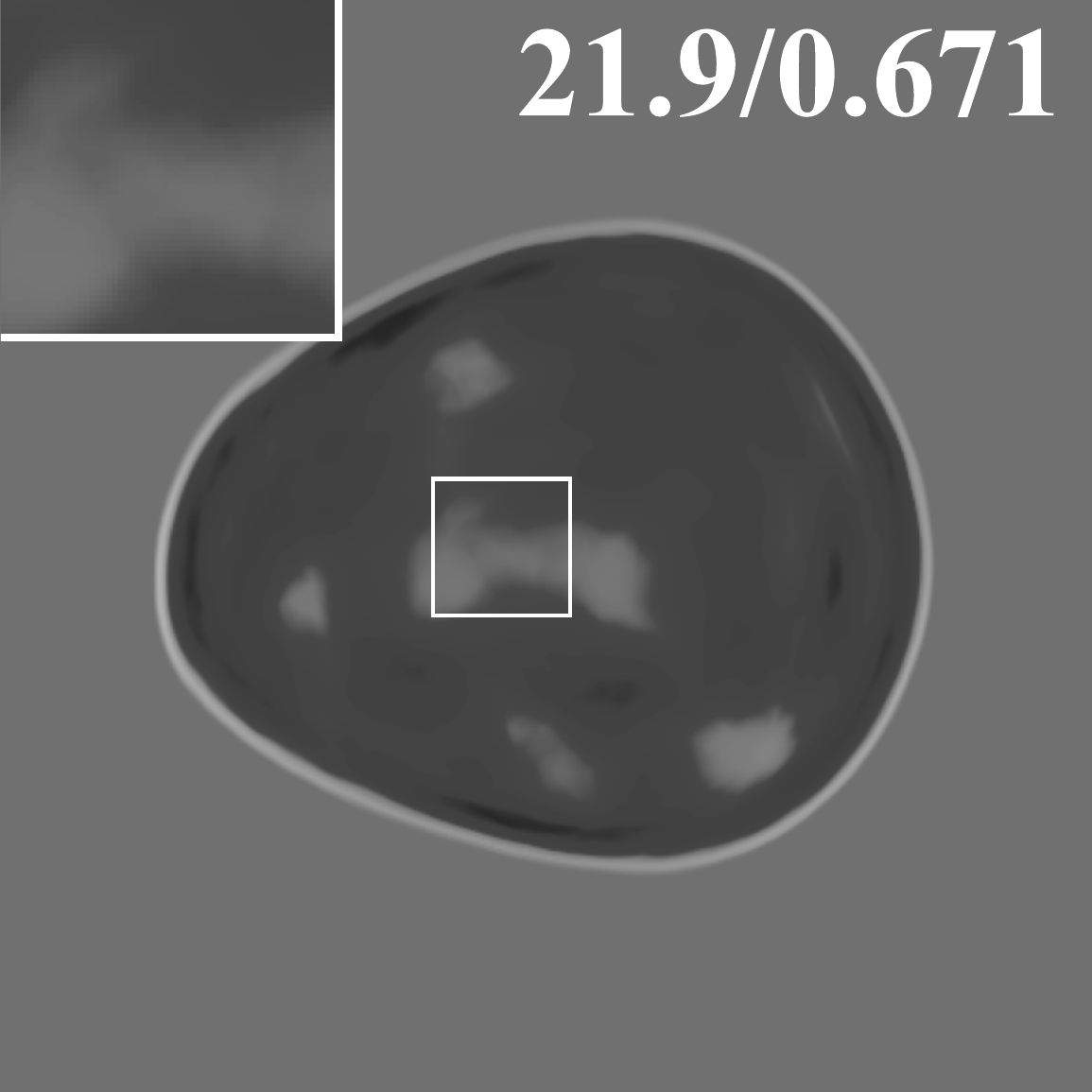}
			  		
			  	\end{overpic}
			  } 
			  \\
				\begin{turn}{90}  \quad\quad \small{FAT} \end{turn} & 
			\multicolumn{1}{c}{
				\begin{overpic}[width=0.15\linewidth]{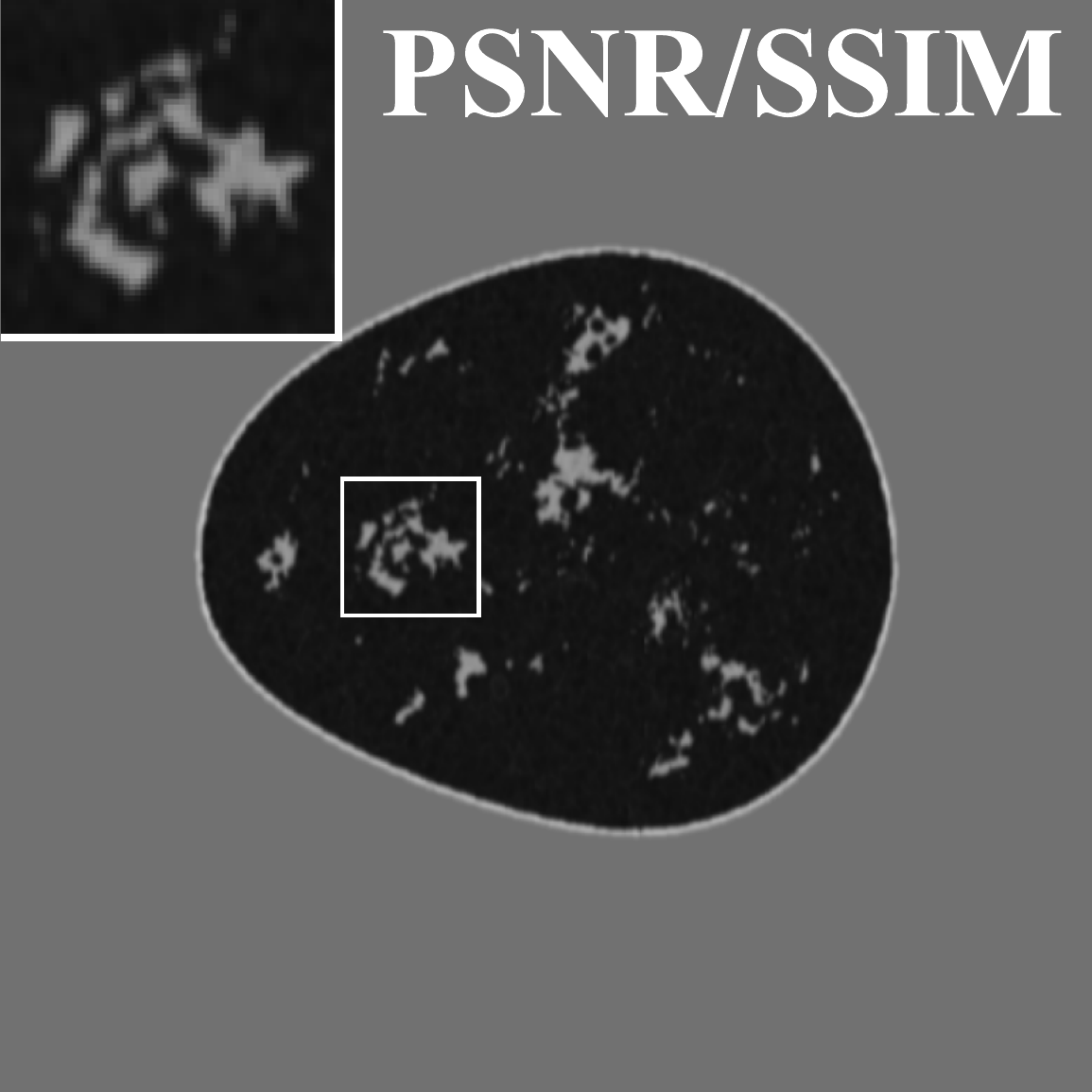}
				\end{overpic}
			}  &
			\multicolumn{1}{c}{
				\begin{overpic}[width=0.15\linewidth]{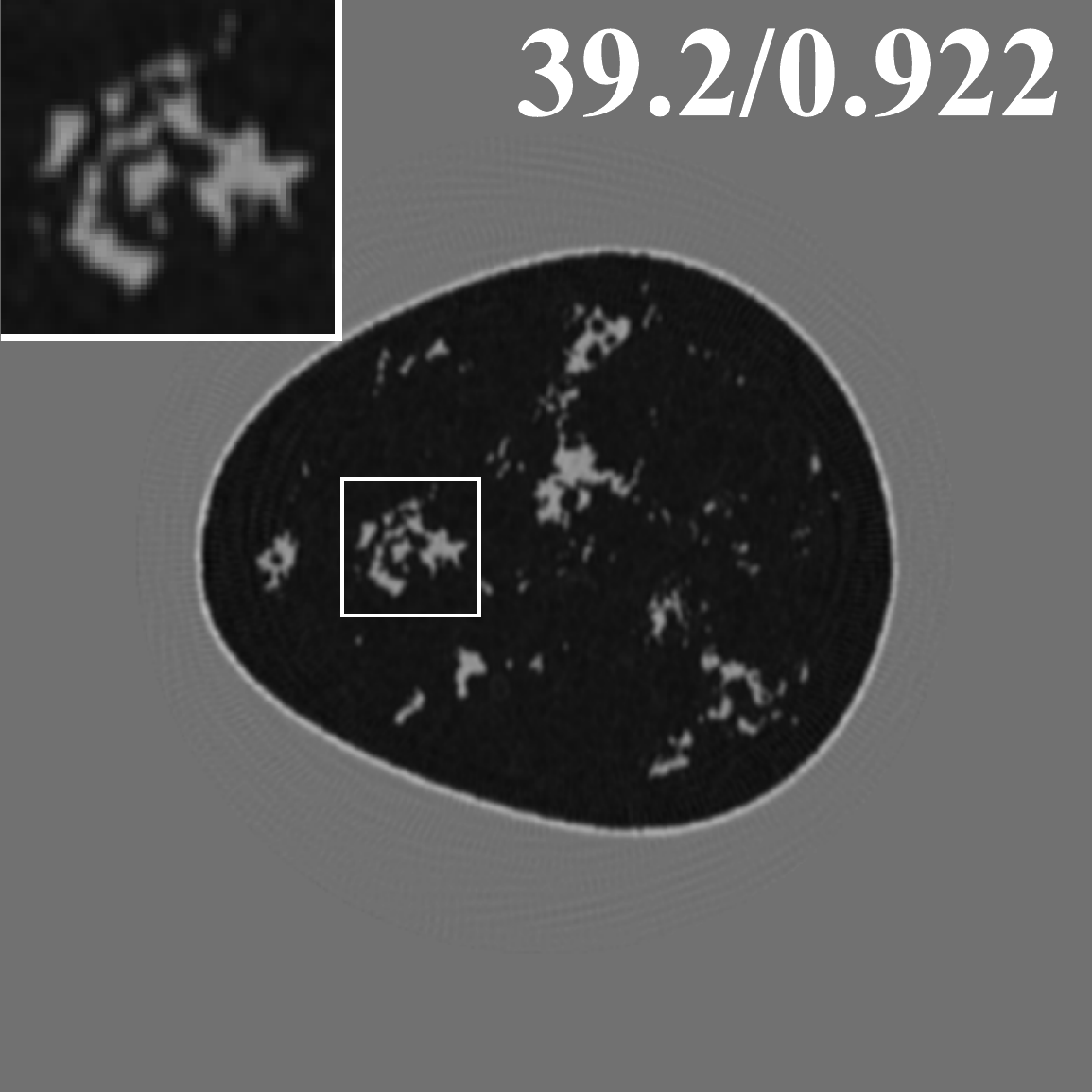}
				\end{overpic}
			}  &
			\multicolumn{1}{c}{
				\begin{overpic}[width=0.15\linewidth]{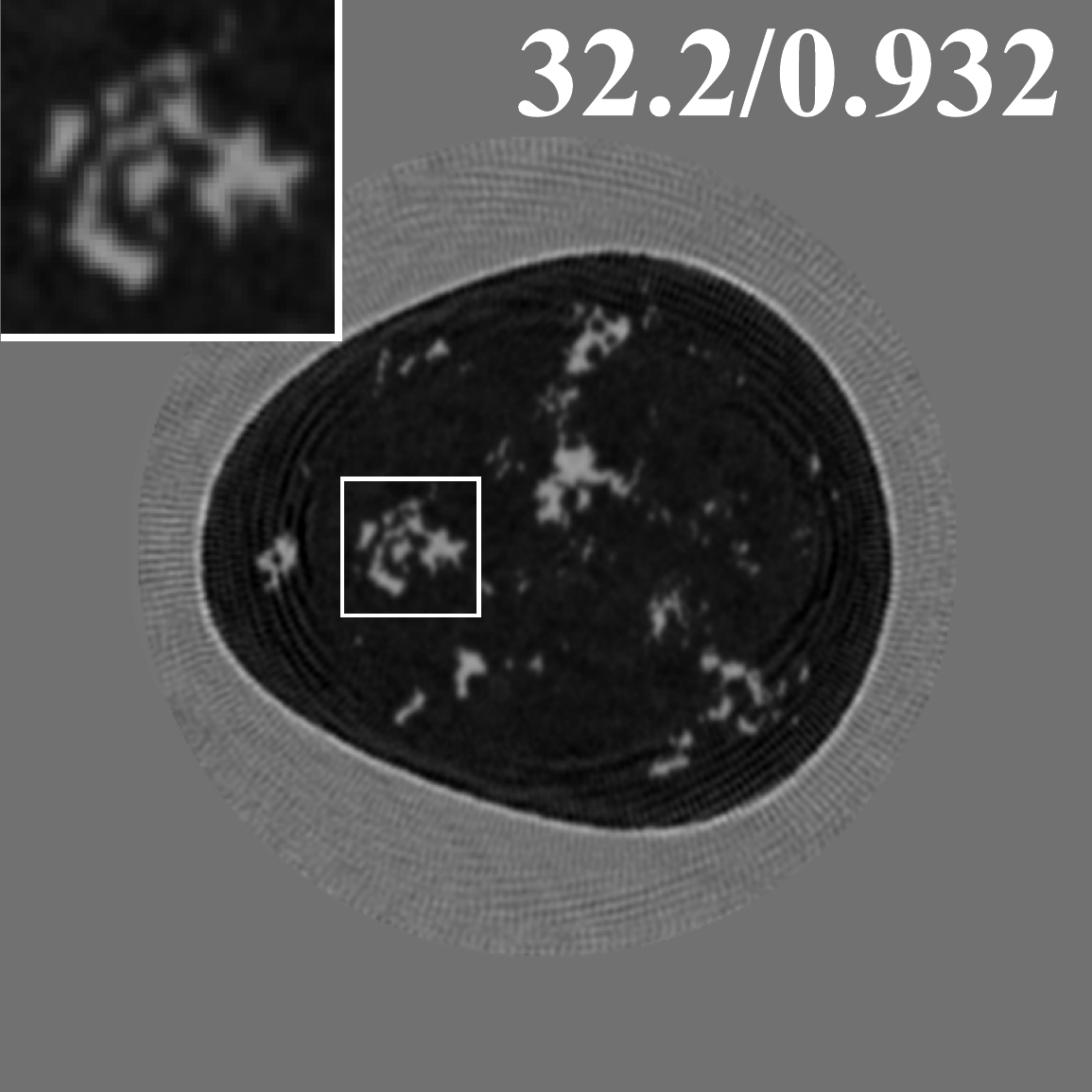}
				\end{overpic}
			}  &
			\multicolumn{1}{c}{
				\begin{overpic}[width=0.15\linewidth]{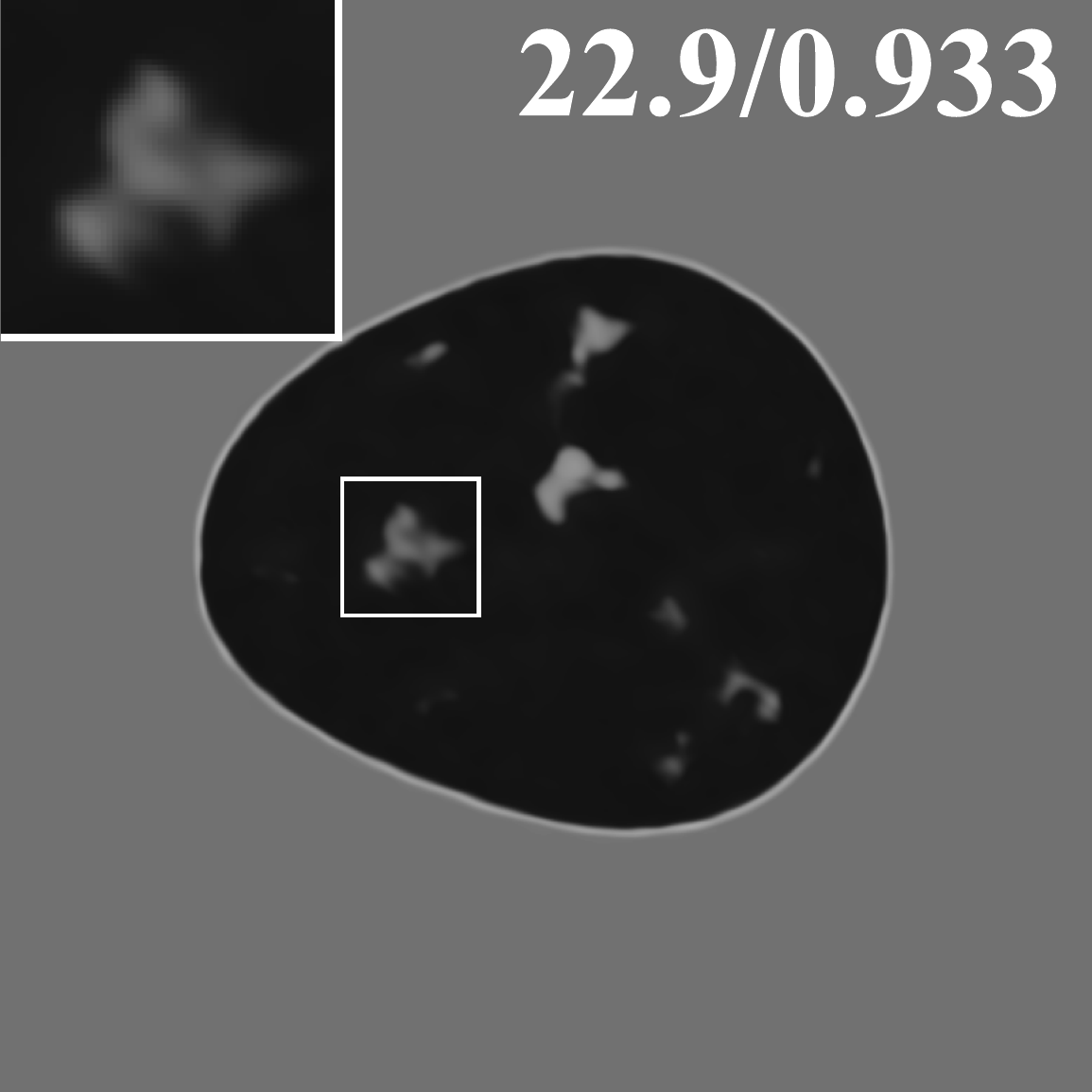}
					
				\end{overpic}
			}  &
			\multicolumn{1}{c}{
				\begin{overpic}[width=0.15\linewidth]{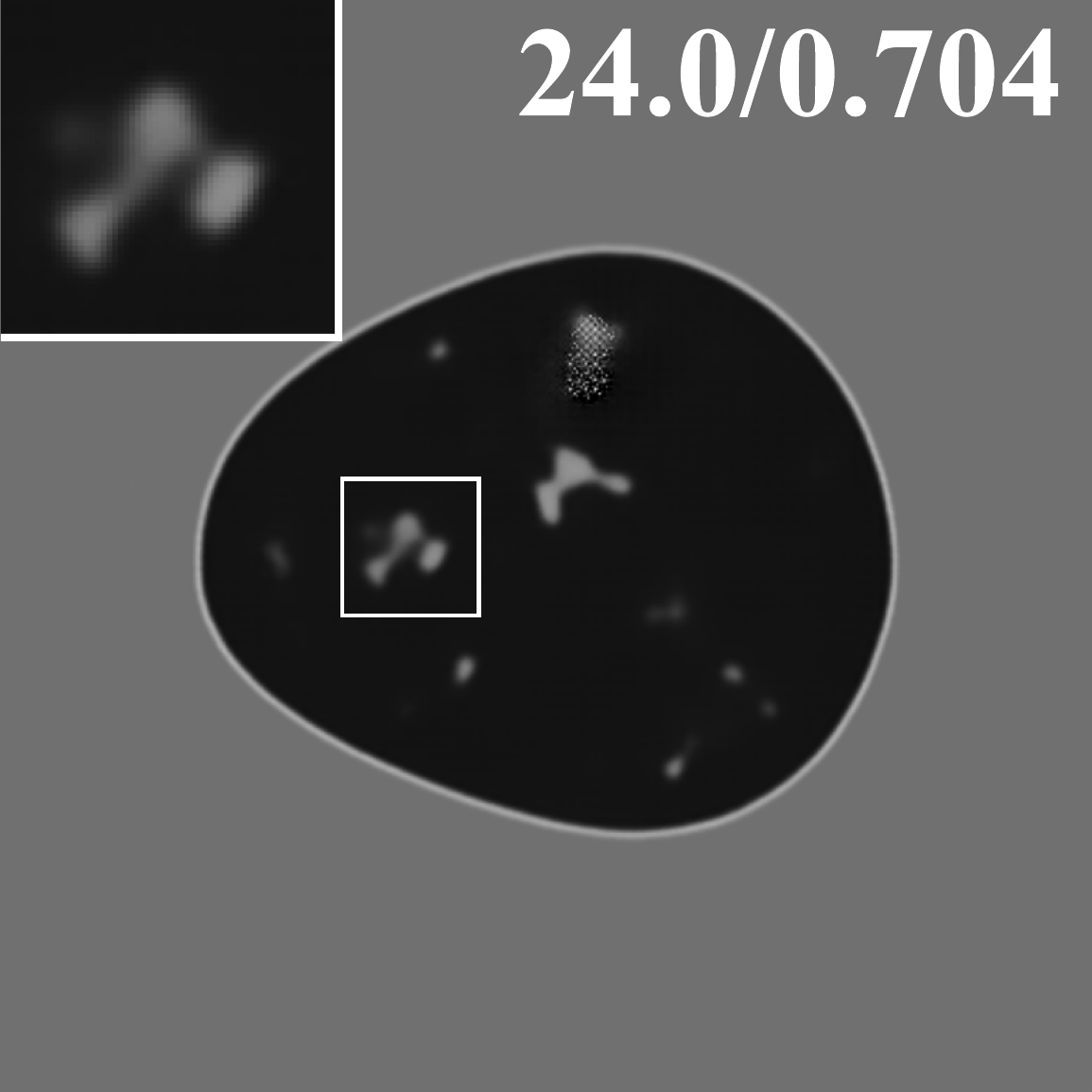}
					
				\end{overpic}
			}  &
			\multicolumn{1}{c}{
				\begin{overpic}[width=0.15\linewidth]{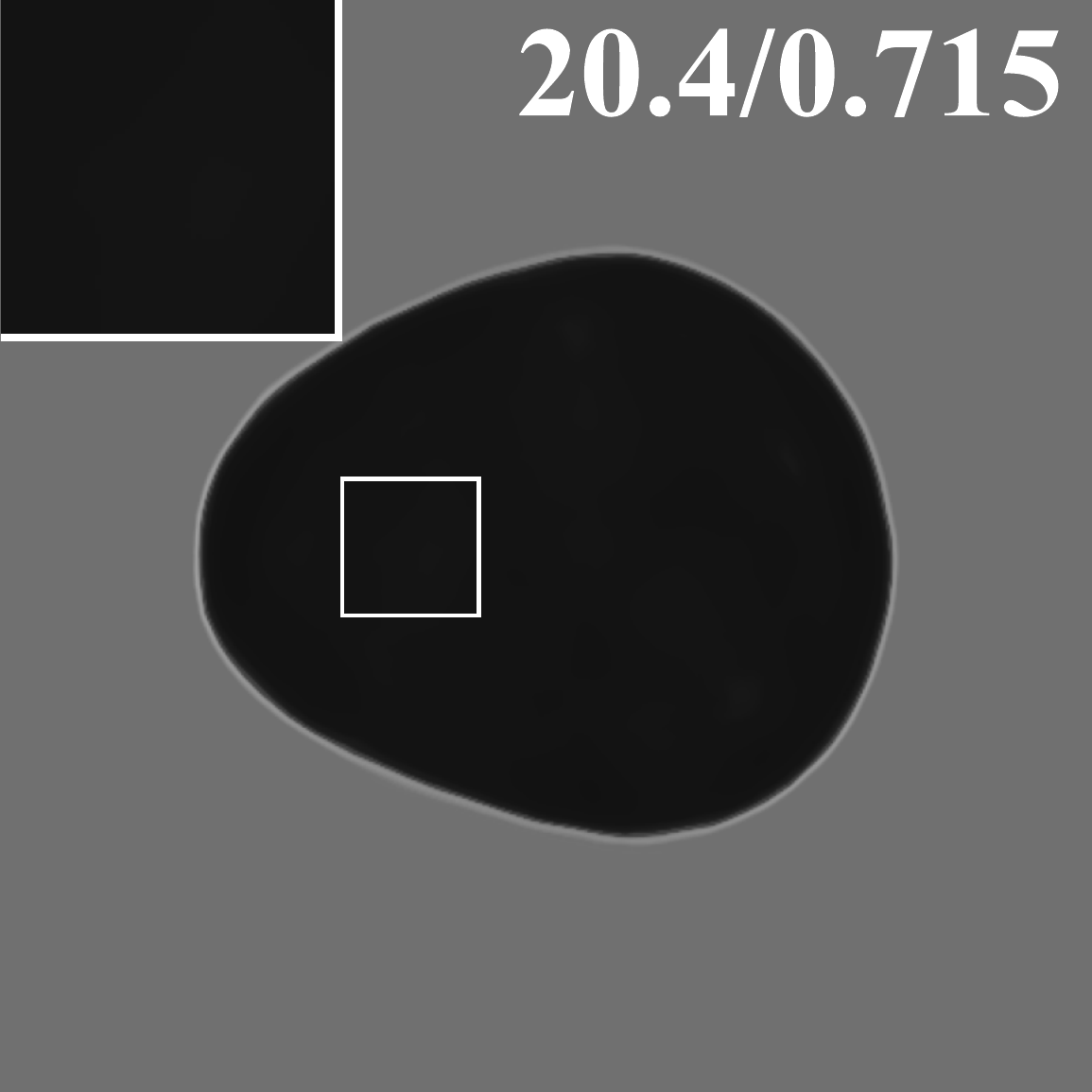}
					
				\end{overpic}
			}

			  \\
			\begin{turn}{90}  \quad\quad \small{EXD} \end{turn} & 
			  \multicolumn{1}{c}{
			  	\begin{overpic}[width=0.15\linewidth]{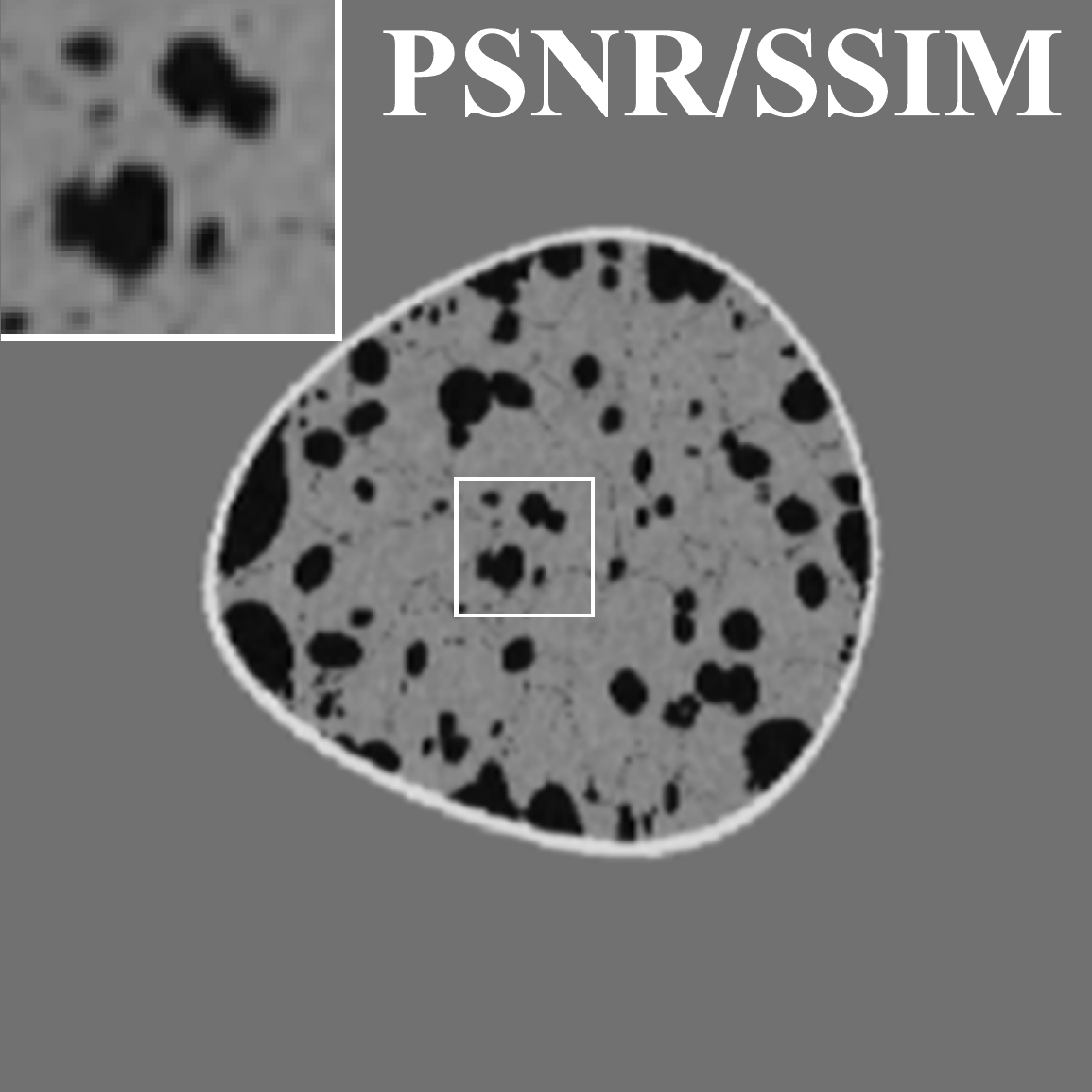}
			  	\end{overpic}
			  }  &
			  \multicolumn{1}{c}{
			  	\begin{overpic}[width=0.15\linewidth]{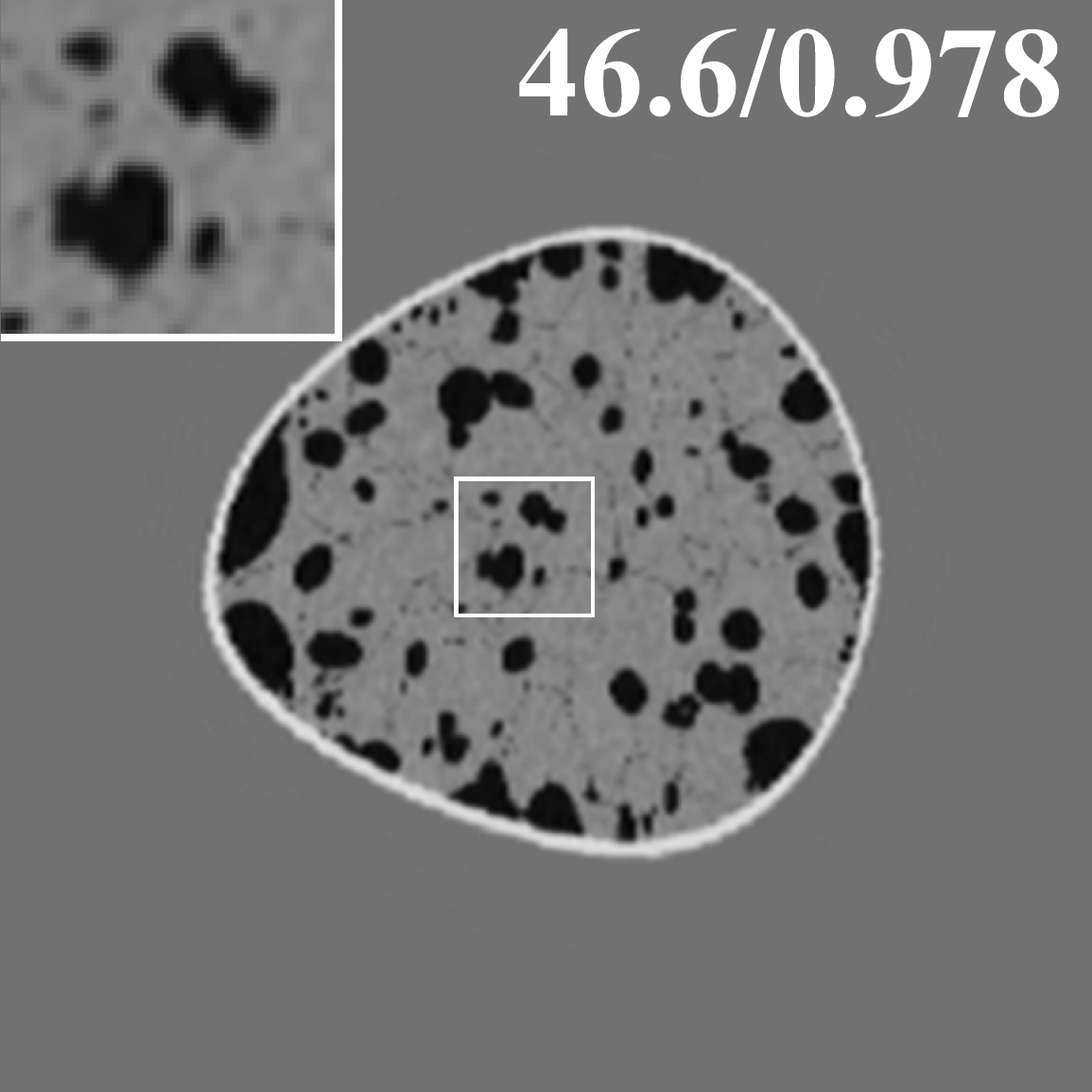}
			  	\end{overpic}
			  }  &
			  \multicolumn{1}{c}{
			  	\begin{overpic}[width=0.15\linewidth]{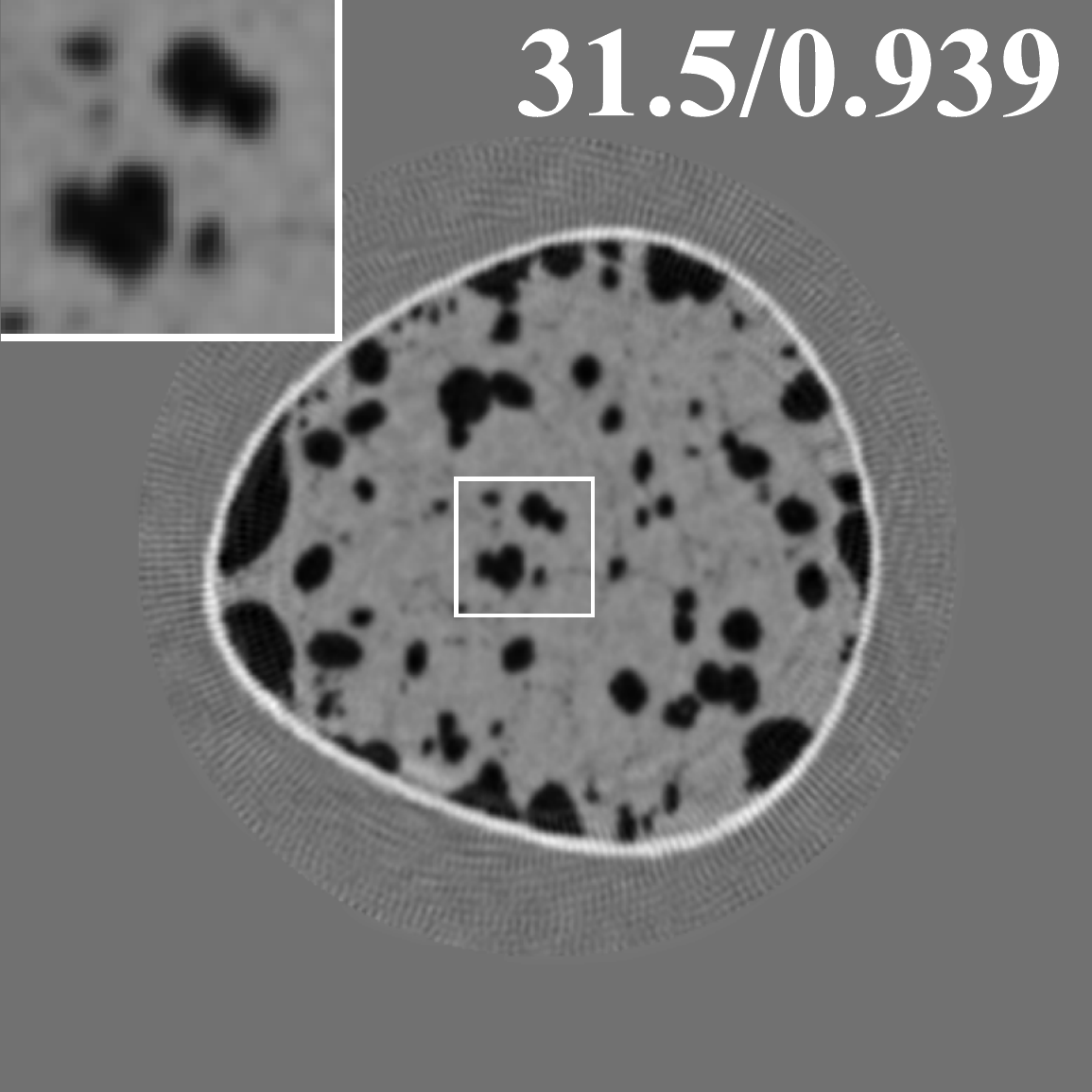}
			  	\end{overpic}
			  }  &
			  \multicolumn{1}{c}{
			  	\begin{overpic}[width=0.15\linewidth]{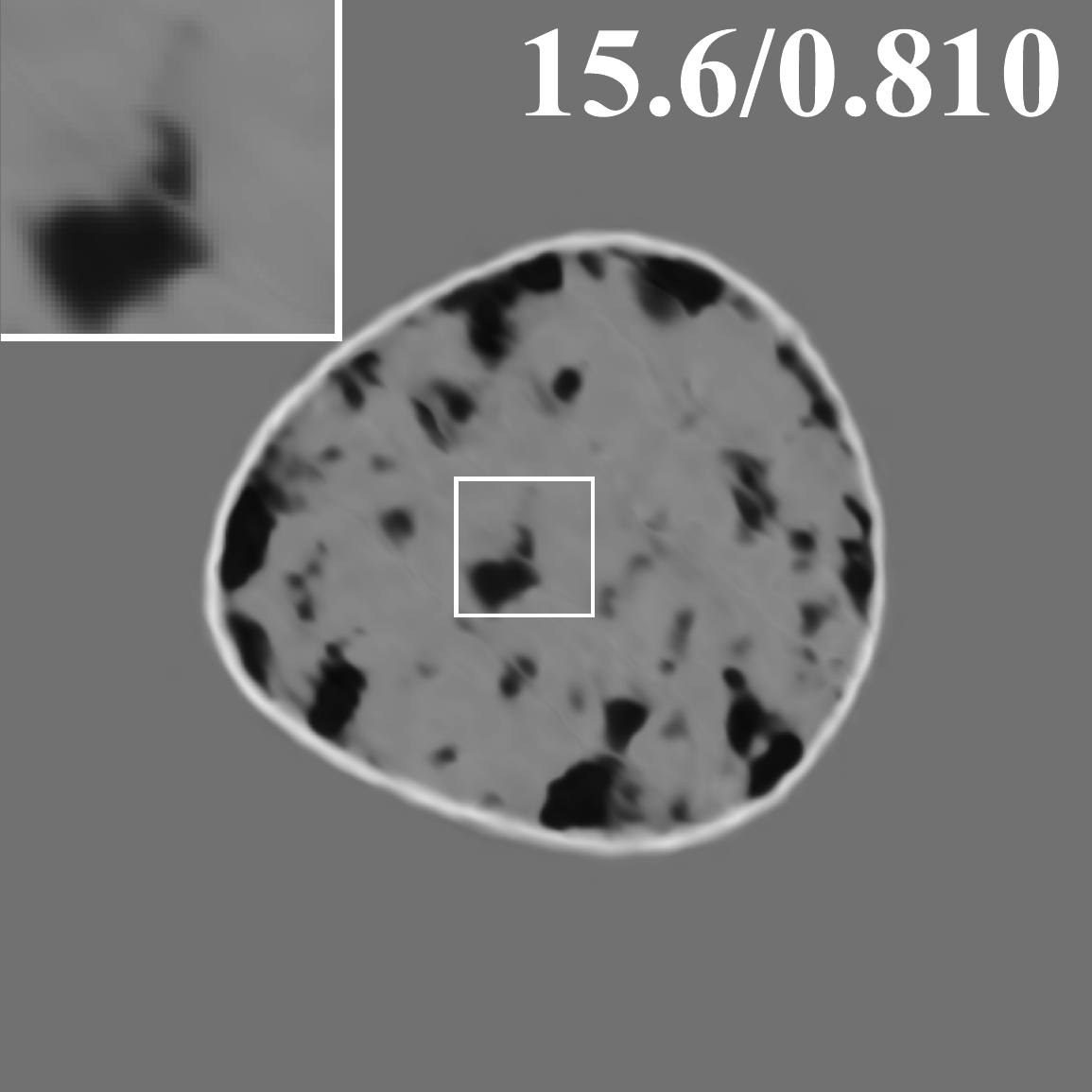}
			  		
			  	\end{overpic}
			  }  &
			  \multicolumn{1}{c}{
			  	\begin{overpic}[width=0.15\linewidth]{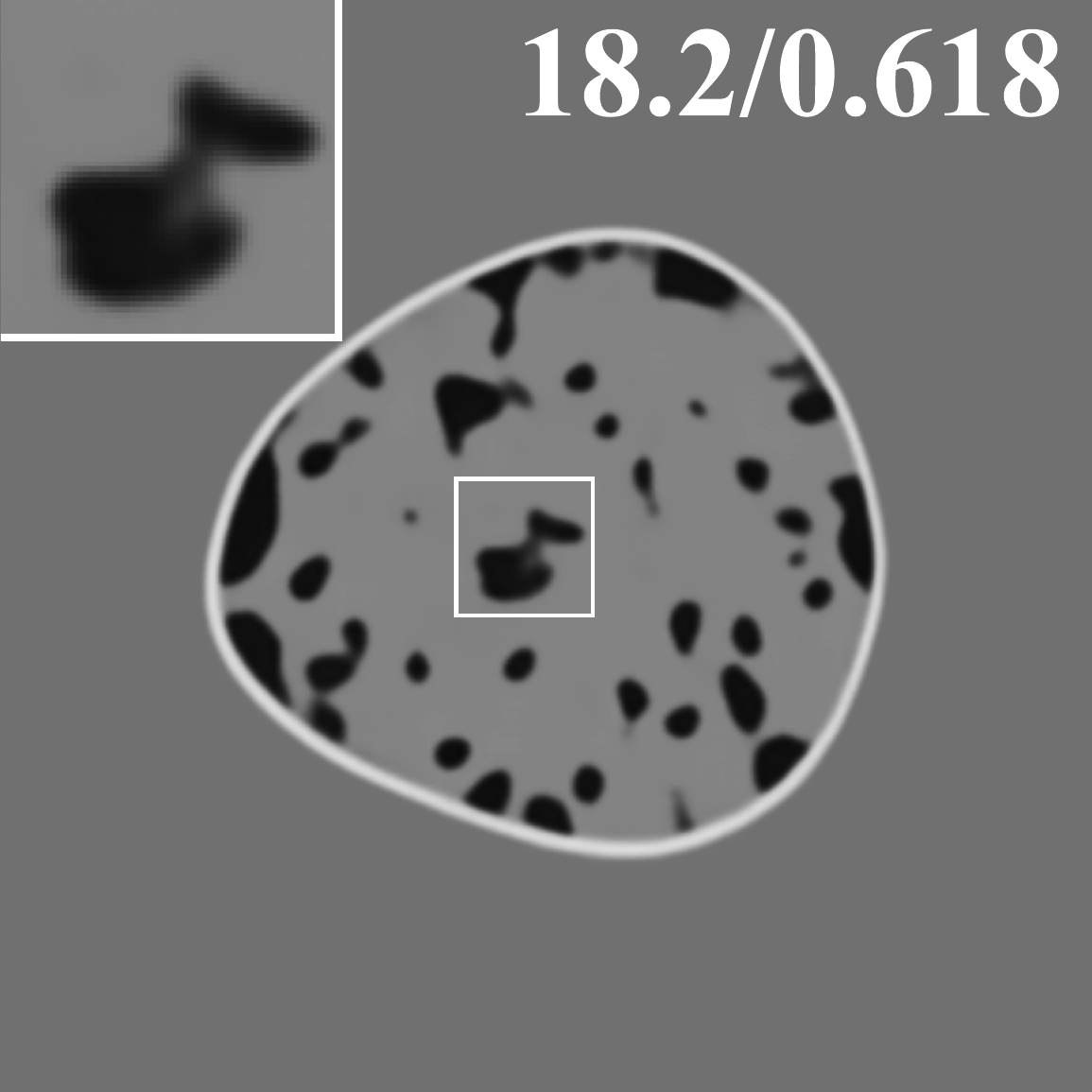}
			  		
			  	\end{overpic}
			  }  &
			  \multicolumn{1}{c}{
			  	\begin{overpic}[width=0.15\linewidth]{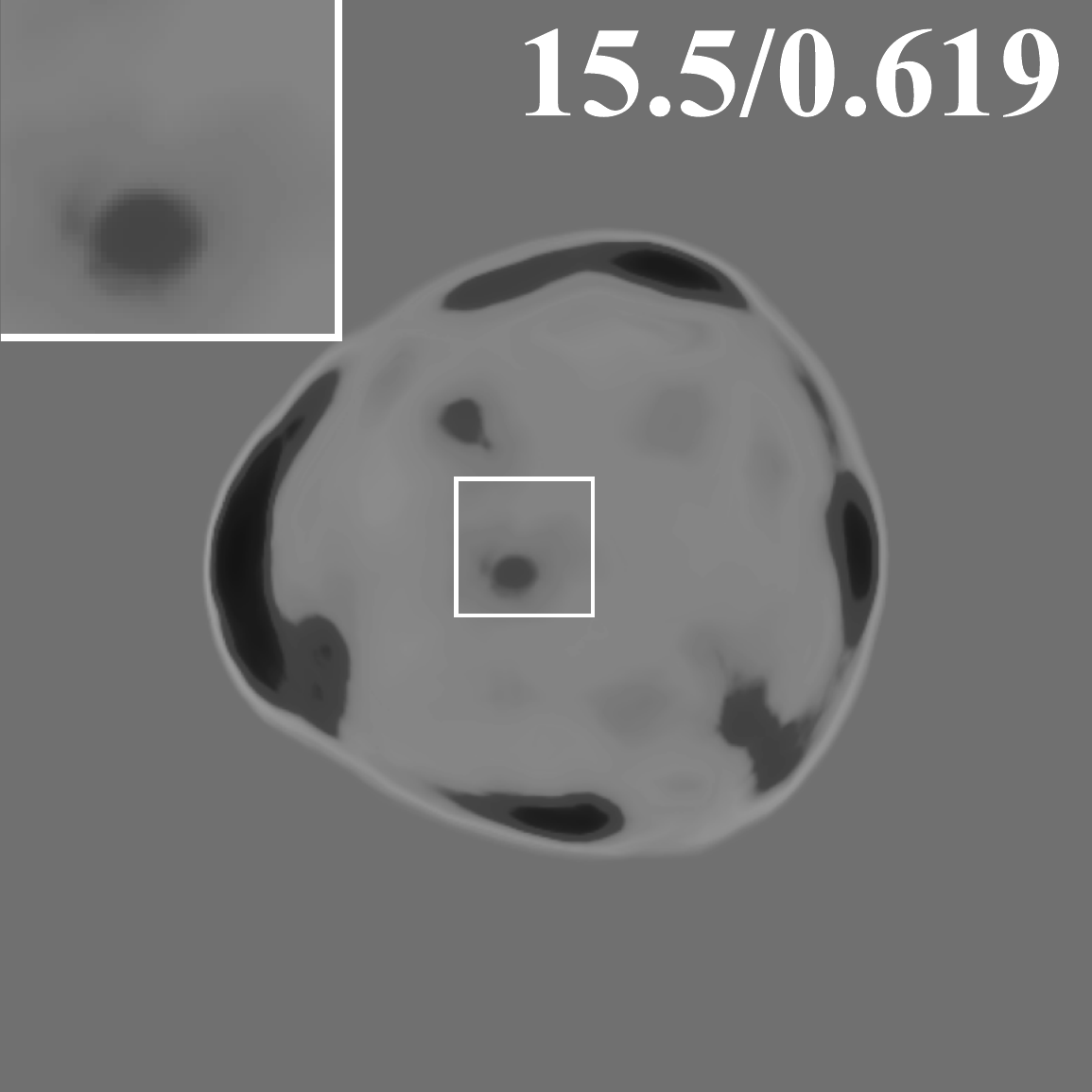}
			  		
			  	\end{overpic}
			  }  
		\end{tabular}
	\end{tabular}
\caption{\textbf{Inverse imaging results.} Comparison of reconstructed breast sound speeds for four breast types using three direct inversion baselines and an optimization-based method with FNO surrogate. Results from gradient-based optimization with a numerical solver (CBS) are provided as a reference.}
\label{fig:inverse_main}
\end{figure*}
\begin{table*}[!hbtp]
\centering
\begin{tabular}{c|ccc|ccccc}
\hline
\multirow{3}{*}{\textbf{Metric}} & \multicolumn{4}{c}{\textbf{Models}}                                             \\ \cline{2-9}
                                           & \multirow{2}{*}{\textbf{DeepONet}} & \multirow{2}{*}{\textbf{InversionNet}} &  \multirow{2}{*}{\textbf{NIO}} & \multicolumn{5}{c}{\textbf{Gradient-based  Optimization Method}} \\ \cline{5-9}
                                           &                                    &                                        &                                &  \textbf{UNet} &\textbf{FNO}  & \textbf{AFNO} & \textbf{BFNO} & \textbf{MgNO}  \\ \hline
PSNR$\uparrow$                             &    16.65                           &    20.26                               &    18.06                       &    20.02      &  25.84  &  24.85     & \underline{27.91}       &  \textbf{30.48}     \\
SSIM$\uparrow$                             &    0.8572                          &    0.8640                              &    0.8692                      &    0.8674     &  0.9104 &  0.8506    & \underline{0.9193}      &  \textbf{0.9381} \\ \hline
\end{tabular}
\caption{\textbf{Quantitative evaluation of inverse imaging baselines.} Performance was evaluated on the test set using PSNR \& SSIM. \textbf{Bold}: Best, \underline{Underlined}: Second Best.}
\label{tab:inverse-baselines}
\end{table*}

Table~\ref{tab:inverse-baselines} and Fig.~\ref{fig:inverse_main} present the wave imaging performance of different methods across four breast types. Although all direct inversion models accurately reconstructed breast size and boundaries and correctly identified breast type, DeepONet’s reconstructions entirely lacked information on interior structure.  Notably, NIO outperformed DeepONet on all breast categories, demonstrating the strength of the global modeling capability provided by the Fourier layer. InversionNet also achieved much better results compared to DeepONet, indicating that convolution-based networks are well-suited for complex image reconstruction tasks. The neural operator-based optimization approach revealed significantly higher resolution than all direct inversion methods, although it incurs higher costs due to the iterative descent process (still much faster than traditional iterative reconstruction with numerical solvers). This suggests that the forward operators better capture the underlying wave physics, while direct inversion pipelines may overly rely on memorizing prior knowledge about the anatomy of the training breasts. Mathematically, the forward neural operatos only need to learn the conditional mapping 
$p(y|x)$, whereas direct inversion networks must approximate $p(x|y) \propto p(y|x)p(x)$, simultaneously modeling both the likelihood and the prior. This dual requirement makes inversion methods more prone to overfitting the data distribution $p(x)$ rather than faithfully encoding the underlying physics. The combination of forward neural operators and gradient-based optimization is consequently more robust than direct inversion networks. In practical FWI applications, it's crucial to balance reconstruction accuracy and computational efficiency.

\subsection{Validation on \textit{in vivo} human breast dataset} 
\label{subsec:clinical data}
To demonstrate the practical applicability of our dataset, we evaluated the performance of two forward surrogate models (BFNO and FNO) and one direct inversion model (NIO), all trained on the OpenBreastUS dataset, on reconstructing \textit{in vivo} human breasts. The clinical datasets were obtained from the Karmanos Cancer Institute under Institutional Review Board approval No. 040912M1F \cite{ali20242}, with detailed descriptions provided in Appendix ~\ref{app:clinic}. Fig.~\ref{subfig:clinical-ground-truth} shows the reconstruction of a breast with a malignant tumor using a 2D FDFD solver with data from 20 frequencies (0.3–1.25MHz) and 1024 sensors, as reported in \cite{ali20242}.  This result serves as a reasonable ground truth for our experiments. Since our neural operators were trained on data at 8 frequencies (300–650 kHz), we evaluated the baseline models using clinical data restricted to these 8 frequencies to ensure a fair comparison.
\begin{figure*}[h]
    \centering
    {%
    \footnotesize
    % 子图 (a)
    \subfloat[]{%
        \includegraphics[width=0.23\linewidth]{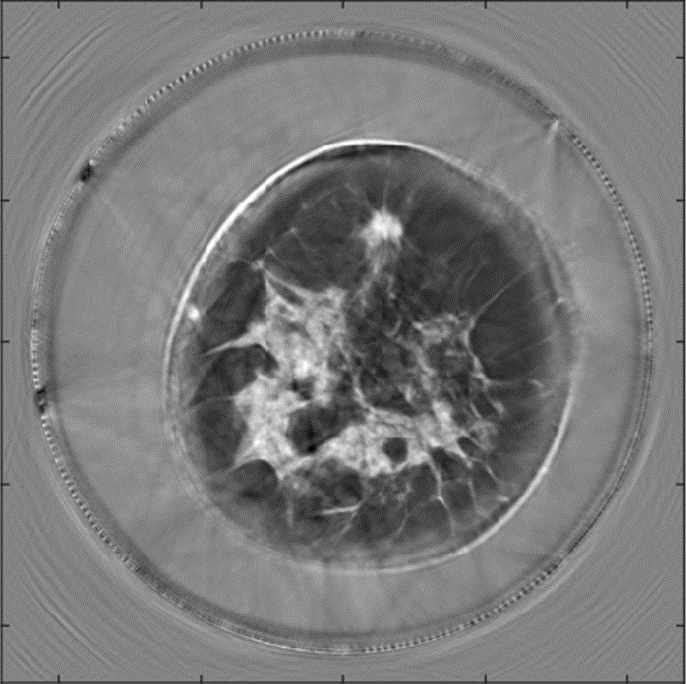}%
        \label{subfig:clinical-ground-truth}%
    }\hfill
    % 子图 (b)
    \subfloat[]{%
        \includegraphics[width=0.23\linewidth]{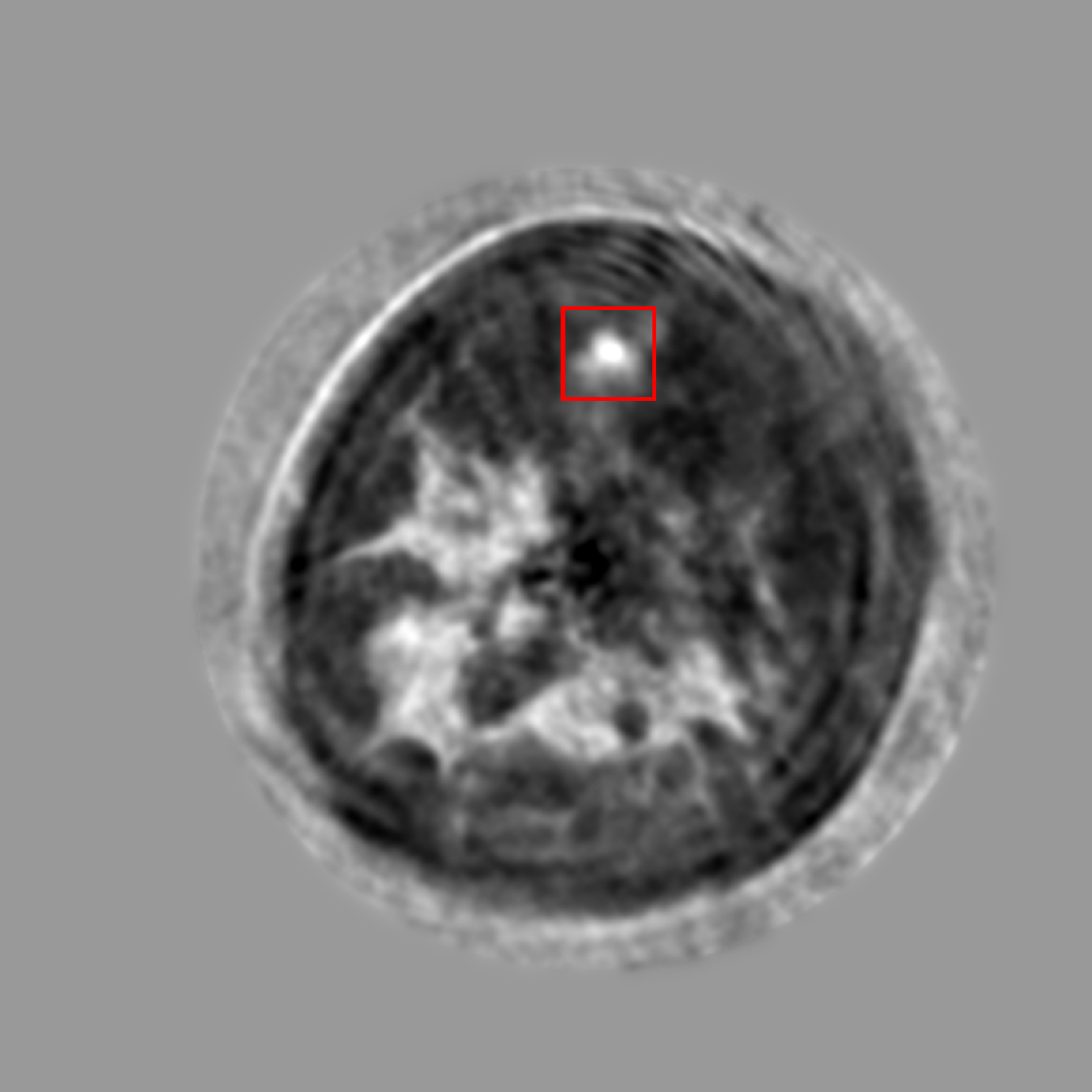}%
        \label{subfig:clinical-bfno}%
    }\hfill
    % 子图 (c)
    \subfloat[]{%
        \includegraphics[width=0.23\linewidth]{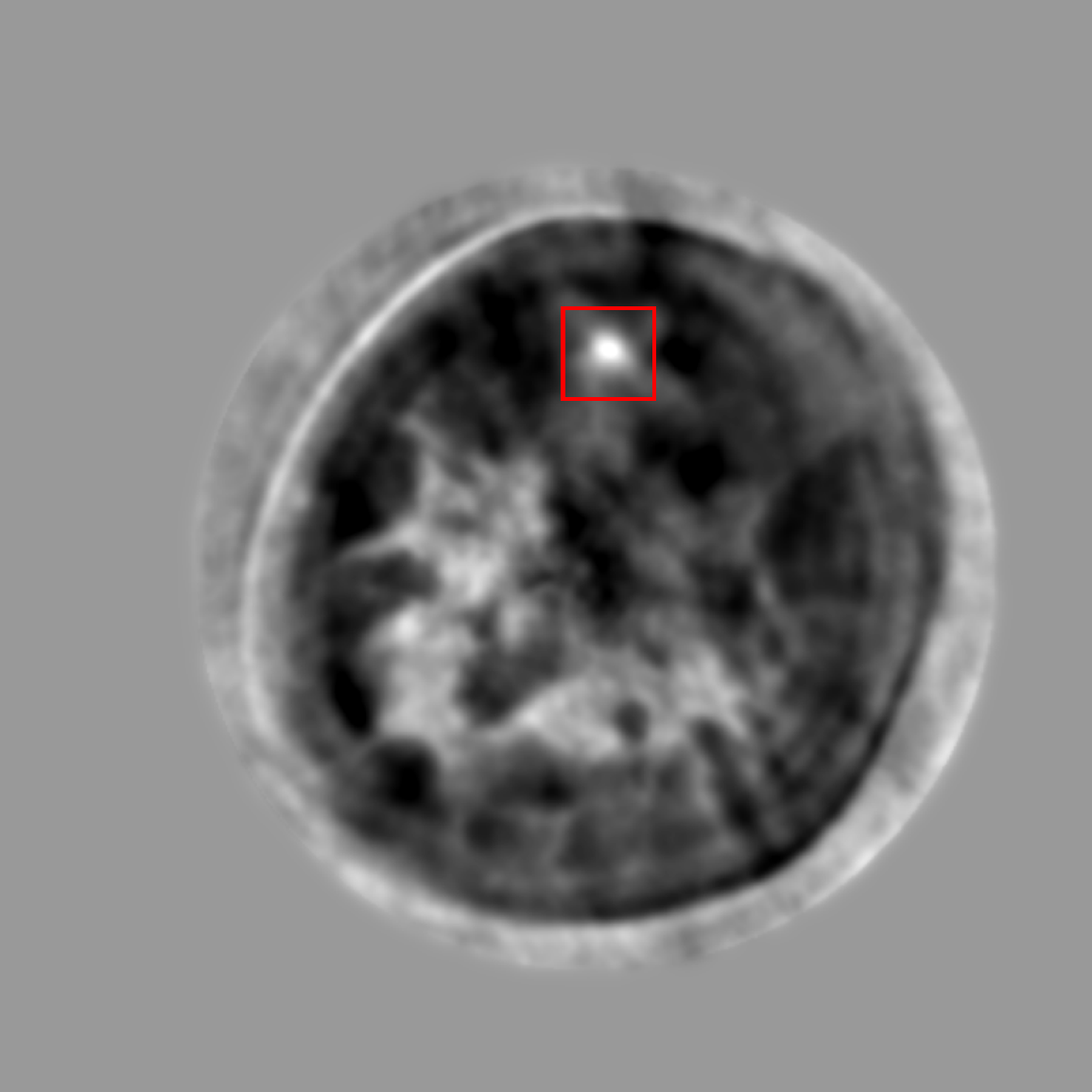}%
        \label{subfig:clinical-fno}%
    }\hfill
    % 子图 (d)
    \subfloat[]{%
        \includegraphics[width=0.23\linewidth]{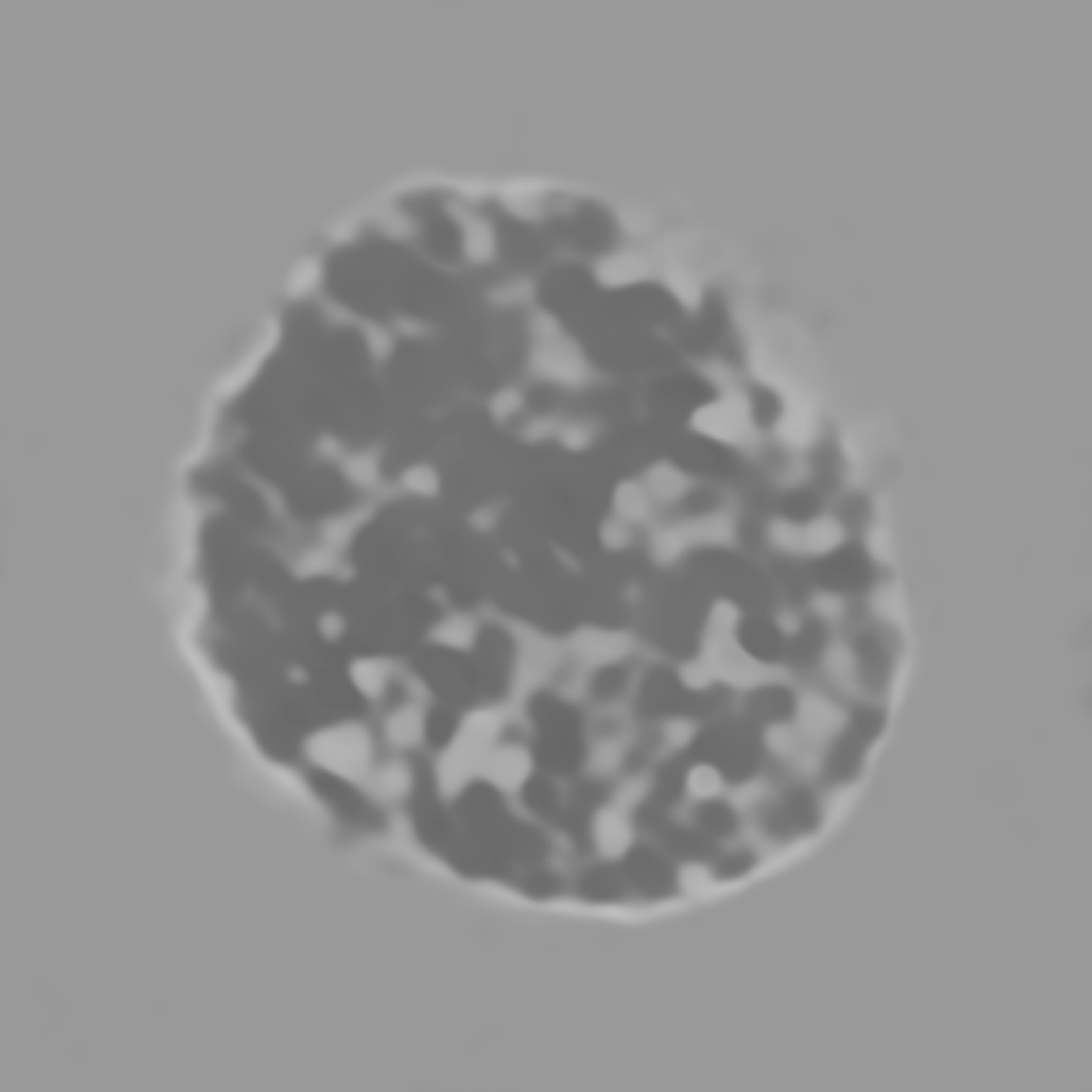}%
        \label{subfig:clinical-nio}%
    }
    }% 结束局部组
    \vspace{-0.1cm}
    \caption{\textbf{Validation on reconstructing \textit{in vivo} human breast with a malignancy using different models.} 
    The reconstruction results on the clinical USCT dataset, using forward models and a direct inversion model trained on OpenBreastUS. 
    (a) Ground Truth: a breast phantom reconstructed using an FDFD solver with 20 frequencies’ data. 
    (b) BFNO with 8 frequencies and 256 transmitters. 
    (c) FNO with 8 frequencies and 256 transmitters. 
    (d) NIO inversion. 
    The malignant tumor is highlighted in the red box.}
    \label{fig:clinical-data-subfloat}
\end{figure*}
As shown in Fig.~\ref{fig:clinical-data-subfloat}, the BFNO model successfully reconstructed a clear malignant tumor along with other tissue structures, such as skin, fat, and glands. These results demonstrate a strong alignment between the anatomical structures in the OpenBreastUS dataset and real breast tissues. The model architecture significantly impacts imaging quality: the FNO model struggled to capture high-frequency details, highlighting its limitations in learning such features. The direct inversion model, NIO, produced images with entirely incorrect structures, revealing its poor generalization to unseen structures and inability to capture the underlying physics compared to gradient-based optimization approaches. 

We also conducted ablation studies to assess the impact of observation settings. As shown in Supplementary Information Fig.~1, reducing the number of sensors or frequencies significantly degrades the quality of reconstructed images and introduces artifacts in BFNO-based FWI reconstruction. This behavior aligns with the widely-observed limitations in FWI reconstruction using classical numerical solvers.

%The sensor placement for this dataset differs slightly from the malignant tumor measurements, as illustrated in Fig.~\ref{subfig:belign-source}.

% In addition to the malignant tumor case, we evaluated our approach using a \textit{in vivo} breast dataset with a benign cyst.   Fig.~\ref{subfig:belign-gt} presents the reference reconstruction obtained using a 2D FDFD solver and 20-frequency data from \cite{ali20242}. As shown in Fig.~\ref{subfig:belign-bfno} and \ref{subfig:belign-fno}, both BFNO and FNO successfully reconstruct the breast structure, including a clearly defined benign cyst, demonstrating that our dataset closely aligns with the characteristic distribution of real breast tissue and that the models generalize well across different instrument configurations (e.g., source locations). However, the results show a noticeable decline in fine structural details and resolution compared to the malignant tumor case. Moreover, the FNO reconstruction exhibits greater feature loss and lower overall quality, underscoring its weaker generalization of high-frequency features relative to BFNO.

\begin{figure*}[htbp]
    \centering
    {%
    \footnotesize
    % 子图 (a)
    \subfloat[]{%
        \includegraphics[width=0.23\linewidth]{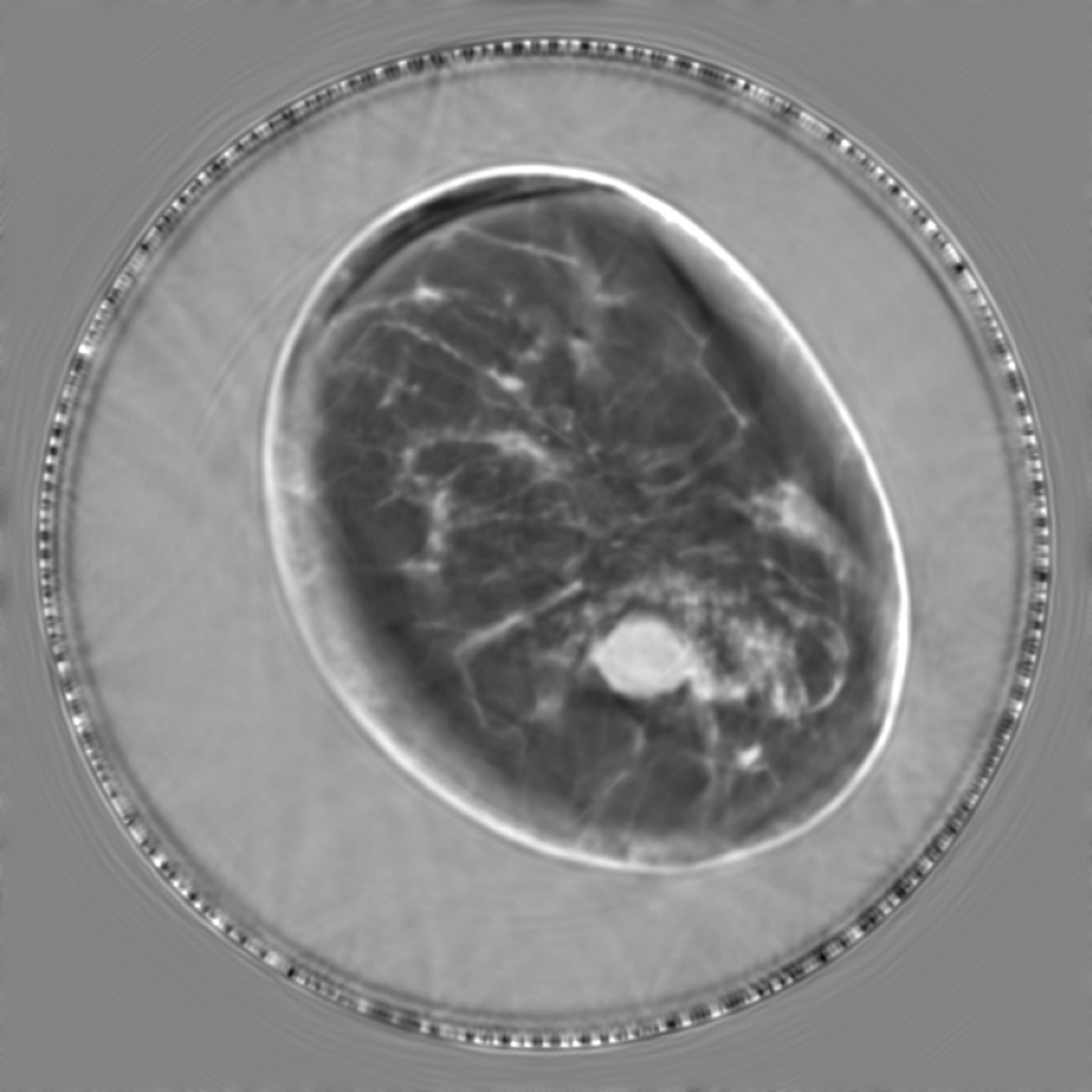}%
        \label{subfig:belign-gt}%
    }\hfill
    % 子图 (b)
    \subfloat[]{%
        \includegraphics[width=0.23\linewidth]{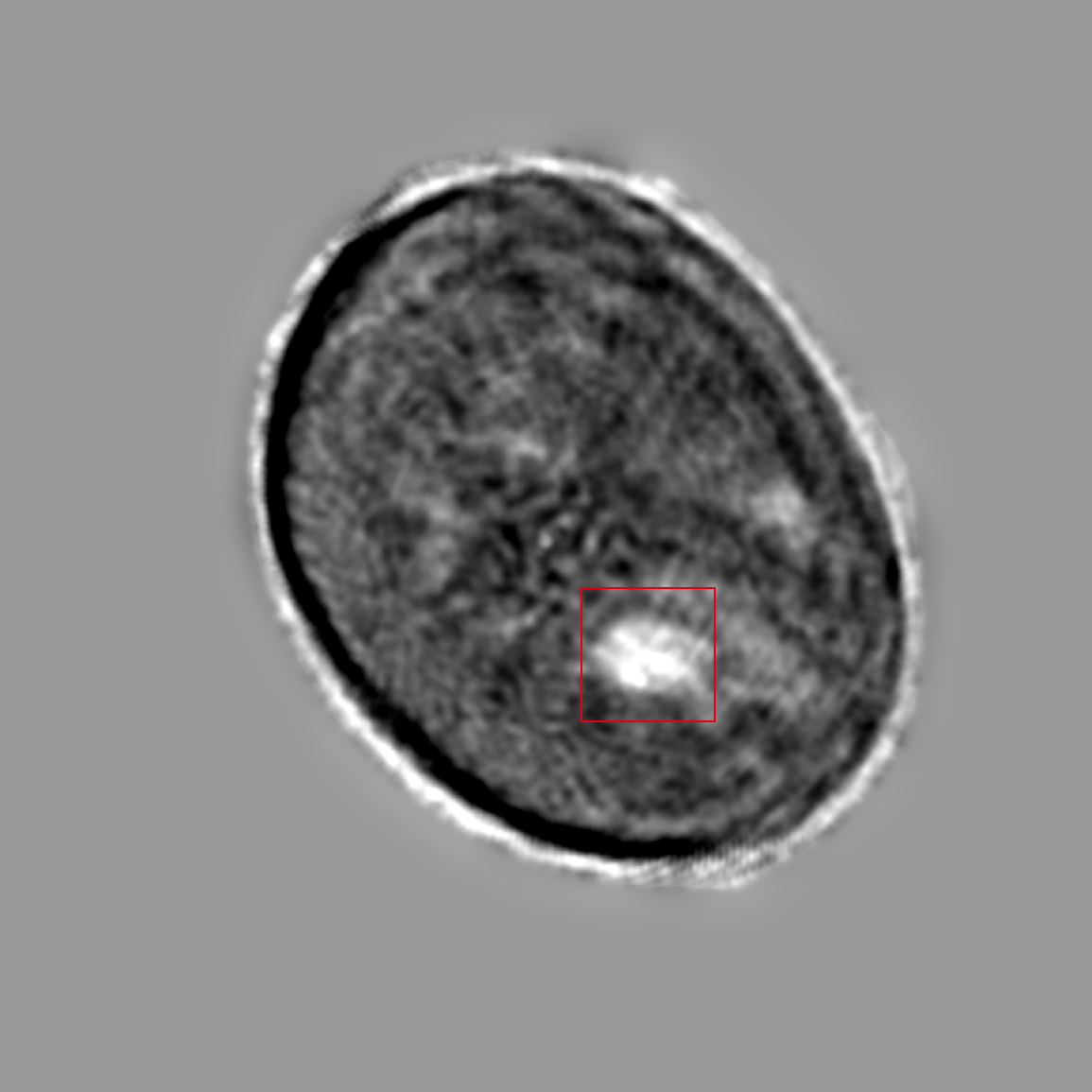}%
        \label{subfig:belign-bfno}%
    }\hfill
    % 子图 (c)
    \subfloat[]{%
        \includegraphics[width=0.23\linewidth]{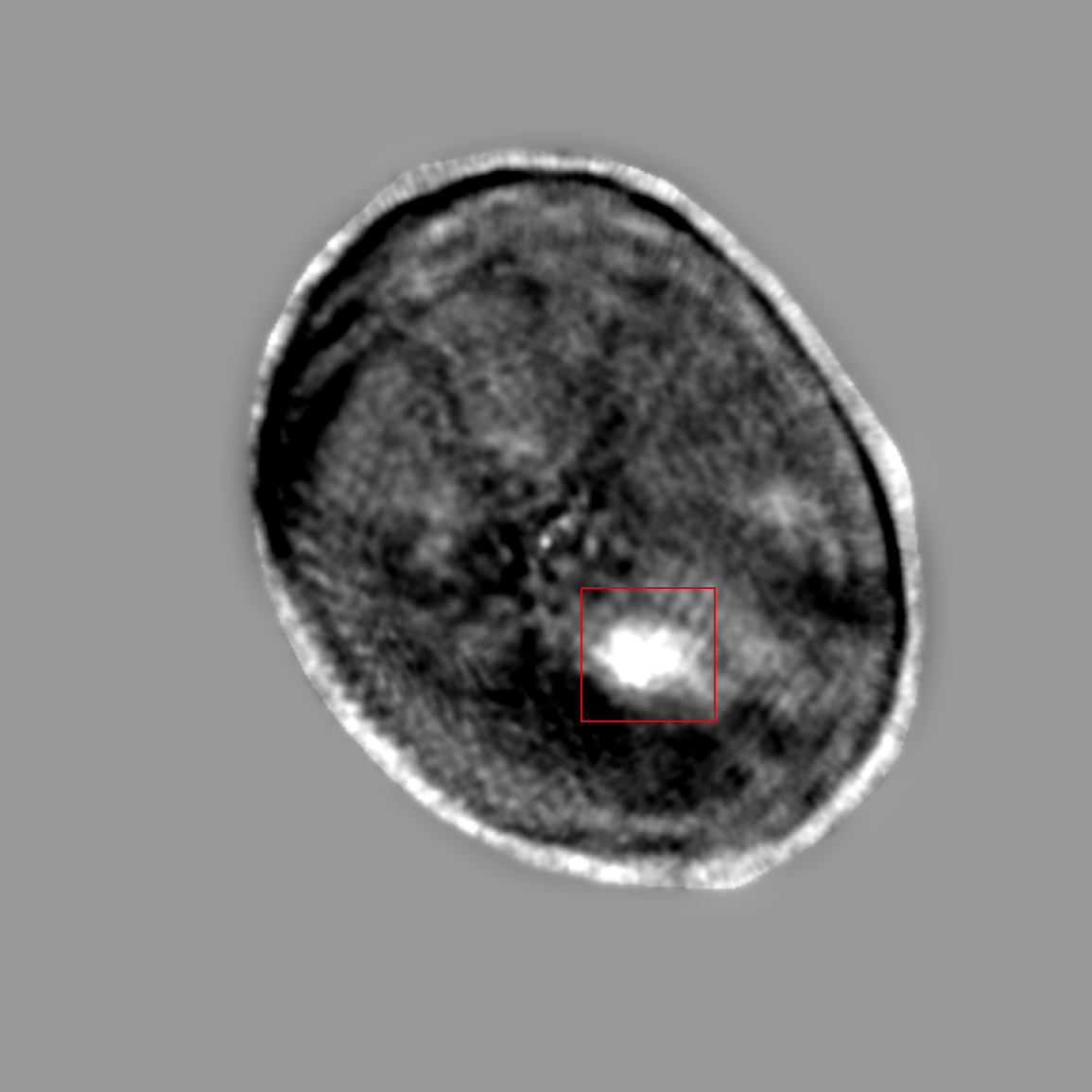}%
        \label{subfig:belign-fno}%
    }\hfill
    % 子图 (d)
    \subfloat[]{%
        \includegraphics[width=0.23\linewidth]{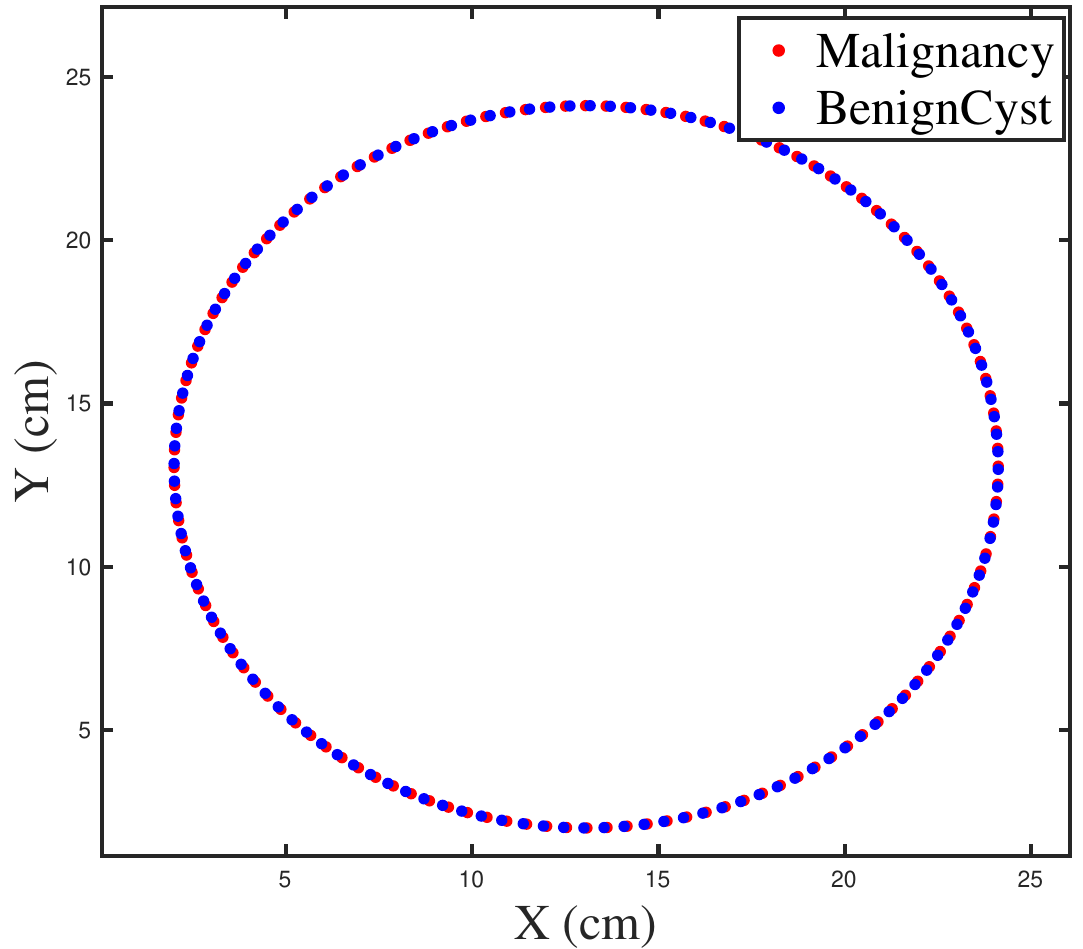}%
        \label{subfig:belign-source}%
    }
    }% 结束局部组
    \caption{\textbf{Reconstruction results of \textit{in vivo} human breast with a benign cyst using different models.} 
    The reconstruction results on the clinical USCT dataset, using forward models and a direct inversion model trained on OpenBreastUS. 
    (a) Ground Truth: reconstruction of a breast phantom using an FDFD solver with 20 frequencies’ data. 
    (b) BFNO with 8 frequencies and 256 transmitters. 
    (c) FNO with 8 frequencies and 256 transmitters. 
    (d) Source location of imaging data with a malignant lesion and imaging data with a benign cyst.}
    \label{fig:clinical data_B-minipage}
\end{figure*}

In addition to the malignant‐tumor case, we assessed our method on a \textit{in vivo} breast dataset with a benign cyst. Fig~\ref{subfig:belign-gt} shows the reference reconstruction produced by a two‐dimensional finite‐difference frequency‐domain (FDFD) solver using 20‐frequency data as reported in \cite{ali20242}. Figure ~\ref{subfig:belign-bfno} and \ref{subfig:belign-fno} demonstrate that both the BFNO and FNO accurately recover the breast anatomy, with the benign cyst sharply delineated. Crucially, gradient‐based optimization with neural surrogates discriminates malignant masses—marked by irregular boundary morphology—from benign lesions, which exhibit smooth, regular boundaries, thereby highlighting our approach’s potential for efficient and precise breast‐disease detection. Moreover, as illustrated in Fig.~\ref{subfig:belign-source}, the experimental imaging configuration (e.g., transducer positions) differed from the simulated setup, confirming that our models generalize across diverse instrument arrangements.

\subsection{Additional Analysis} \label{subsec:analysis}
\subsubsection{Data Complexities of Different Breast Types}
 Different breast categories have distinct internal structures, leading to significant variations in sound speed distribution and scattering effects within the tissue. As observed in Fig.~\ref{fig:forward-300k}, the heterogeneous and extremely-dense  breasts exhibit the most complex structures and the strongest scattering due to their higher densities, while the fibroglandular and fatty breasts show the weakest scattering. Figure~\ref{fig:forward-errors} and \ref{fig:inverse-errors} in Appendix present the forward prediction accuracy and inversion quality of various neural operators across all breast categories. Notably, heterogeneous and extremely dense breasts exhibit higher errors, whereas fibroglandular and fatty breasts are more readily modelled, resulting in comparatively lower errors.

\subsubsection{Data Complexities of Different Frequencies}
\begin{figure*}[htbp]
    \centering
    {%
    \footnotesize
    % 子图 (a)
    \subfloat[]{%
        \includegraphics[width=0.32\linewidth]{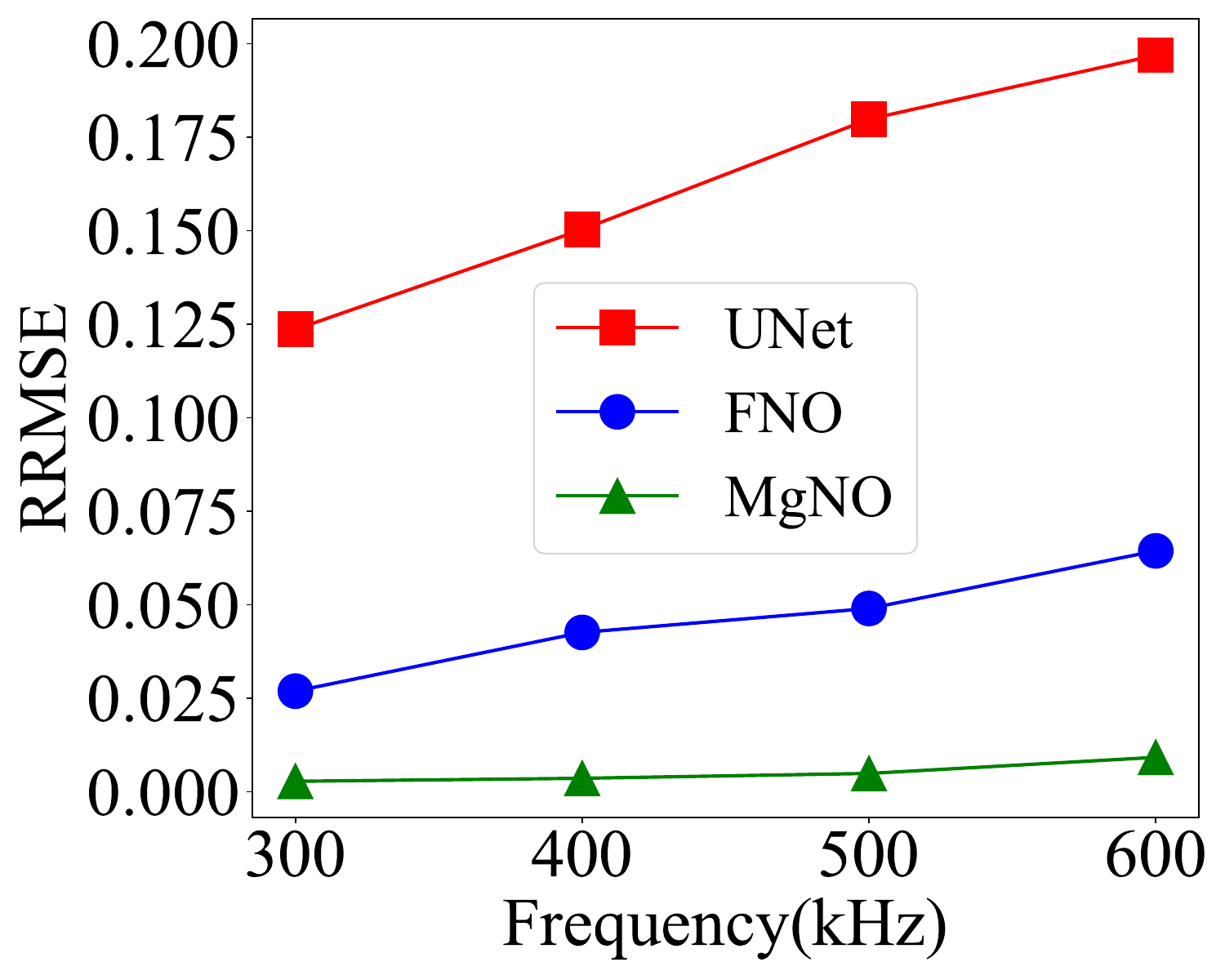}%
        \label{subfig:freq_ablation}%
    }\hfill
    % 子图 (b)
    \subfloat[]{%
        \includegraphics[width=0.32\linewidth]{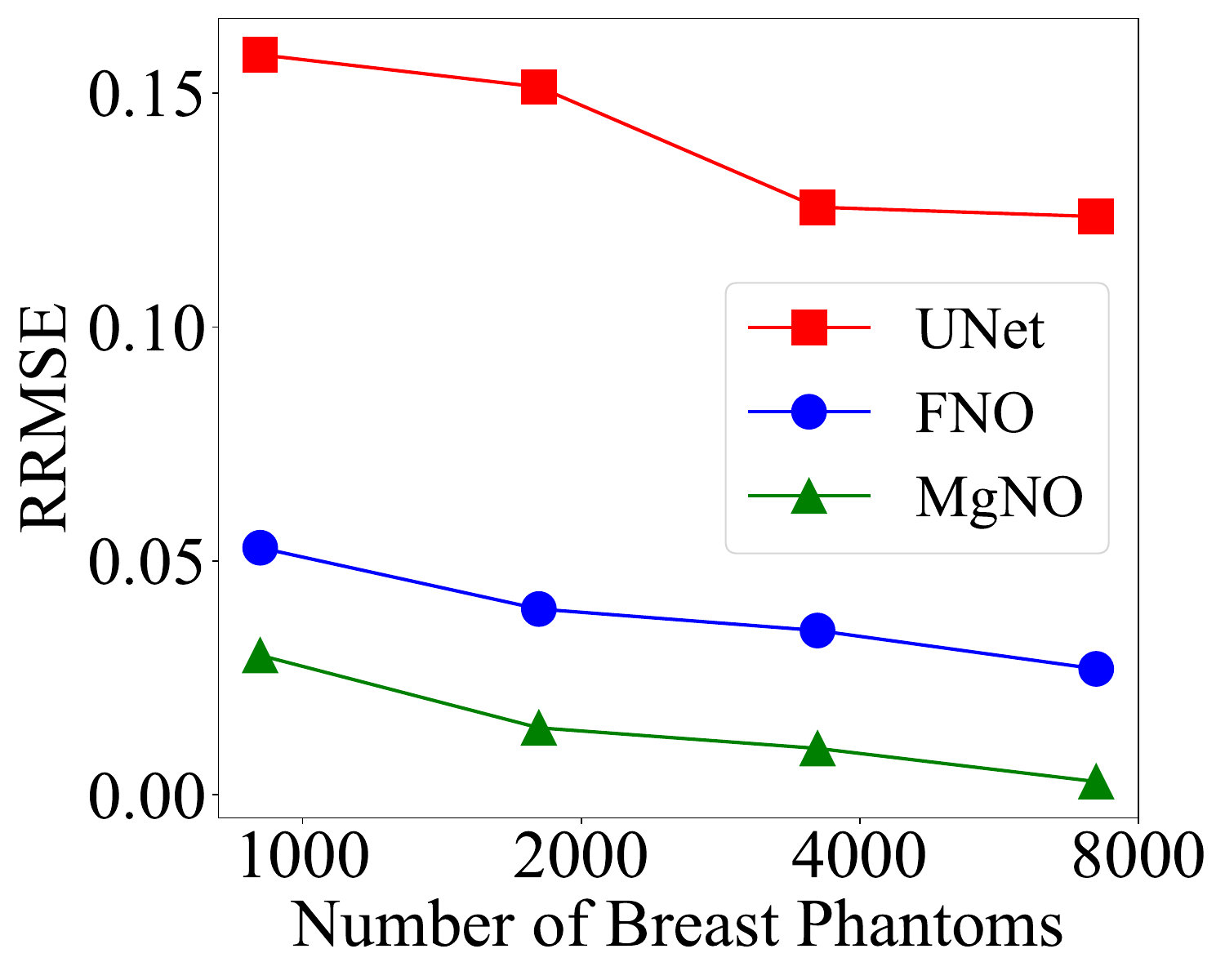}%
        \label{subfig:num_breast_ablation}%
    }\hfill
    % 子图 (c)
    \subfloat[]{%
        \includegraphics[width=0.32\linewidth]{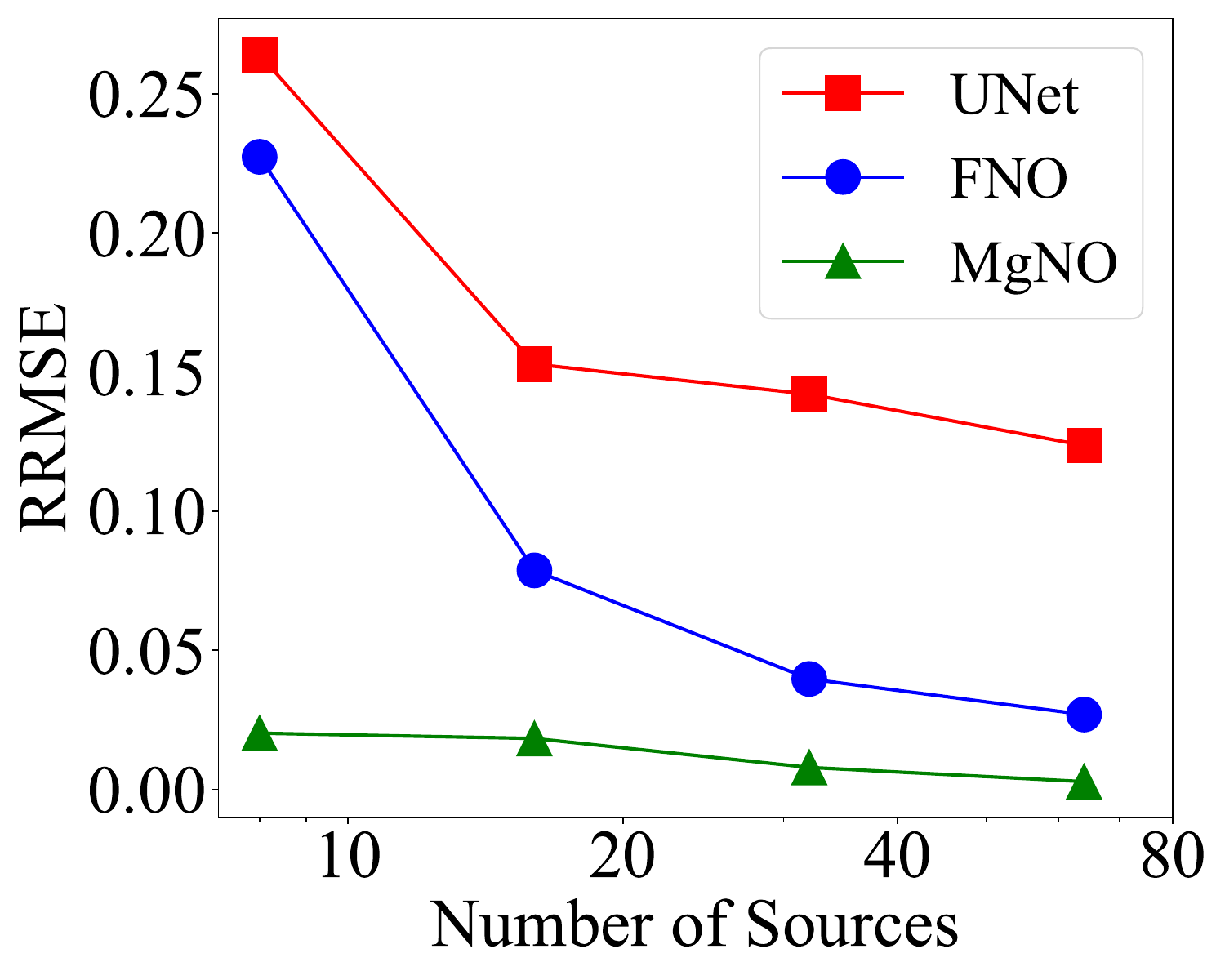}%
        \label{subfig:num_source_ablation}%
    }
    }% 结束局部组
    \caption{\textbf{Analysis of Data Complexity, Model Scalability, and Generalization.} (a) RRMSE variation of neural operators trained on data at different frequencies. (b) RRMSE variation of neural operators trained with different numbers of breast phantoms. (c) RRMSE variation of neural operators trained with different numbers of source locations.}
    \label{fig:scaling-law}
\end{figure*}

From a theoretical perspective, \cite{engquist2018approximate} indicates that higher frequencies are known to increase the complexity of solution operators. Numerically, solving high-frequency problems typically requires more precise methods and denser grid points. All neural operators experience performance degradation when learning wavefields at higher frequencies, as shown in Fig.~\ref{subfig:freq_ablation}.  Among all baselines, the UNet degrades the fastest, while the FNO and MgNO show less pronounced error increases. This suggests that incorporating global and multiscale features is crucial for achieving high-accuracy approximations in operator learning across frequencies.
\subsubsection{Scaling with Dataset Size}
Figure~\ref{subfig:num_breast_ablation} examines how the performance of different forward neural operators scales with the size of the training dataset. An increased amount of training data consistently enhances wave simulation accuracy, validating the scaling law of operator learning and underscoring the necessity of creating large-scale datasets for studying neural operator frameworks. Neural operator architectures scale differently with increasing training data. Notably, MgNO and the FNO family show continued improvement as the number of training phantoms increases from 4,000 to 8,000, demonstrating better data efficiency than UNet, which shows limited improvement with additional data.
\begin{figure*}[!hbtp]
    \centering
    % ========== 第一排: GRF 媒体 ==========
    \begin{minipage}[b]{0.16\linewidth}
        \centering
        {\footnotesize Sound Speed}\\[1mm]
        \includegraphics[width=\linewidth]{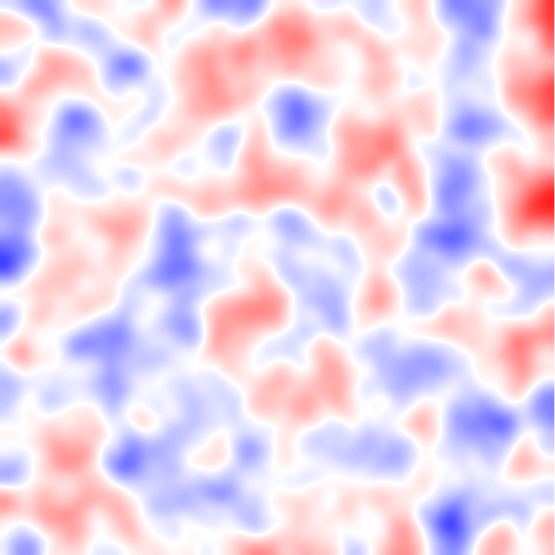}
    \end{minipage}\hfill
    \begin{minipage}[b]{0.16\linewidth}
        \centering
        {\footnotesize Ground Truth}\\[1mm]
        \includegraphics[width=\linewidth]{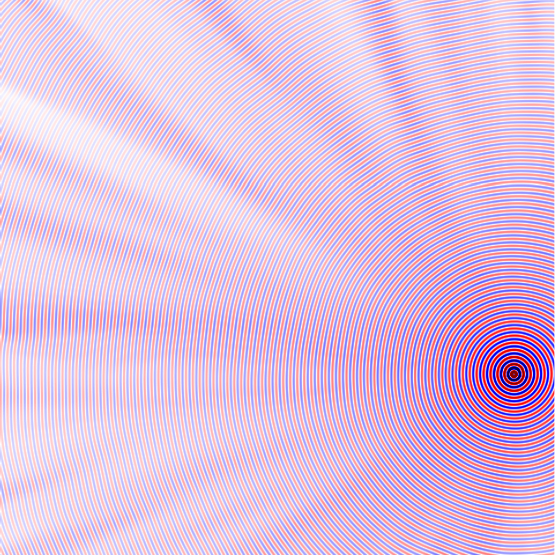}
    \end{minipage}\hfill
    \begin{minipage}[b]{0.16\linewidth}
        \centering
        {\footnotesize FNO(GRF)}\\[1mm]
        \includegraphics[width=\linewidth]{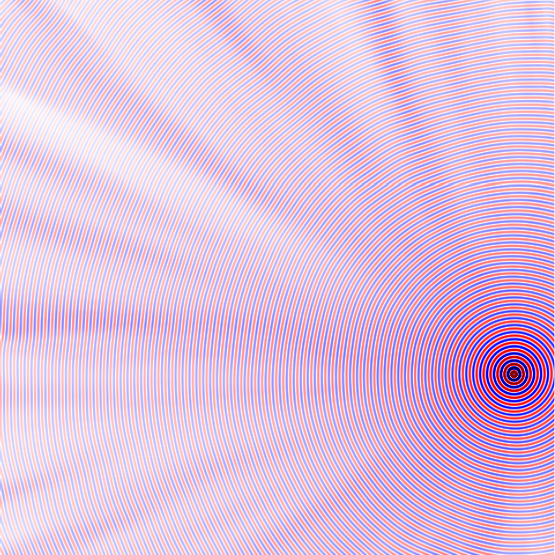}
    \end{minipage}\hfill
    \begin{minipage}[b]{0.16\linewidth}
        \centering
        {\footnotesize UNet(GRF)}\\[1mm]
        \includegraphics[width=\linewidth]{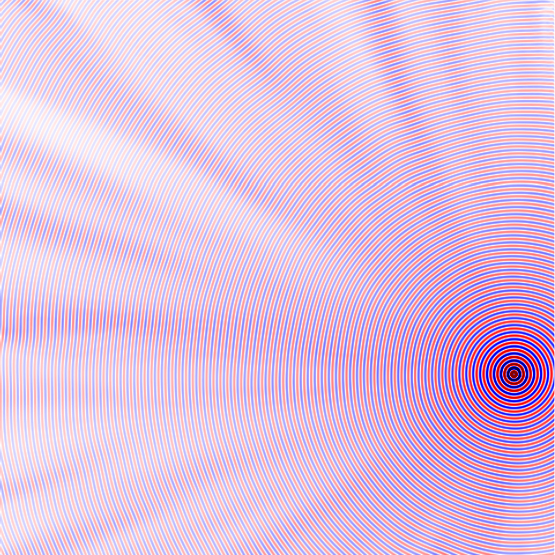}
    \end{minipage}\hfill
    \begin{minipage}[b]{0.16\linewidth}
        \centering
        {\footnotesize FNO(OpenBreastUS)}\\[1mm]
        \includegraphics[width=\linewidth]{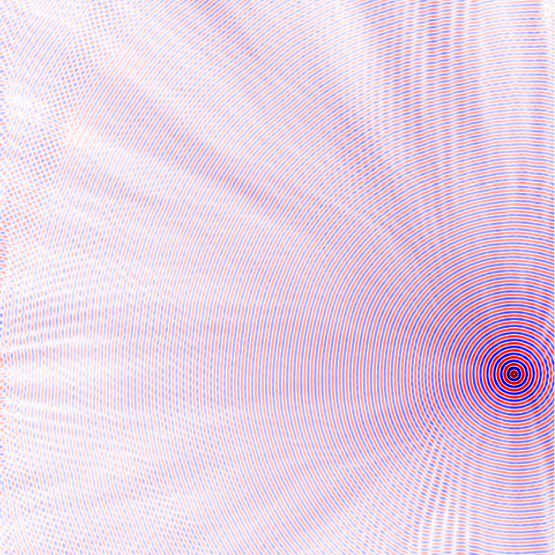}
    \end{minipage}\hfill
    \begin{minipage}[b]{0.16\linewidth}
        \centering
        {\footnotesize UNet(OpenBreastUS)}\\[1mm]
        \includegraphics[width=\linewidth]{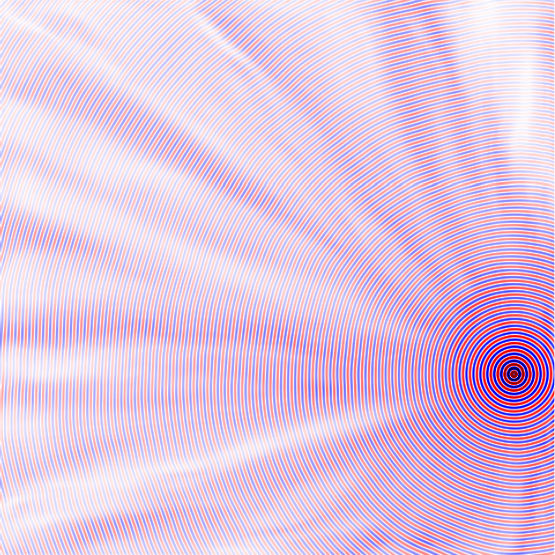}
    \end{minipage}

    \vspace{0.5em} % 第一排和第二排之间的垂直间距

    % ========== 第二排: Breast Phantom ==========
    \begin{minipage}[b]{0.16\linewidth}
        \centering
        {\footnotesize Sound Speed}\\[1mm]
        \includegraphics[width=\linewidth]{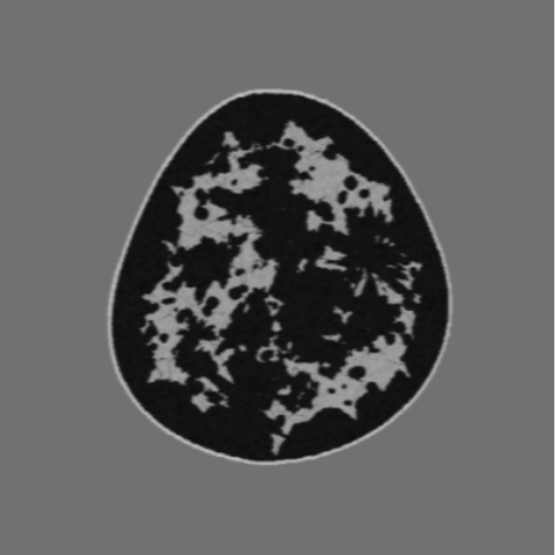}
    \end{minipage}\hfill
    \begin{minipage}[b]{0.16\linewidth}
        \centering
        {\footnotesize Ground Truth}\\[1mm]
        \includegraphics[width=\linewidth]{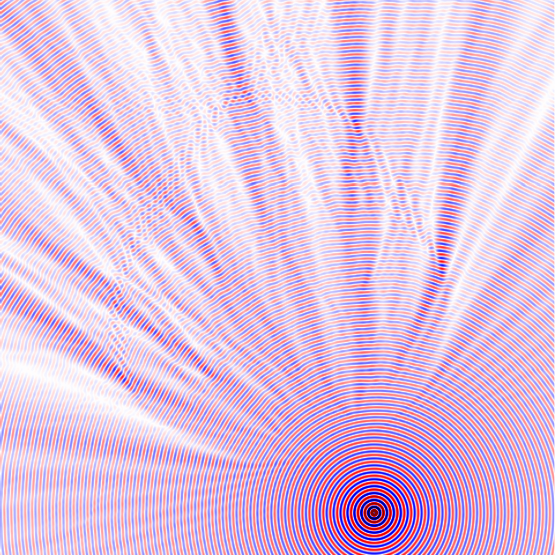}
    \end{minipage}\hfill
    \begin{minipage}[b]{0.16\linewidth}
        \centering
        {\footnotesize FNO(GRF)}\\[1mm]
        \includegraphics[width=\linewidth]{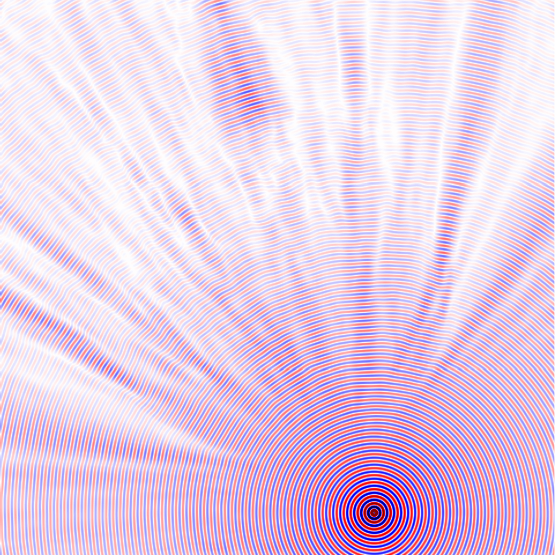}
    \end{minipage}\hfill
    \begin{minipage}[b]{0.16\linewidth}
        \centering
        {\footnotesize UNet(GRF)}\\[1mm]
        \includegraphics[width=\linewidth]{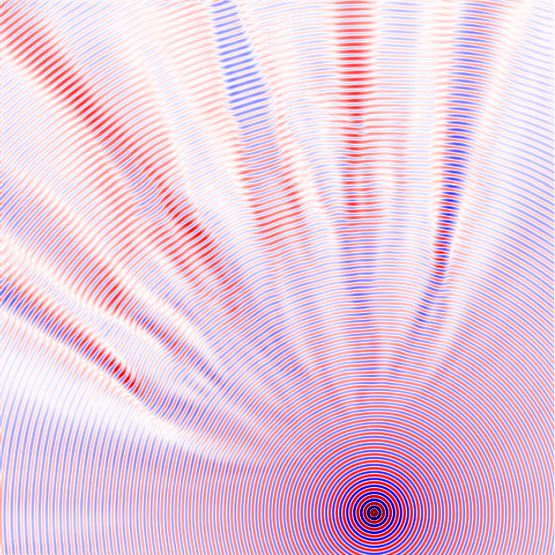}
    \end{minipage}\hfill
    \begin{minipage}[b]{0.16\linewidth}
        \centering
        {\footnotesize FNO(OpenBreastUS)}\\[1mm]
        \includegraphics[width=\linewidth]{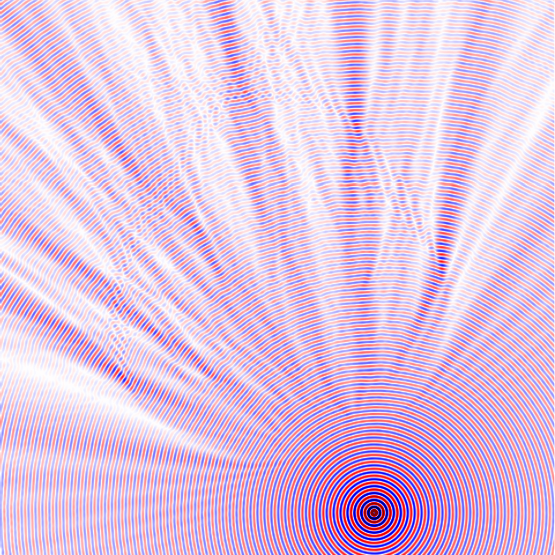}
    \end{minipage}\hfill
    \begin{minipage}[b]{0.16\linewidth}
        \centering
        {\footnotesize UNet(OpenBreastUS)}\\[1mm]
        \includegraphics[width=\linewidth]{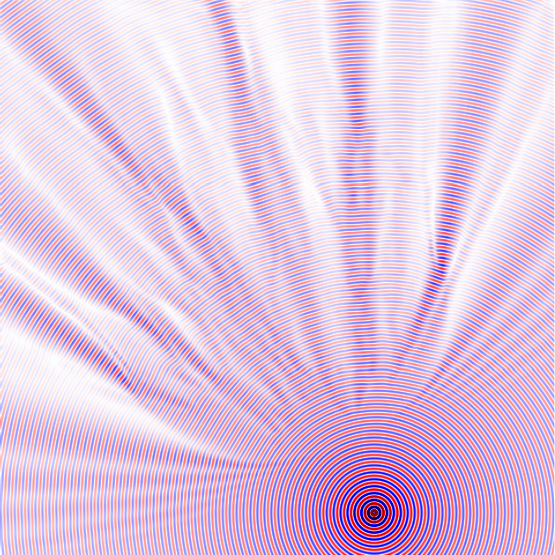}
    \end{minipage}

   \caption{\textbf{Forward simulation results of GRF media and breast phantom at 500kHz.} Comparison of wavefield prediction of GRF media and breast phantom using FNO and UNet. The (GRF) suffix denotes models trained on the GRF dataset, while the (OpenBreastUS) suffix indicates models trained on the OpenBreastUS dataset.}
    \label{fig:grf_openwave_comparison-manual}
\end{figure*}
\subsubsection{Generalization Capability}
In this section, we investigate the out-of-distribution (OOD) generalization capabilities of the representative FNO and NIO models.

\textbf{Source Locations} Figure ~\ref{subfig:num_source_ablation}  illustrates the baseline models' ability to generalize to different wave source locations. We trained the forward neural operators on datasets with varying numbers of source locations (8, 16, 32, 64) and validated them on datasets with unseen sources. As expected, prediction accuracy improves as more training source locations are added. MgNO demonstrates strong generalization to new source locations by effectively capturing the underlying physical principles, even with limited data. As the number of sources increases, FNO's accuracy approaches that of MgNO, while UNet's performance fails to improve, indicating its difficulty in modeling wave propagation. Detailed performance for different models is provided in the Appendix Table~\ref{tab:forward-ood-sources}.

\textbf{Breast Types} Table~\ref{tab:inverse-ood} and \ref{tab:forward-ood} in Appendix show the performance of forward and inverse neural operators trained on selected breast types and tested across all categories. The results show that, for both forward and inverse tasks, performance on OOD samples degrades significantly compared to ID samples. However, neural operators trained on more complex breast types (e.g., heterogeneous) tend to generalize better than those trained on simpler types. Training neural operators on two significantly different breast types (e.g., heterogeneous + fatty) also enhances generalization. Additionally, the performance degradation is less pronounced in forward simulation than in inverse imaging, again suggesting that forward models better capture the underlying physics, while inverse models may tend to memorize anatomical structures.

\textbf{Broader Medium Types} 
To further evaluate the generalization of neural operators trained on OpenBreastUS and to expose the limitations of relying on oversimplified scattering media, we generated a dataset of 6,000 isotropic Gaussian random-field (GRF) phantoms \cite{liu2024wavebench} and simulated their wavefields for 64 sources at eight frequencies . We then compared the performance of models trained on this GRF-based dataset against those trained on OpenBreastUS.

Figure \ref{fig:grf_openwave_comparison-manual} compares the wavefield prediction results of two neural operators (FNO and UNet) trained on the GRF and OpenBreastUS datasets. To evaluate their generalization capabilities, we assessed their performance not only on their respective training data but also on unseen counterparts—breast data for models trained on GRF and GRF data for models trained on OpenBreastUS. The results clearly show that models trained on OpenBreastUS better capture the underlying physics: they perform well on breast data and predict reasonable scattering patterns for GRF media, whereas models trained on GRF fail to generalize to breast phantoms. Notably, while the U-Net trained on the simplified GRF dataset delivers accurate predictions on its in-distribution samples, the U-Net trained on OpenBreastUS fails to produce effective results even on its own in-distribution breast phantoms. This further underscores the tendency of the Simplified dataset to overestimate model performance.

\section{Conclusion}
We introduced OpenBreastUS, a large-scale, anatomically realistic USCT dataset designed to bridge the gap between numerical studies of wave equations and practical imaging applications. OpenBreastUS provides over 16 million frequency-domain wave simulations based on a real USCT system, featuring anatomically accurate human breast phantoms across four density categories. We benchmarked baseline methods for both forward and inverse neural operators in wave imaging tasks, comparing their performance. Our results highlight the strengths and limitations of existing neural operator architectures, providing insights into their generalization capabilities and scalability. OpenBreastUS offers a valuable platform for developing and benchmarking neural wave imaging solvers, enabling their application in real-world imaging tasks involving complex wave phenomena.

While OpenBreastUS represents a significant step toward realistic benchmarking of neural surrogates, it has certain limitations. The dataset is currently limited to breast phantoms; including other organs like brains would enhance its applicability. Simulations are restricted to 2D due to computational constraints; incorporating 3D data would provide a more accurate representation of real-world scenarios. The dataset primarily varies sound speed as the tissue property; incorporating other properties like attenuation could further enhance realism. Additionally, our study focuses on neural operator architectures without extensively exploring the influence of their hyperparameters such as the number of FNO layers. Future work will address these limitations by expanding the dataset's scope and conducting more comprehensive analyses, aiming to provide even more valuable resources for the development of robust neural operators in imaging sciences.

\section*{Acknowledgments}
This work was supported by the National Key Research and Development Program of China (2022YFC3401100), the National Natural Science Foundation of China (12474461) and the Joint Research Project of the Shijiazhuang-Peking University Cooperation Program.

\appendices

\section{VICTRE Generation Implementation}\label{app:VICTRE}
To ensure shape diversity and anatomical realism, we adjusted five key parameters—a1b, a1t, a2l, a2r, and a3—that control the breast’s bottom, top, left, right, and outward scales, respectively (units in cm). These were sampled from truncated Gaussian distributions: 
\[
\begin{array}{ccc}
  a1b, a1t, a2l, a2r   &\sim  & \mathcal{TN}(5.0, 2.0, 3.5, 7.5) \\
 a3/a1b    &  \sim& \mathcal{TN}(1.4, 0.1, 1.0, 1.5)
\end{array}
\]
with \(\sim \mathcal{TN}(\mu, \sigma, a, b)\) representing a truncated Gaussian distribution in the interval \((a, b)\).

To model the internal structure, we first adjusted the \textit{targetFatFrac} parameter to control fat distribution, as it mainly determines the division of four breast types. Typically, the \textit{targetFatFrac} parameter of Extremely Dense, Heterogeneous, Fibroglandular, and Fatty types is respectively in the range of \((0, 0.25)\), \((0.25, 0.5)\), \((0.5, 0.75)\), and \((0.75, 1.0)\). We also fine-tuned the \textit{backFatBufferFrac} parameter in the range of \((0, 0.01)\) to ensure a small fraction of the nipple region to be fat. To define skin properties, we mainly adjusted the following parameters: 
\textit{SkinScale} in the range of \((200,400)\), \textit{SkinScaleNippleDir} in the range of \((5,20)\), and \textit{skinStrength} in the range of \((0.5,2.0)\).

\begin{table*}[!htbp]
\centering
\setlength{\tabcolsep}{3pt}
\begin{tabular}{ccc|ccc}
\hline
\multicolumn{3}{c|}{\textbf{Forward Wave Simulation Baselines}}                                   & \multicolumn{3}{c}{\textbf{Inverse Wave Imaging Baselines}}                                                  \\ \hline
\textbf{Model} & \textbf{\# Parameters} & \textbf{Inference time {[}s{]}}& \textbf{Model}                                                  & \textbf{\# Parameters}  & \textbf{Inference time {[}s{]}} \\ \hline
UNet           & 36.0M                  & 0.015                         & DeepONet                                                         & 36.3M                   & 0.089                         \\
FNO            & 734M                   & 0.018                         & InversionNet                                                     & 55.6M                   & 0.058                         \\
AFNO           & 58.6M                  & 0.013                         & NIO                                                              & 56.3M                   & 0.077                         \\
BFNO           & 104M                   & 0.024                         & \multirow{2}{*}{\makecell{Gradient-based \\ Optim (FNO)}} & \multirow{2}{*}{-}  & \multirow{2}{*}{$\sim$300}        \\
MgNO           & 26.6M                  & 0.015                         &                                                                  &                        &                               \\ \hline
\end{tabular} 
\caption{\textbf{Model size and computational cost.} Comparison of the number of parameters and inference time for baseline models in both forward (Left) and inverse (Right) tasks.}
\label{tab:baseline-overview}
\end{table*}
\section{Implementation Details} \label{app:implementation}
\subsubsection{Forward Baselines}
We trained all forward simulation baseline models for 30 epochs using the AdamW optimizer, with an initial learning rate of 5e-3, decayed by a StepLR scheduler (0.5 decay rate, 10-step size). We used relative L2 loss for training and RRMSE for validation. The detailed architecture of each network is provided below:\\
\textbf{UNet}:We implement UNet using the same structure as \cite{ronneberger2015u} but a increased model size to other baselines for the sake of fairness.  We use the UNet structure with resolution size sequence
 $\{[60]\times 6,[120]\times 5,[240]\times 5,[480]\times 4\}$ and 4 skip channels for Upsample blocks. An input block with the downsample structure using stride 1 is added to the beginning. \\
\textbf{FNO}:We use a vanilla FNO model with 7 FNO layers whose modes are $\{[128]\times 7\}$ and width is $40$ to enlarge the representative ability.\\
\textbf{BFNO}:
The modes and width are set to match those of FNO. Due to its parameter-sharing architecture, BFNO has a smaller parameter size compared to FNO, but its inference time is longer.\\
\textbf{AFNO}:
The adaptive FNO uses multi-head Fourier layers that combines the attention mechanism and Fourier convolution. We set $\text{head} = 4$ and $\text{feature} = 512$ with modes list as $[40]\times 11$. The lifting operator uses Conv2d with patch size = $[4,4]$.\\
\textbf{MgNO}: The model is based on the standard MgNO 
 architecture. In this adaptation, the \textsc{MgConv} modules are modified for the OpenBreastUS dataset by replacing the standard convolution operation with \textsc{dynamical convolution}. The MgNO consists of 6 layers of \textsc{MgConv}. In each \textsc{MgConv}, the number of channels in each convolutional layer increases progressively as the model moves from fine to coarse levels. Specifically, the channel sizes at the five levels are $[24, 32, 64, 128, 256]$.

% \textbf{Training Configuration}:For the forward tast, we trained the forward baseline models for 30 epochs using the AdamW optimizer. The learning rates were initially set to 5e-3 and then decayed by a StepLR scheduler with 0.5 decay rate and 10 stepsize. We employed the relative L2 loss for training and RRMSE for evaluation in all our problems.
\subsubsection{Inversion Baselines}
We trained the three direct inversion baseline models for 500 epochs using the AdamW optimizer, with an initial learning rate of 1e-3 and a weight decay of 1e-6. L1 loss was used for training to preserve edges and fine details in the images, while SSIM and PSNR were used for evaluation.

\textbf{NIO}:In this paper, we modified the original setting of convolutional layers' setting in the Branch net to adapt to the resolution of this problem. For the DeepONet, a CNN with 10 Conv2d layers is applied to obtain a 512 feature coefficients and a linear layer is then applied to map it into 25 basis. The trunk net uses an 8 layer MLP with 100 hidden neurons. For the FNO part, we use 4 Fourier layers with 40 modes and 32 width.\\
\textbf{InversionNet}:WIn this paper, we train the encoder and decoder of InversionNet in a supervised manner, using USCT observations from multiple sources as inputs and predicting 2D sound speed maps (width $\times$ height) as outputs. The convolutional layers are adjusted to accommodate the resolution of this dataset. Additionally, in the USCT setting, we use frequency domain input structured as $\text{Frequencies} \times \text{Receiver} \times \text{Source}$ to correspond with the time domain input $\text{Source} \times \text{Receiver} \times \text{Time}$ as used in seismic FWI, which improves the performance, which improves the model's performance.\\
\textbf{DeepONet}:The implementation of DeepONet is the same as the DeepONet part of NIO. We further use a MLP to map the final 25 basis functions to the outputs.
\subsubsection{Model size and Inference time evaluation}
Efficient computation is critical in wave-based modeling and imaging tasks, particularly for large-scale or real-time applications. Table~\ref{tab:baseline-overview} presents a comprehensive comparison of baseline models in terms of model size (number of parameters) and inference time.
\begin{table*}[!h]
\centering
\begin{tabular}{c|cccc|cccc}
\toprule[1pt]
\multicolumn{1}{c|}{\textbf{Metric}} & \multicolumn{4}{c|}{\textbf{PSNR}$\uparrow$  }                                   & \multicolumn{4}{c}{\textbf{SSIM}$\uparrow$}                                     \\\hline
\multicolumn{1}{c|}{\diagbox{\textbf{Train}}{\textbf{Test}}}                           & \textbf{HET}      & \textbf{FIB}      & \textbf{FAT}      & \textbf{EXD}      & \textbf{HET}  & \textbf{FIB}    & \textbf{FAT} & \textbf{EXD} \\ \hline
HET                                &  15.56               &  11.88               &  6.99                &  12.31               &  0.8194          &  0.7349             &  0.6708         &  0.6826               \\
FIB                                &  11.86               &  \underline{19.22}   &  9.83                &  12.18               &  0.7400          &  \underline{0.8625} &  0.8077         &  0.6353         \\
FAT                                &  7.44                &  9.27                &  \underline{17.69}   &  8.74                &  0.7372          &  0.7739             &\underline{0.9048} &  0.6645         \\
EXD                                &  10.33               &  10.86               &  6.89                &  \underline{17.03}   &  0.6938          &  0.6352             &  0.6650         & \textbf{0.8385 }        \\
All                                   &  \textbf{16.62}      &  \textbf{19.68}      &  \textbf{18.01}      &  \textbf{17.35}      & \textbf{0.8379}    & \textbf{0.8657}    & \textbf{0.9135} & \underline{0.8371}         \\
HET+FAT                               &  \underline{15.92}   &  12.20               &  16.99               &  8.33                & \underline{0.8248} &  0.7247            &  0.9031         &  0.6845         \\ 
FIB+EXD                               &  10.99               &  15.95               &  5.51                &  13.39               & 0.6660             &  0.8396            &  0.6458         &  0.7940         \\ \bottomrule[1pt]
\end{tabular}
\caption{\textbf{Quantitative evaluation of direct inversion baseline (NIO) on OOD breasts.} Each row indicates the breast type(s) used for training, and each column indicates the breast type used for testing. \textbf{Bold}: Best, \underline{Underlined}: Second Best.}
\label{tab:inverse-ood}
\end{table*}

\begin{table*}[h]
\centering
\begin{tabular}{c|cccc|cccc}
\toprule[1pt]
\multicolumn{1}{c|}{\textbf{Metric}} & \multicolumn{4}{c|}{\textbf{RRMSE}$\downarrow$}                                    & \multicolumn{4}{c}{\textbf{Max Error}$\downarrow$}                                \\ \hline
\multicolumn{1}{c|}{\diagbox{\textbf{Train}}{\textbf{Test}}}                           & \textbf{HET}  & \textbf{FIB}    & \textbf{FAT}    & \textbf{EXD}    & \textbf{HET}  & \textbf{FIB}   & \textbf{FAT}   & \textbf{EXD} \\ \hline
HET            &0.0738            & 0.1413             &  0.8113            &  0.5210            & 0.1412           &  0.2033           &  1.0129            &  0.7653          \\
FIB            & 0.2425           & 0.0208             &  0.9284            &  0.6434            & 0.4730           & 0.0523            &  1.0702            &  0.8136         \\
FAT            & 0.4640           & 0.7257             & \textbf{0.0244}    &  0.4966            & 0.6552           &  0.8339           &  \textbf{0.0404}   &  0.9889         \\
EXD            & 0.2668           & 0.6802             &  1.2434            & \textbf{0.0292}    & 0.5269           &  0.9783           &  2.1184            &  0.0687         \\
All              & \textbf{0.0236}   & \textbf{0.0187}    & \underline{0.0270} & 0.0302             & \textbf{0.0417}  & \textbf{0.0318}   &  \underline{0.0446}&  \textbf{0.0584}         \\
HET+FAT          & \underline{0.0269}& 0.5241             & 0.0287             & 0.3147             & \underline{0.0484}& 0.7941            &  0.0545            &  0.5821        \\
FIB+EXD          &  0.1918           & \underline{0.0169} & 0.8983             & \underline{0.0300} & 0.3753            & \underline{0.0349} &  1.0160            &  \underline{0.0610}  \\ \bottomrule[1pt]
\end{tabular}
\caption{\textbf{Quantitative evaluation of forward simulation baseline (FNO) on OOD breasts.} Each row indicates the breast type(s) used for training, and each column indicates the breast type used for testing. \textbf{Bold}: Best, \underline{Underlined}: Second Best.}
\vspace{-0.1in}
\label{tab:forward-ood}
\end{table*}

\begin{table*}[!h]
\centering
\begin{tabular}{c|l|ccccc}
\hline
\multirow{2}{*}{\textbf{Frequency(kHz)}} & \multicolumn{1}{c|}{\multirow{2}{*}{\textbf{Metric}}}              & \multicolumn{5}{c}{\textbf{Models}}                                          \\ \cline{3-7} 
                                    & \multicolumn{1}{c|}{}                                             & \textbf{UNet} & \textbf{FNO} &\textbf{AFNO}         & \textbf{BFNO}        & \textbf{MgNO}         \\ \hline
\multirow{2}{*}{300}               & RRMSE$\downarrow$                                                  &  0.1237       &  0.0347      &  0.0567              & \underline{0.0115}   & \textbf{0.0041}       \\
                                   & Max Error$\downarrow$                                              &  0.2551       &  0.0927      &  0.4447              & \underline{0.0610}   & \textbf{0.0131}       \\ \hline
\multirow{2}{*}{400}               & RRMSE$\downarrow$                                                  &  0.1532       &  0.0426      &  0.1656              & \underline{0.0151}   & \textbf{0.0108}       \\
                                   & Max Error$\downarrow$                                              &  0.2858       &  0.1172      &  1.3172              & \underline{0.0840}   & \textbf{0.0246}       \\ \hline
\multirow{2}{*}{500}               & RRMSE$\downarrow$                                                  &  0.1877       &  0.0632      &  0.2184              & \underline{0.0212}   & \textbf{0.0183}       \\
                                   & Max Error$\downarrow$                                              &  0.3524       &  0.1843      &  1.5160              & \underline{0.0854}   & \textbf{0.0416}       \\ \hline
\end{tabular}
\caption{\textbf{Quantification of the model's generalization to OOD source locations.} Performance was evaluated by training models on the 64 source locations and testing them on the whole 256 sources (192 unseen). \textbf{Bold}:Best, \underline{Underlined}:Second Best}
\label{tab:forward-ood-sources}
\end{table*}
\begin{figure}[!h]
    \centering
    % 用一个局部组把下面两条 \subfloat 都包装起来，这样整个子图都会使用 \footnotesize
    {%
    \footnotesize
    \subfloat[]{%
        \includegraphics[width=0.45\linewidth]{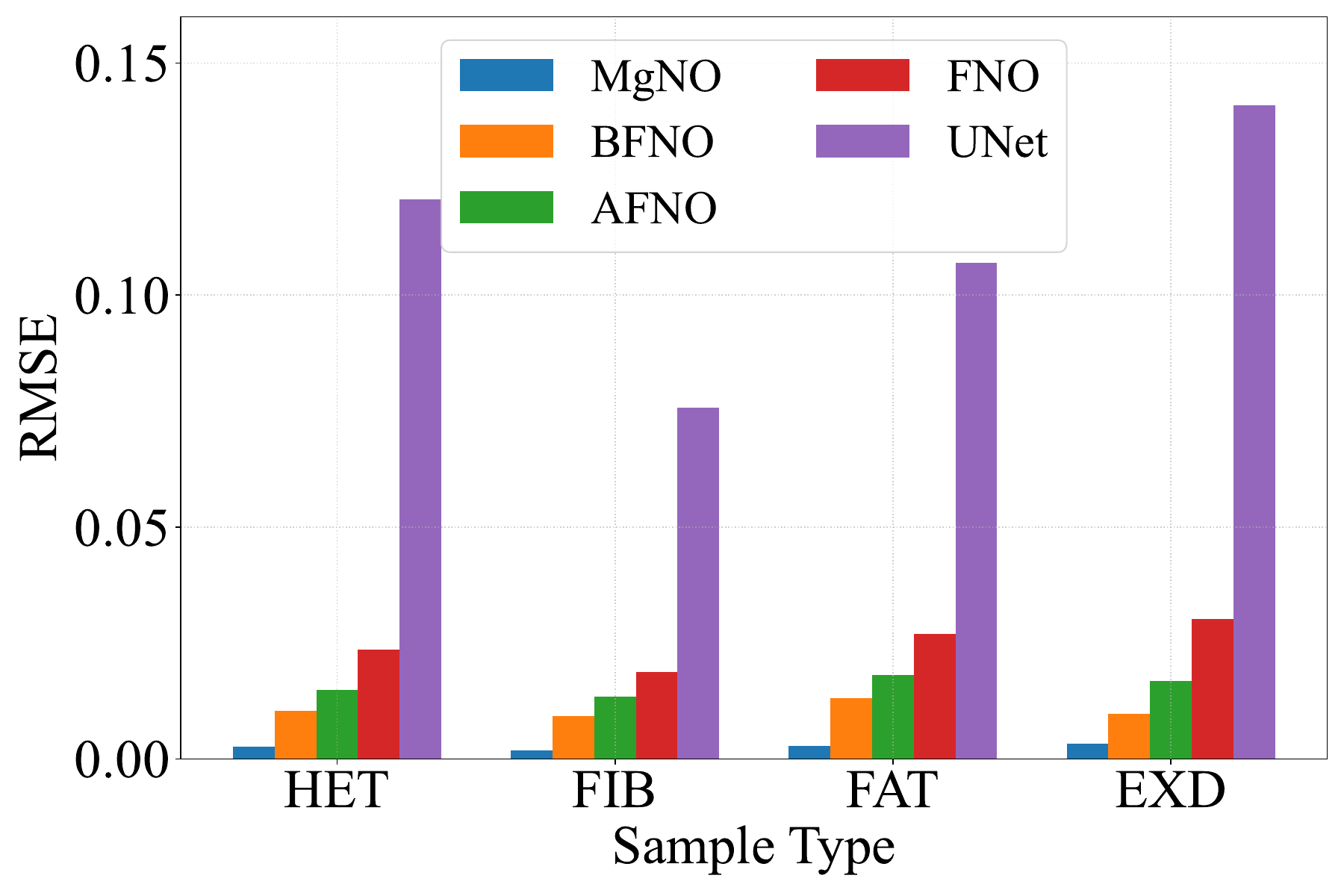}%
        \label{subfig: RRMSE of forward model in types}%
    }\hfill
    \subfloat[]{%
        \includegraphics[width=0.45\linewidth]{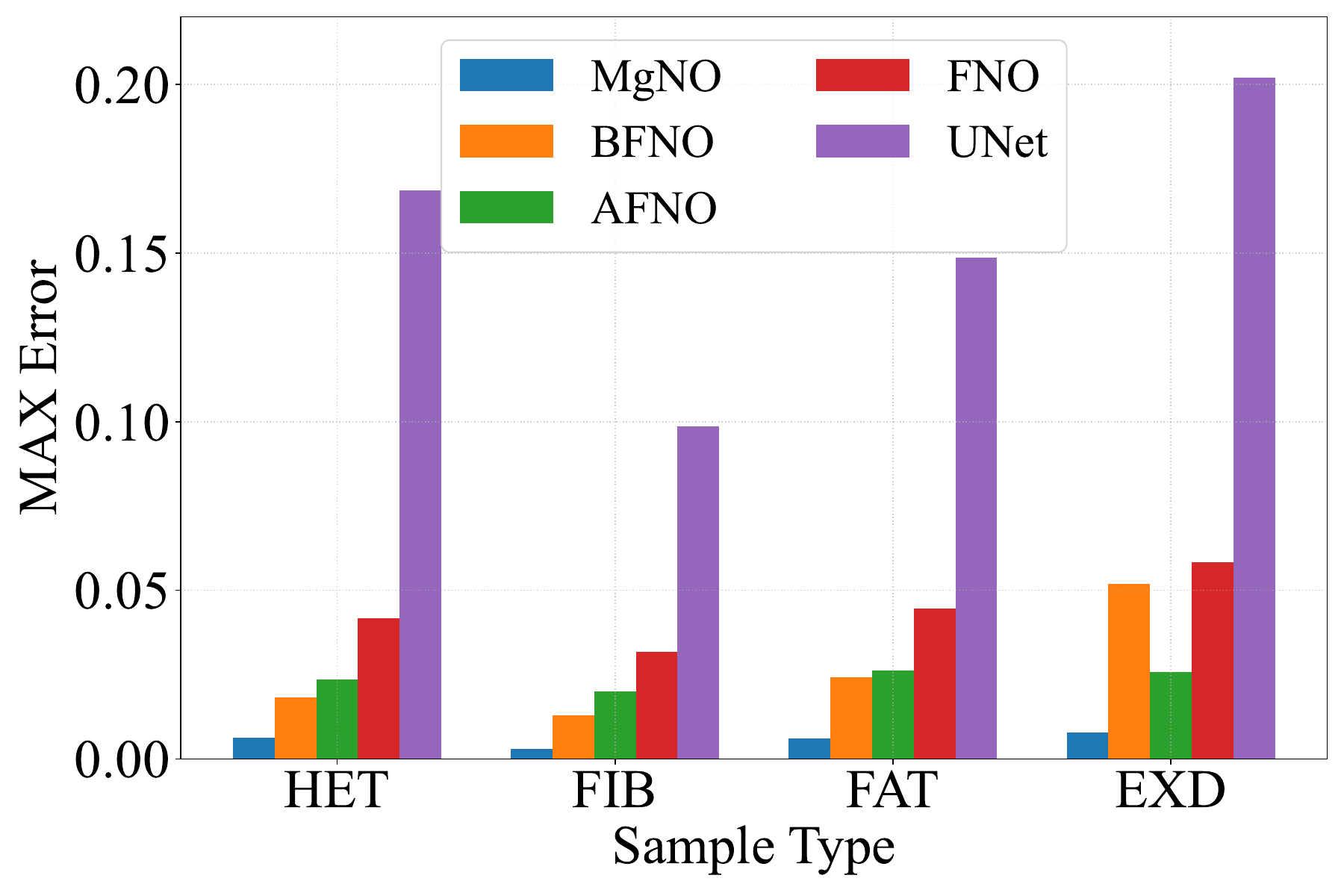}%
        \label{subfig: Max-error of forward model in types}%
    }
    }% 结束局部组
    \vspace{-0.1cm}
 \caption{\textbf{Comparison of forward simulation errors across different breast categories.} RRMSE (a) and Max Errors (b) of five forward simulation baselines are reported across four breast categories. Larger errors in heterogeneous and extremely dense breasts indicate that their more complex internal tissue structures lead to stronger scattering effects and more challenging learning problems.}
    \label{fig:forward-errors}
\end{figure}
\begin{figure}[!htbp]
    \centering
    % 用一个局部组把下面两条 \subfloat 都包装起来，这样整个子图都会使用 \footnotesize
    {%
    \footnotesize
    \subfloat[]{%
        \includegraphics[width=0.45\linewidth]{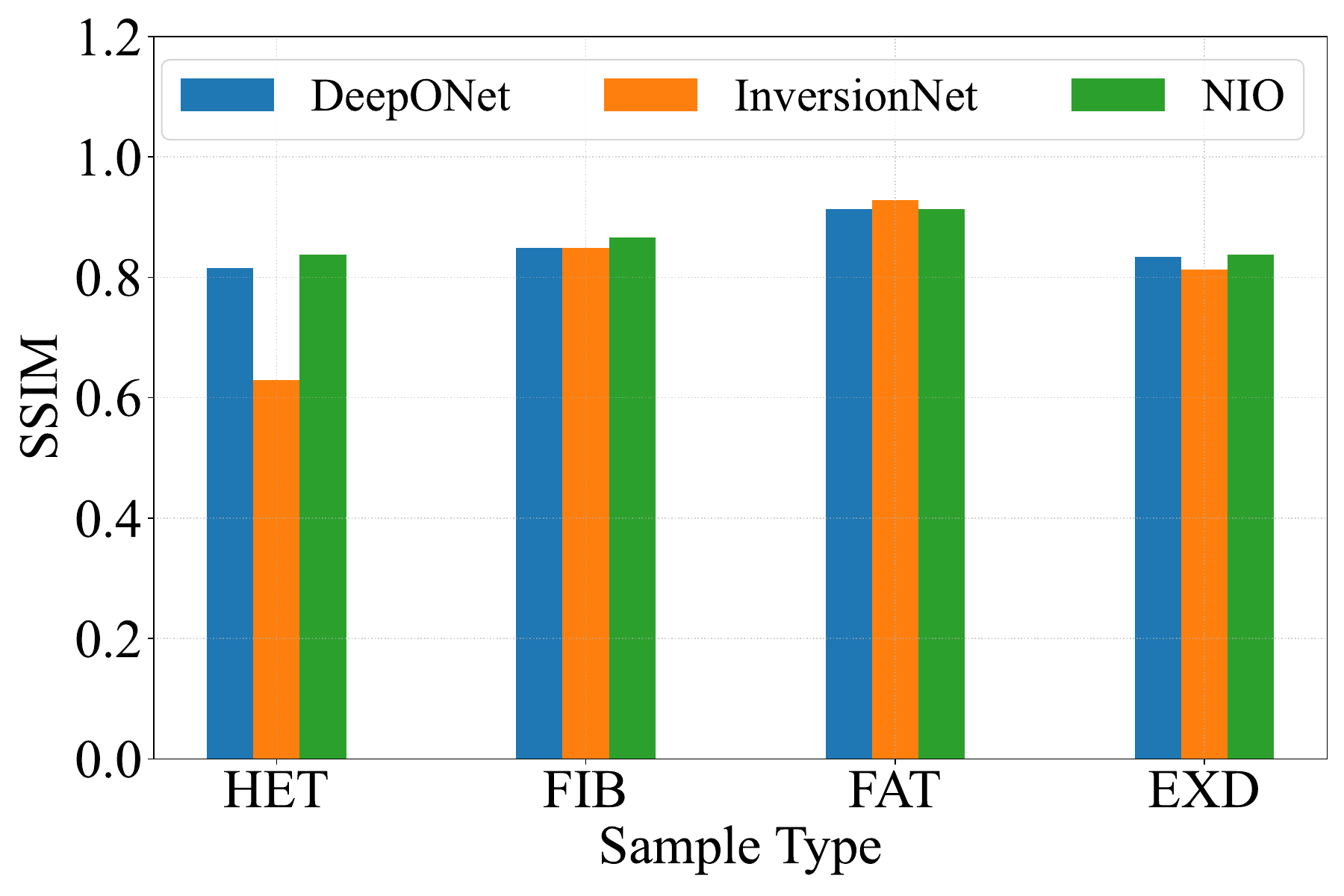}%
        \label{subfig:SSIM_of_inverse}%
    }\hfill
    \subfloat[]{%
        \includegraphics[width=0.45\linewidth]{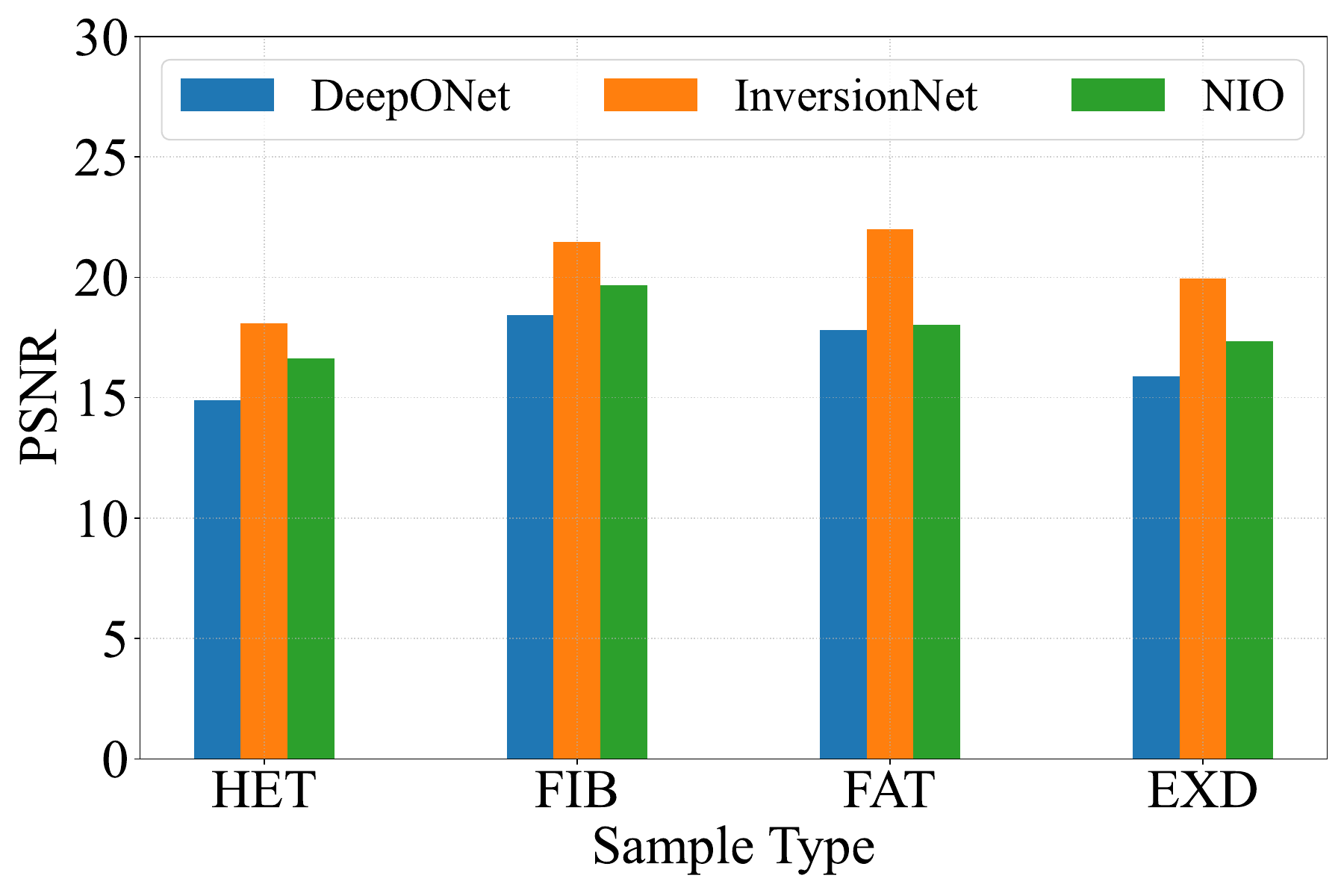}%
        \label{subfig:PSNR_of_inverse}%
    }
    }% 结束局部组
    \vspace{-0.1cm}
    \caption{\textbf{Comparison of direct inversion quality across different breast categories.} 
    SSIM (a) and PSNR (b) of three direct inversion baselines are reported for four breast categories. Lower reconstruction quality in heterogeneous and extremely dense breasts suggests that their more complex internal tissue structures lead to stronger scattering effects and more challenging learning tasks.}
    \label{fig:inverse-errors}
\end{figure}
\begin{figure}[!htbp]
    \centering
    % 用一个局部组把下面两条 \subfloat 都包装起来，这样整个子图都会使用 \footnotesize
    {%
    \footnotesize
    \subfloat[]{%
        \includegraphics[width=0.45\linewidth]{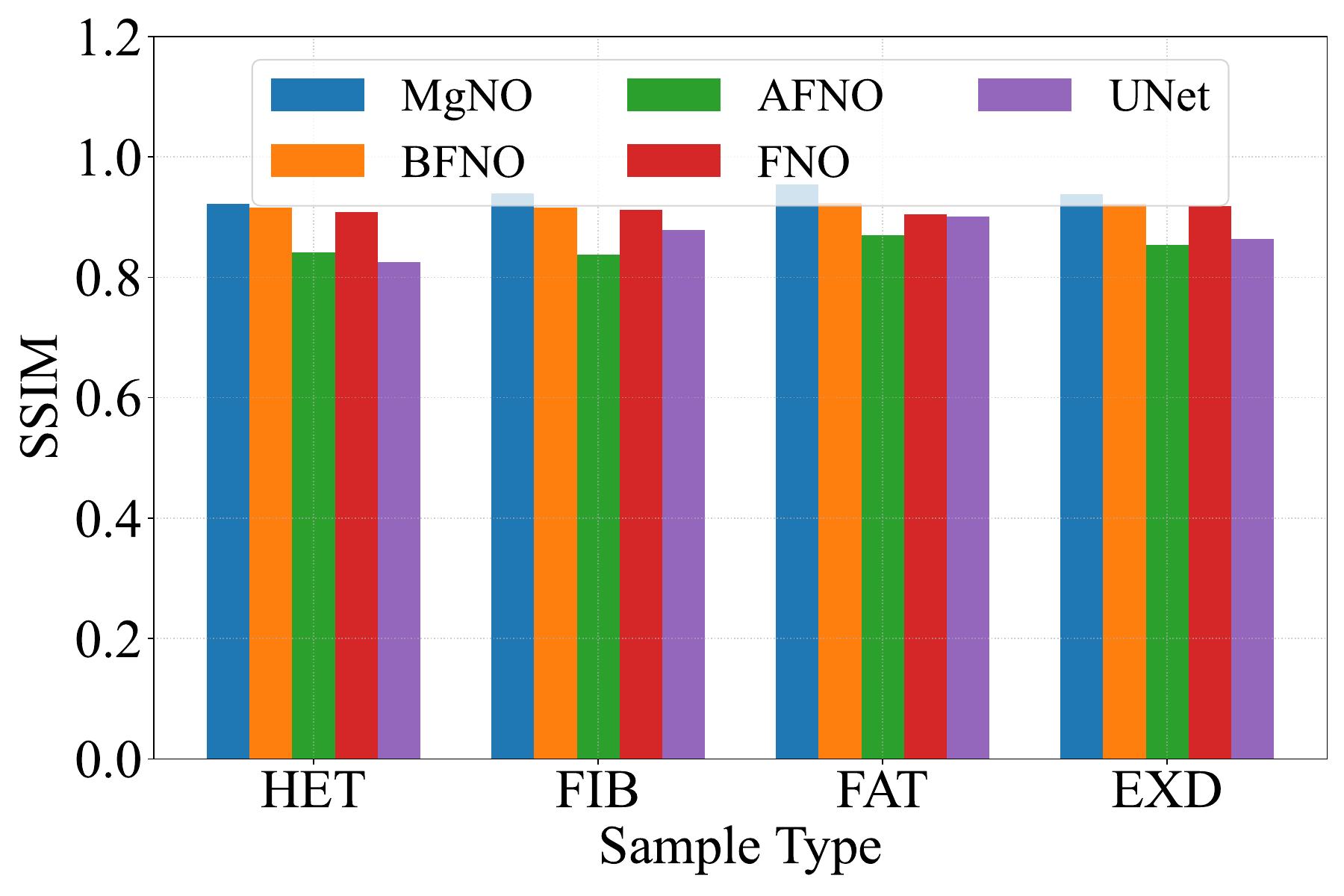}%
        \label{subfig:SSIM_of_inverse_no}%
    }\hfill
    \subfloat[]{%
        \includegraphics[width=0.45\linewidth]{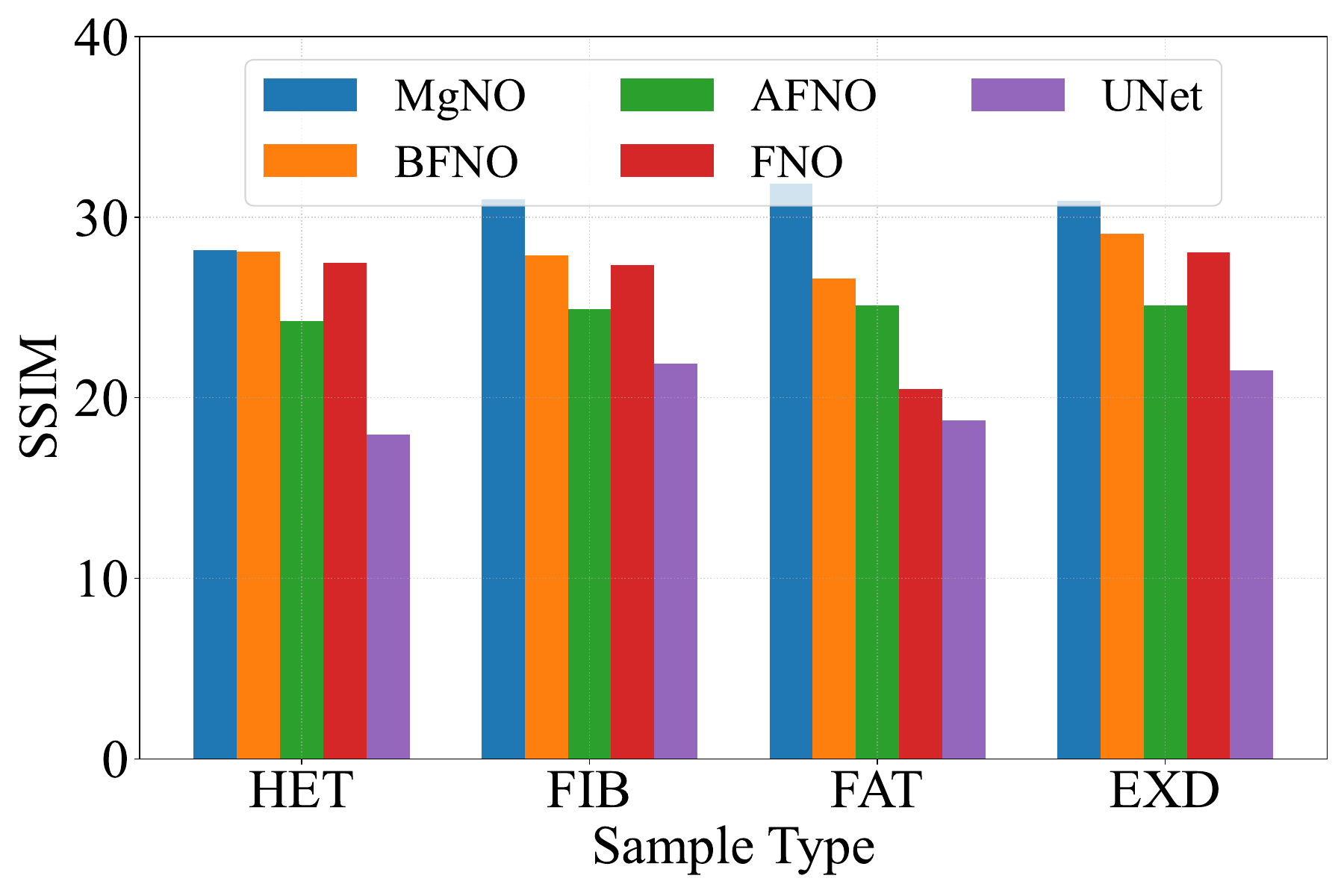}%
        \label{subfig:PSNR_of_inverse_no}%
    }
    }% 结束局部组
    \vspace{-0.1cm}
    \caption{\textbf{Comparison of inversion quality of Gradient-Based Methods across different breast categories.} SSIM (a) and PSNR (b) of three direct inversion baselines are reported for four breast categories.}
    \label{fig:inverse-errors-neural operator}
\end{figure}
\section{Public  USCT Clinical dataset}\label{app:clinic}
The clinical dataset was collected at the Karmanos Cancer Institute (KCI) under Institutional Review Board (IRB) approval No. 040912M1F \cite{ali20242}.
The USCT instrument for data collection employed a 22 cm-diameter ring transducer array with 1024 elements and a pulse center frequency of 2.5 MHz. The system recorded time-series channel data from all 1024 receivers for each individual emitter on the ring, producing a full 1024×1024 matrix. The received channel data was windowed for each transmission, and the discrete-time Fourier transform (DTFT) was applied to isolate the frequencies used in waveform inversion.

\FloatBarrier
\newpage
\bibliographystyle{IEEEtran}
\bibliography{main.bib}

\end{document}